\PassOptionsToPackage{unicode}{hyperref}
\PassOptionsToPackage{hyphens}{url}
\PassOptionsToPackage{dvipsnames,svgnames*,x11names*}{xcolor}
\documentclass[
  11pt,
]{article}
\usepackage{lmodern}
\usepackage{setspace}
\usepackage{amssymb,amsmath}
\usepackage{ifxetex,ifluatex}
\ifnum 0\ifxetex 1\fi\ifluatex 1\fi=0 
  \usepackage[T1]{fontenc}
  \usepackage[utf8]{inputenc}
  \usepackage{textcomp} 
\else 
  \usepackage{unicode-math}
  \defaultfontfeatures{Scale=MatchLowercase}
  \defaultfontfeatures[\rmfamily]{Ligatures=TeX,Scale=1}
\fi
\IfFileExists{upquote.sty}{\usepackage{upquote}}{}
\IfFileExists{microtype.sty}{
  \usepackage[]{microtype}
  \UseMicrotypeSet[protrusion]{basicmath} 
}{}
\makeatletter
\@ifundefined{KOMAClassName}{
  \IfFileExists{parskip.sty}{%
    \usepackage{parskip}
  }{
    \setlength{\parindent}{0pt}
    \setlength{\parskip}{6pt plus 2pt minus 1pt}}
}{
  \KOMAoptions{parskip=half}}
\makeatother
\usepackage{xcolor}
\IfFileExists{xurl.sty}{\usepackage{xurl}}{} 
\IfFileExists{bookmark.sty}{\usepackage{bookmark}}{\usepackage{hyperref}}
\hypersetup{
  pdftitle={Represented but misread: Post-training changes how language models read out what they represent},
  colorlinks=true,
  linkcolor=blue,
  filecolor=Maroon,
  citecolor=Blue,
  urlcolor=blue,
  pdfcreator={LaTeX via pandoc}}
\usepackage[margin=2.5cm]{geometry}
\usepackage{longtable,booktabs}
\usepackage{etoolbox}
\makeatletter
\patchcmd\longtable{\par}{\if@noskipsec\mbox{}\fi\par}{}{}
\makeatother
\IfFileExists{footnotehyper.sty}{\usepackage{footnotehyper}}{\usepackage{footnote}}
\makesavenoteenv{longtable}
\usepackage{graphicx}
\makeatletter
\def\maxwidth{\ifdim\Gin@nat@width>\linewidth\linewidth\else\Gin@nat@width\fi}
\def\maxheight{\ifdim\Gin@nat@height>\textheight\textheight\else\Gin@nat@height\fi}
\makeatother
\setkeys{Gin}{width=\maxwidth,height=\maxheight,keepaspectratio}
\makeatletter
\def\fps@figure{htbp}
\makeatother
\usepackage{caption}
\usepackage{float}
\usepackage{newunicodechar}
\usepackage{etoolbox}
\AtBeginEnvironment{longtable}{\footnotesize\setlength{\tabcolsep}{4pt}}

\emergencystretch=3em \newunicodechar{→}{\ensuremath{\rightarrow}}
\newunicodechar{←}{\ensuremath{\leftarrow}}
\newunicodechar{√}{\ensuremath{\surd}}
\newunicodechar{≈}{\ensuremath{\approx}}
\newunicodechar{≤}{\ensuremath{\leq}}
\newunicodechar{≥}{\ensuremath{\geq}}
\newunicodechar{ᵃ}{\textsuperscript{a}}
\ifluatex
  \usepackage{selnolig}  
\fi
\newlength{\cslhangindent}
\newenvironment{cslreferences}%
  {\setlength{\parindent}{0pt}%
  \everypar{\setlength{\hangindent}{\cslhangindent}}\ignorespaces}%
  {\par}

\title{Represented but misread: Post-training changes how language
models read out what they represent}
\author{Jochen Laubrock\\ {\normalsize University of Potsdam}}
\date{19 November 2026}

\begin{document}
\maketitle
\begin{abstract}
Does post-training change what a language model represents, or how this
representation is read out into an answer? We compared linear probes
with the models' own answers, analyzed with signal detection theory, in
six open-weight base models and their post-trained versions. For
judgments of whether German texts were written by a human or an AI,
probes and word n-gram classifiers separated the classes well, but the
answers did not and often ordered the texts inversely. Post-training
lowered the sensitivity of the answers on average (Δd′ = −0.34 across
model families; −0.29 across three prompts). It also changed the use of
a stated base rate: Base models shifted their log-odds by less than the
stated change in prior odds, most post-trained models by more. In two
models with stage-wise checkpoints, the change occurred during
supervised fine-tuning, model stitching implicated the early layers, and
from mid-depth on internal states at positions after the text, not those
of the text, transferred the change between checkpoints. Our corpora
cannot be shared, so we add a case reproducible from public data, the
truth of negated statements. Here, post-training acted in the opposite
direction: Probes decoded truth at or near ceiling in every model, the
answers of several base models did not use it, and post-training
repaired them; by mid-depth the repair resided at the positions after
the statement; below it, the representation of the statement
contributed, strongly in one of four models. In a training study, adding
2,000 correctly labeled authorship judgments of news texts to otherwise
identical fine-tuning data improved the answers on our newspaper and
textbook corpora and inverted them on the chats, where they followed a
news-trained probe of the base model. Post-training thus changed how a
judgment is read out of what a model represents, and a small training
signal can set its direction.
\end{abstract}

\setstretch{1.15}
\hypertarget{introduction}{%
\section{1 Introduction}\label{introduction}}

What does post-training change in a language model? After pretraining on
large text corpora, language models are turned into assistants in
several stages of post-training: supervised fine-tuning (SFT), in which
the model learns to imitate example dialogues, and stages that optimize
it for human or AI preferences, e.g., direct preference optimization
(DPO), in which it learns to prefer the better of two candidate answers.
An influential view holds that these stages add little knowledge: A
model learns what it knows during pretraining, and post-training mainly
teaches it in which format and style to respond. Zhou et al.
(\protect\hyperlink{ref-zhou2023lima}{2023}) formulated this as the
superficial alignment hypothesis, and several findings are consistent
with it, among them that a few well-chosen examples or a suitable prompt
to a base model can produce much of the behavior of a fully post-trained
model (Lin et al., \protect\hyperlink{ref-lin2023urial}{2023}), and that
instructions change how a model expresses information more than how it
encodes it (Waldis et al.,
\protect\hyperlink{ref-waldis2026instructions}{2026}). If this view is
correct, post-training should leave a model's internal representations
largely intact while changing how they are read out into an answer. This
distinction matters beyond its theoretical interest: A model whose
answers no longer reflect what it represents knows something but says
something else, which is a central concern for assessing the honesty of
language models. Testing the distinction, however, requires a judgment
for which we know what the model should represent, can measure whether
it does, and can pinpoint where the representation diverges from the
answer.

We argue that judging the authorship of a text, i.e., whether it was
written by a human or generated by an AI, is such a test case, for
several reasons. The correct answer is known for every text, so that
both the model's representation and its answer can be scored against the
same ground truth. Authorship is a property of the input text, which the
model reads in full before it answers; the information is therefore
available in principle and does not depend on retrieving facts from
memory. Whether the distinguishing information is present can be
established independently of the model, with simple text statistics.
Texts can be chosen that the models are unlikely to have seen, or cannot
have seen, during training, which rules out memorization. Humans make
the same judgment, which connects the question to a long tradition of
research on human judgment. The judgment is also of practical relevance,
because AI-generated text has become hard to tell apart from human
writing in education, journalism, and science, human readers perform
close to chance (Clark et al.,
\protect\hyperlink{ref-clark2021human}{2021}; Jakesch et al.,
\protect\hyperlink{ref-jakesch2023human}{2023}; Jones \& Bergen,
\protect\hyperlink{ref-jones2025turing}{2025}; Köbis \& Mossink,
\protect\hyperlink{ref-kobis2021angelou}{2021}; Porter \& Machery,
\protect\hyperlink{ref-porter2024poetry}{2024}), and detectors trained
specifically for this purpose are in wide use (Thai et al.,
\protect\hyperlink{ref-thai2026editlens}{2025}). Human judgments rely on
plausible but invalid heuristics, such as taking references to personal
experience or first-person pronouns as signs of human authorship
(Jakesch et al., \protect\hyperlink{ref-jakesch2023human}{2023}), and
only readers who frequently use language models for their own writing
appear to detect AI-generated text reliably (Russell et al.,
\protect\hyperlink{ref-russell2025frequent}{2025}). Because the texts
that make authorship a clean test case cannot be shared, we complement
it with a second judgment on public data that meets the same
requirements and can be reproduced in full: whether a statement is true,
for statements with and without negation (Marks \& Tegmark,
\protect\hyperlink{ref-marks2024geometry}{2024}).

One might expect language models to do better. They produce such texts
themselves, and the regularities that distinguish their output from
human writing are, in principle, available to them. However, language
models asked directly whether a text was written by a human or by an AI
are weak and biased judges: Their errors lean in one direction, and the
direction differs between models and model generations (Bhattacharjee \&
Liu, \protect\hyperlink{ref-bhattacharjee2024fighting}{2024}; Yuan et
al., \protect\hyperlink{ref-yuan2026crossgen}{2026}). When GPT-3.5 and
GPT-4 judged transcripts of Turing-test conversations, their accuracy
was even below chance (Rathi et al.,
\protect\hyperlink{ref-rathi2024gpt4}{2024}). In pilot work with the
German chat messages used here, four commercial models judged each
message with five prompt versions and rated their confidence. Two of
them, GPT-5.4 and Claude Opus 4.6, had written the AI messages
themselves. None of the four models recognized these messages: With any
prompt version, each model called at most 16\% of them AI, and the two
generators called at most 7\% and 14\% of their own messages AI. The
confidence ratings ordered the messages below chance in the smallest
model (GPT-5.4-mini: AUC = .20--.34) and at or above chance in the
larger ones (GPT-5.4: .53--.73; Claude Sonnet 4.6: .63--.76; Claude Opus
4.6: .69--.90). Why would a model that writes like an AI fail to
recognize AI-written text, and why would the training that makes it a
helpful assistant make it worse at this?

Two classic frameworks from psychology are helpful in asking this
question precisely. Signal detection theory separates the sensitivity of
a judge, i.e., the ability to distinguish signal from noise, from the
criterion, i.e., the tendency to give one answer rather than the other
(Macmillan \& Creelman,
\protect\hyperlink{ref-macmillan2005detection}{2004}). A bias toward
answering ``human'' only shifts the criterion for all texts alike; it
cannot make the evidence for ``AI'' stronger for human than for AI
texts. Sensitivity below chance therefore indicates that the evidence on
which the judgment is based is ordered inversely. Brunswik's lens model
adds the distinction between the validity of the cues that a stimulus
offers and the way in which a judge utilizes them (Brunswik,
\protect\hyperlink{ref-brunswik1956perception}{1956}; Karelaia \&
Hogarth, \protect\hyperlink{ref-karelaia2008lens}{2008}). Poor judgments
can result from a lack of valid cues or from valid cues that are not
used, or used in the wrong way. For human judges, the utilization of
cues can only be inferred from their judgments. For language models, it
can be examined directly, because their internal representations are
accessible. The judgment of a language model can then be described by
two lenses in series (Figure 1): In a first lens, the cues of the text
are encoded in the model's internal representation; in a second lens,
this representation is read out into an answer to the question.
Splitting the judge side of the lens model in this way has precedents in
research on personality judgment, which distinguishes the detection of
cues from their utilization (Funder,
\protect\hyperlink{ref-funder1995accuracy}{1995}), and in neuroscience,
which distinguishes the information that a neural population encodes
from the information that is read out for behavior (Panzeri et al.,
\protect\hyperlink{ref-panzeri2017cracking}{2017}). Because psychology
and machine learning often use different words for the same idea, and
sometimes the same word for different ideas, Appendix A lists the
correspondences between their terms.

\begin{figure}
\centering
\includegraphics[width=1\textwidth,height=0.68\textheight]{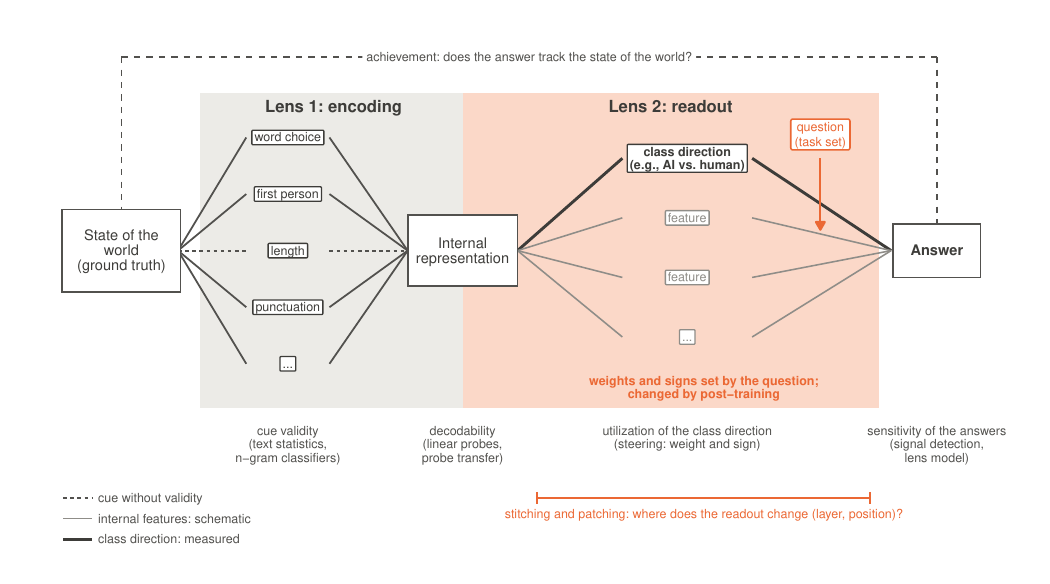}
\caption{Brunswik's lens model, applied to a language model as judge and
extended by an internal lens. A state of the world (Brunswik's distal
criterion; the ground truth, here whether a text was written by an AI,
whether a statement is true, or whether a review is genuine) is
reflected in observable cues of the text, some of which are valid and
some of which are not (dashed). In the first lens (encoding), the model
integrates these cues into its internal representation; whether the
ground truth can be decoded from this representation is measured with
linear probes, and whether two checkpoints represent it in the same way
with probe transfer. In the second lens (readout), the representation is
read out into an answer to the question by weighting internal features.
Only one of these features was measured directly: the class direction,
i.e., the direction that separates the two classes in the
representation; its relation to the answer was examined with steering at
the answer position (Appendix D). The other features are schematic.
Achievement, i.e., the agreement of the answer with the state of the
world, requires both lenses. Post-training changes mainly the second
lens (orange), and in several cases also the first (Sections 3.4 and
3.6); model stitching and activation patching locate the change in depth
and in the prompt. Splitting the judge side of the lens model has
precedents in personality judgment (detection vs.~utilization of cues;
Funder, \protect\hyperlink{ref-funder1995accuracy}{1995}) and in
neuroscience (information encoded in a neural population vs.~read out
for behavior; Panzeri et al.,
\protect\hyperlink{ref-panzeri2017cracking}{2017}). Note: for the
authorship corpora, probes were fitted both on the text positions and on
the last prompt position; for the truth and review sets, only on the
last prompt position.}
\end{figure}

A brief description of how a language model processes a prompt may be
helpful for readers from psychology. The prompt is split into a sequence
of tokens (words or parts of words), and in each of its several dozen
layers the model computes an internal state for every position of this
sequence. Through the attention mechanism, the state at a position can
draw on the states at all earlier positions, but not on later ones. The
answer is generated from the state at the last position after the final
layer. In a prompt that presents a text and then asks a question about
it, information about the text can therefore reach the answer only
through the positions that follow the text, i.e., those of the question
and of the cue for the answer (Figure 2A). We refer to these parts of
the prompt as the text positions, the question positions, and the last
position. All analyses and interventions reported below are carried out
on these internal states of an artificial neural network, i.e., of a
computer program whose every state can be recorded, replaced, and
restored.

\begin{figure}
\centering
\includegraphics[width=1\textwidth,height=0.68\textheight]{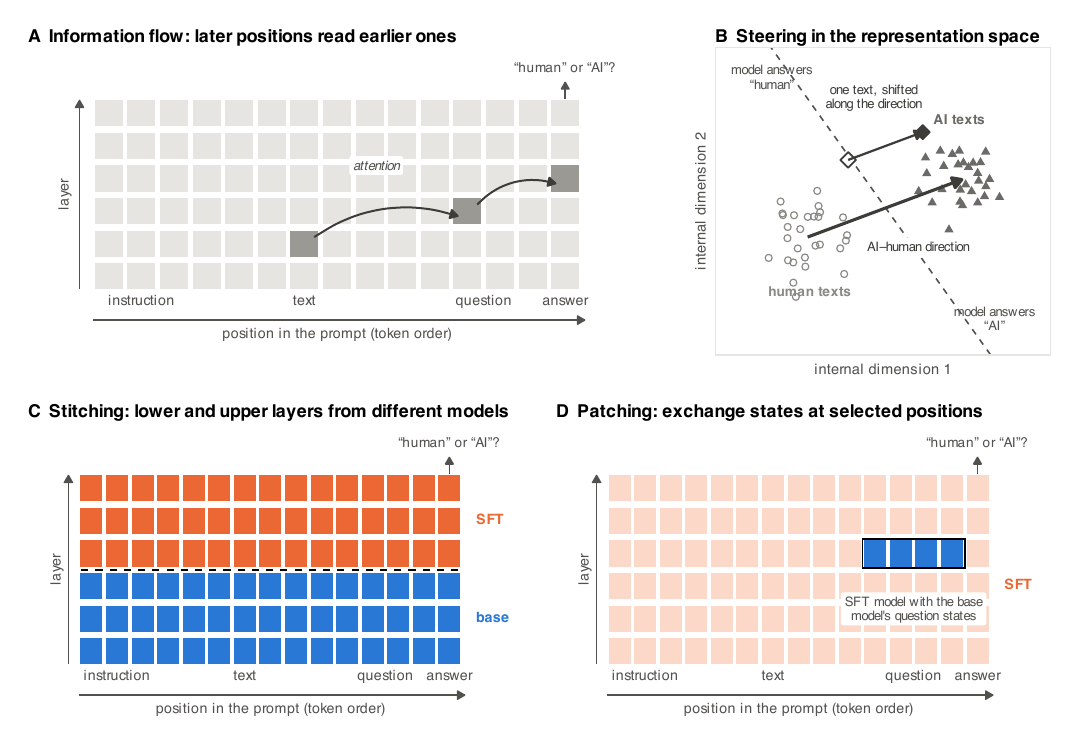}
\caption{How the interventions work. In panels A, C, and D, the
horizontal axis shows the positions of the prompt in reading order (the
instruction, the text to be judged, the question, and the position at
which the answer is produced), and the vertical axis shows the layers of
the network. (A) Information flow: through attention, later positions
read information from earlier ones. The answer is produced at the last
position. Information from the text reaches it through the positions
after the text: the question (dark tiles) and the last position itself,
which also reads the text directly at every layer (the schematic shows
only the route via the question). (B) Steering, sketched in a
two-dimensional version of the high-dimensional representation space:
human texts (open circles) and AI texts (triangles) differ along an
AI--human direction (arrow between the class means). Steering moves the
state of one text along this direction (diamonds) and measures how the
model's answer changes; the dashed line marks where the answer switches.
The two need not be orthogonal: the angle between the AI--human
direction and the model's decision boundary determines how strongly a
shift along the direction changes the answer. If the model read the
direction correctly, a shift toward the AI texts would move the answer
toward ``KI''. (C) Model stitching combines the lower layers of one
checkpoint (here the base model, blue) with the upper layers of another
(here the SFT model, orange). (D) Activation patching replaces the
internal states of one model (here SFT) at selected positions and one
layer with those of the other model (here the base model's states at the
question positions).}
\end{figure}

This possibility has been used to show that language models often
represent more than they say. Simple linear classifiers (``probes'')
trained on a model's internal activations can recover properties such as
the truth of a statement (Burns et al.,
\protect\hyperlink{ref-burns2023discovering}{2024}; Hu \& Levy,
\protect\hyperlink{ref-hu2023prompting}{2023}; Marks \& Tegmark,
\protect\hyperlink{ref-marks2024geometry}{2024}) or the correctness of
the model's own answers (Orgad et al.,
\protect\hyperlink{ref-orgad2025know}{2025}) better than the model's
answers do, and prompting a model for a judgment can underestimate what
it knows (Hu \& Levy, \protect\hyperlink{ref-hu2023prompting}{2023}).
The same dissociation has been reported for authorship: Models encode
whether they wrote a text themselves although they fail to report it (Y.
Zhou et al., \protect\hyperlink{ref-zhou2025cosur}{2026}), and a
direction in the residual stream that tracks self-authorship can be used
to steer what a model claims to have written (Ackerman \& Panickssery,
\protect\hyperlink{ref-ackerman2025inspection}{2025}; see also
Panickssery et al.,
\protect\hyperlink{ref-panickssery2024evaluators}{2024}).
Interpretability research at the scale of frontier models has made the
same distinction between the presence of a feature and the behavior it
supports (Templeton et al.,
\protect\hyperlink{ref-templeton2024scaling}{2024}), and has identified
a small subspace of representations that a model is able to verbalize,
whose content is shaped by post-training (Gurnee et al.,
\protect\hyperlink{ref-gurnee2026verbalizable}{2026}). Why represented
information fails to reach the answer is less clear. One recent account
attributes such gaps to a miscalibrated decision threshold (Villuri et
al., \protect\hyperlink{ref-villuri2026probes}{2026}), i.e., to the
criterion in terms of signal detection theory, which cannot explain
sensitivity below chance. Another line of work shows that steering along
a direction that discriminates a concept can promote the opposite
behavior (Torop et al.,
\protect\hyperlink{ref-torop2026inverted}{2026}), which suggests that
the readout of represented information can itself be reversed.

What post-training does to this relation has been studied mainly by
comparing base and post-trained models. Representations learned during
pretraining largely transfer to the post-trained model (Kissane et al.,
\protect\hyperlink{ref-kissane2024saetransfer}{2024}), fine-tuning tends
to enhance existing mechanisms rather than to create new ones (Prakash
et al., \protect\hyperlink{ref-prakash2024finetuning}{2024}). Model
diffing, i.e., the systematic comparison of the internal representations
of a base model and its post-trained version, has been carried out with
crosscoders, sparse dictionaries of features that are learned from both
models at once and reveal which features are shared and which are
specific to one of them (Lindsey et al.,
\protect\hyperlink{ref-lindsey2024crosscoders}{2024}). It suggests that
much of what chat fine-tuning adds is concentrated at the tokens of the
chat template (Minder et al.,
\protect\hyperlink{ref-minder2025crosscoders}{2026}; see also Leong et
al., \protect\hyperlink{ref-leong2025template}{2025}), i.e., the special
tokens that mark the beginning and the end of the user's and the
assistant's turns and that, in a prompt, lie between the user's text and
the answer. Likewise, exchanging the key-value cache, i.e., the stored
internal states of earlier positions that later positions read, between
fine-tuned and base models transfers fine-tuned behaviors mainly through
system-prompt and template positions (Loos \& Eberle,
\protect\hyperlink{ref-loos2026kvpatching}{2026}), and crossing the
lower and upper layers of base and post-trained models has shown that
changes in the late layers can depend on altered earlier computation (Y.
Zhou, \protect\hyperlink{ref-zhou2026convergence}{2026}). For example,
the representation of whether a model is being evaluated, tracked across
the published checkpoints of one model, is already present in the base
model and amplified during supervised fine-tuning (Heidari et al.,
\protect\hyperlink{ref-heidari2026evalawareness}{2026}). To our
knowledge, however, these approaches have not been combined to ask, for
a judgment with known ground truth, whether post-training changes the
representation of a property or the way in which the answer reads it
out, and where in the network this happens. Models whose developers
released checkpoints after every stage of post-training make this
possible. Combining the lower layers of one checkpoint with the upper
layers of another, a technique known as model stitching (Bansal et al.,
\protect\hyperlink{ref-bansal2021stitching}{2021}; Lenc \& Vedaldi,
\protect\hyperlink{ref-lenc2015equivariance}{2015}), locates a change in
depth; exchanging internal states between checkpoints at selected
positions of the prompt, an application of activation patching
(Heimersheim \& Nanda,
\protect\hyperlink{ref-heimersheim2024patching}{2024}; Meng et al.,
\protect\hyperlink{ref-meng2022rome}{2023}; Vig et al.,
\protect\hyperlink{ref-vig2020causal}{2020}), locates it in the prompt.

Here we use authorship judgments as a test case for representation
versus readout under post-training. We studied six open-weight base
models and their post-trained versions with three corpora of German
texts: chat messages that students wrote for this study and that were
never published, so that they cannot have been part of any training
data; segments of articles from a local newspaper that, to our
knowledge, were not available in digital form and are therefore unlikely
to have been part of the training data; and summaries of textbook
chapters. First, we tested whether the information that distinguishes AI
from human texts is available in the models' representations, and
compared it with simple text statistics. Second, we measured with signal
detection theory how the models' own answers use this information, how
post-training changes it from stage to stage, whether labeled examples
can correct it, and whether a stated base rate moves the answers as it
would for an ideal observer. Third, we located the effect of
post-training by combining the lower and upper layers of base and
post-trained checkpoints (``model stitching'') and by exchanging
internal states between checkpoints at selected positions of the prompt
(``activation patching''). Fourth, because our corpora cannot be shared,
we repeated the central analyses with a judgment on public data that can
be reproduced in full, the truth of statements with and without
negation. Fifth, we manipulated the training data directly, in a
training study with control conditions, to test whether the training
signal sets the direction of the readout. We find that the authorship
information was readily available but that the models' answers did not
use it, and that post-training lowered the sensitivity of the answers on
average. In the two models with stage-wise checkpoints, the weights that
carried this change resided in the early layers, and a sweep over layers
showed that from mid-depth onward the states at the positions after the
text, and not those of the text, transferred it between checkpoints;
below mid-depth, the text states contributed in one of the two models.
For negated statements, post-training acted in the opposite direction:
It repaired answers that several base models gave poorly although the
truth could be decoded at ceiling, through weights in the middle layers;
by mid-depth the repair resided at the positions after the statement,
and below that the representation of the statement itself contributed in
the four models in which we traced it, strongly in one of them. In the
training study, a small number of labeled examples shifted the readout
in the direction that their labels implied, with little change of the
probes at the text positions. A further judgment, whether a hotel review
is genuine, gave less conclusive results (Appendix C). Insofar as the
representation of the text is concerned, our results are largely
consistent with the superficial alignment hypothesis; at the same time,
they show that the change is not superficial in the sense of affecting
only the final stages of processing, and that it can make a judgment
considerably better or worse.

\hypertarget{method}{%
\section{2 Method}\label{method}}

\hypertarget{materials}{%
\subsection{2.1 Materials}\label{materials}}

We used three corpora of German texts in which human-written texts are
matched with AI-generated counterparts. In each corpus, the human
original and all texts derived from it form one group (\texttt{source}),
and all analyses treat this group as the unit of resampling and
cross-validation, so that no result can rest on recognizing a topic.

\emph{Chat messages.} The human messages were written in a session of an
experimental psychology lab course: Students were asked to write five
short chat or social-media posts each on current topics; example topics
were provided, but students were free to choose their own, and several
did so; not every student provided five messages. The messages were
never published. AI counterparts were generated by two large language
models (Claude Opus 4.6 and GPT-5.4) on the same topics, in three steps.
All prompts placed the model in a private German group chat with friends
or family and asked for a single message of about one to four short
sentences, without hashtags. The basic prompt (``naive'') added only
that the message should sound natural, spontaneous, and everyday, as in
a real group chat rather than an essay. The style-matched prompt
additionally provided the other messages of the same student, on other
topics, as style references (tone, sentence length, directness,
colloquialism, punctuation, spontaneity). Finally, each style-matched
message was revised according to written feedback on what still sounded
AI-like, which the students gave on messages that they had not written
themselves (``optimized''). The students' messages and their feedback
cannot be shared, because the students have not yet been asked for
consent to publication; the prompt templates, which contain placeholders
only, are included in the project repository (\texttt{prompts/}).
Because AI and human messages differed considerably in length, we
truncated every message to its first 25 words and discarded shorter
messages. The resulting corpus comprised 300 texts (61 human, 239 AI)
from 61 groups. We verified the labels of this corpus: All human
messages are verbatim the students' submissions, all AI messages trace
to the generation pipeline, and human participants of a separate
experiment recognized the messages in the labeled direction (AUC = .82;
Rosin \& Laubrock, in preparation). Whether individual students used AI
tools to write their messages cannot be determined. Because the
generation pipeline had also stored the style-matched messages before
the feedback revision, we additionally scored these 122 messages (same
generators and truncation) as a control that separates the effect of
style matching from that of the revision.

\emph{Newspaper segments.} Articles written by a local journalist for a
local newspaper, which to our knowledge were not available in digital
form and are therefore unlikely to have been part of any model's
training data, were paired with generated articles. For each article, a
locally run language model (Gemma 4 27B) extracted a summary of the
topic, the key facts (names, places, dates, numbers), and the length and
number of paragraphs, and it derived a profile of the author's style
from the author's other articles (leave-one-out). GPT-5.4 then wrote an
article of the same length and paragraph count that incorporated all key
facts, either from the topic and facts alone (``naive'') or additionally
following the style profile (``style-matched''); both prompts were
written in English, did not use web search, and instructed the model not
to reveal AI authorship. The style-matched articles were finally revised
to avoid nine surface features that the students of the lab course had
identified as AI markers in a sample of the articles (e.g., dashes,
lists of exactly three items, ``not only \ldots{} but also''
constructions, effusive closing sentences, uniform paragraph lengths,
too few numbers; ``optimized''). The templates for the extraction and
writing steps are included in the project repository; the revision
prompt, which contains the students' list of features, is not. We used
the 103 segments (38 human, 65 AI) from 14 articles that had been
presented in a human judgment experiment, each preceded by its headline
and subheadline, as in that experiment. These texts are confidential and
were processed only on university machines.

\emph{Textbook summaries.} The third corpus consisted of summaries of
119 chapters of German textbooks in biology, psychology and business
administration, each available as the human summary from the textbook
and as a summary generated by ChatGPT 5.3 (Rosin \& Laubrock, in
preparation). The 238 texts form 119 matched pairs. The human summaries
are copyrighted and were processed only on university machines.

Because these corpora cannot be shared, we repeated the central analyses
with truth judgments on public statement sets (Sections 2.8 and 3.6),
which can be reproduced in full.

\hypertarget{models}{%
\subsection{2.2 Models}\label{models}}

We studied six open-weight base models with 4 to 12 billion parameters
from four developers (Table 1): Gemma 3 (Gemma Team et al.,
\protect\hyperlink{ref-gemma2025gemma3}{2025}), Qwen3 (Yang et al.,
\protect\hyperlink{ref-yang2025qwen3}{2025}), OLMo 2 (Team OLMo et al.,
\protect\hyperlink{ref-olmo2024olmo2}{2025}), OLMo 3 (Team Olmo et al.,
\protect\hyperlink{ref-olmo2025olmo3}{2026}), and Apertus
(Hernández-Cano et al., \protect\hyperlink{ref-apertus2025}{2026};
Project Apertus et al., \protect\hyperlink{ref-apertus2026v15}{2026}).
For each base model, we compared the pretrained model with its final
post-trained version. We included the OLMo and Apertus models because
they are open beyond their weights: Their training data are known, and
for these three base models the developers released checkpoints after
each stage of post-training, which allowed us to follow the change step
by step: OLMo 2 7B (base, supervised fine-tuning {[}SFT{]}, direct
preference optimization {[}DPO{]}, and a final reinforcement-learning
stage with verifiable rewards), Apertus 1.5 8B (base, SFT, reinforcement
learning, DPO, and a final alignment stage), and OLMo 3 7B, for which
two lines of post-training start from the same base model (one optimized
for explicit reasoning, ``Think'', one for conversation, ``Instruct'')
and a third applies reinforcement learning to the base model without any
SFT (``RL-Zero'').

\textbf{Table 1.} Models (Hugging Face identifiers).

\begin{longtable}[]{@{}lll@{}}
\toprule
\begin{minipage}[b]{0.30\columnwidth}\raggedright
Model\strut
\end{minipage} & \begin{minipage}[b]{0.30\columnwidth}\raggedright
Base\strut
\end{minipage} & \begin{minipage}[b]{0.30\columnwidth}\raggedright
Post-training checkpoints\strut
\end{minipage}\tabularnewline
\midrule
\endhead
\begin{minipage}[t]{0.30\columnwidth}\raggedright
Gemma 3 4B\strut
\end{minipage} & \begin{minipage}[t]{0.30\columnwidth}\raggedright
google/gemma-3-4b-pt\strut
\end{minipage} & \begin{minipage}[t]{0.30\columnwidth}\raggedright
google/gemma-3-4b-it\strut
\end{minipage}\tabularnewline
\begin{minipage}[t]{0.30\columnwidth}\raggedright
Gemma 3 12B\strut
\end{minipage} & \begin{minipage}[t]{0.30\columnwidth}\raggedright
google/gemma-3-12b-pt\strut
\end{minipage} & \begin{minipage}[t]{0.30\columnwidth}\raggedright
google/gemma-3-12b-it\strut
\end{minipage}\tabularnewline
\begin{minipage}[t]{0.30\columnwidth}\raggedright
Qwen3 8B\strut
\end{minipage} & \begin{minipage}[t]{0.30\columnwidth}\raggedright
Qwen/Qwen3-8B-Base\strut
\end{minipage} & \begin{minipage}[t]{0.30\columnwidth}\raggedright
Qwen/Qwen3-8B\strut
\end{minipage}\tabularnewline
\begin{minipage}[t]{0.30\columnwidth}\raggedright
OLMo 2 7B\strut
\end{minipage} & \begin{minipage}[t]{0.30\columnwidth}\raggedright
allenai/OLMo-2-1124-7B\strut
\end{minipage} & \begin{minipage}[t]{0.30\columnwidth}\raggedright
-SFT, -DPO, -Instruct\strut
\end{minipage}\tabularnewline
\begin{minipage}[t]{0.30\columnwidth}\raggedright
Apertus 1.5 8B\strut
\end{minipage} & \begin{minipage}[t]{0.30\columnwidth}\raggedright
swiss-ai/Apertus-v1.5-8B-base\strut
\end{minipage} & \begin{minipage}[t]{0.30\columnwidth}\raggedright
-SFT, -RL, -DPO, final (Apertus-v1.5-8B)\strut
\end{minipage}\tabularnewline
\begin{minipage}[t]{0.30\columnwidth}\raggedright
OLMo 3 7B Think\strut
\end{minipage} & \begin{minipage}[t]{0.30\columnwidth}\raggedright
allenai/Olmo-3-1025-7B\strut
\end{minipage} & \begin{minipage}[t]{0.30\columnwidth}\raggedright
Olmo-3-7B-Think-SFT, -Think-DPO, -Think\strut
\end{minipage}\tabularnewline
\begin{minipage}[t]{0.30\columnwidth}\raggedright
OLMo 3 7B Instruct\strut
\end{minipage} & \begin{minipage}[t]{0.30\columnwidth}\raggedright
(same base)\strut
\end{minipage} & \begin{minipage}[t]{0.30\columnwidth}\raggedright
Olmo-3-7B-Instruct-SFT, -Instruct-DPO, -Instruct\strut
\end{minipage}\tabularnewline
\begin{minipage}[t]{0.30\columnwidth}\raggedright
OLMo 3 7B RL-Zero\strut
\end{minipage} & \begin{minipage}[t]{0.30\columnwidth}\raggedright
(same base)\strut
\end{minipage} & \begin{minipage}[t]{0.30\columnwidth}\raggedright
Olmo-3-7B-RL-Zero-Mix\strut
\end{minipage}\tabularnewline
\bottomrule
\end{longtable}

\hypertarget{eliciting-authorship-judgments}{%
\subsection{2.3 Eliciting authorship
judgments}\label{eliciting-authorship-judgments}}

We asked every model whether a text had been written by a human or
generated by an AI, in German, and read the answer off the model's
next-word probabilities rather than from generated text. In the plain
completion format (``raw''), which works for base and post-trained
models alike, the prompt stated that the following text had been written
either by a human or by an AI, presented the text in quotation marks,
and asked: ``Wurde dieser Text von einem Menschen geschrieben oder von
einer KI generiert? Antworte mit einem Wort (Mensch oder KI).'' followed
by ``Antwort:''. Post-trained models were in addition queried with their
own chat template (``chat''), with the system message ``Du bewertest
Texte.'' and the same question as user message. For reasoning models
whose template opens a reasoning block, the block was closed empty, so
that the next word is the answer.

The decision variable was the difference in log probability between the
two answers, dv = log P(``KI'') − log P(``Mensch''). Each answer
probability summed over its spellings with and without a leading space
and included the probability that the word ends after the answer tokens
(Pimentel \& Meister,
\protect\hyperlink{ref-pimentel2024compute}{2024}). To remove position
effects, every text was judged with both orders of the answer options
(``Mensch oder KI'', ``KI oder Mensch''), and dv was averaged over the
two orders. As a manipulation check, we recorded the answer probability,
i.e., the summed probability of the two answers; conditions in which a
model placed less than 30\% of its probability on the two answers were
excluded from inferential comparisons.

\hypertarget{signal-detection-measures-and-statistics}{%
\subsection{2.4 Signal detection measures and
statistics}\label{signal-detection-measures-and-statistics}}

Treating AI texts as signal, we computed sensitivity as d′ = √2 ·
z(AUC), where AUC is the area under the ROC curve of dv. For this
transformation, the AUC is limited to the interval {[}1/(2m), 1 −
1/(2m){]}, where m is the size of the smaller class, so that d′ has a
maximum that depends on the number of texts (e.g., 3.64 for 100
statement pairs and 4.54 for 748); values at this maximum mean that all
or nearly all pairs were ordered correctly (any AUC at or above the
bound yields the maximum), and differences near it are compressed. This
measure does not depend on where the model sets its decision threshold,
and negative values indicate that the model judges AI texts as less
AI-like than human texts. As a measure of response bias, we computed the
criterion c at the model's natural threshold (dv = 0), with a log-linear
correction of hit and false-alarm rates (Hautus,
\protect\hyperlink{ref-hautus1995corrections}{1995}); positive values
indicate a bias toward answering ``Mensch''. Because dv is a graded
preference, d′ characterizes how the evidence for the answer is ordered
across texts, in the manner of a confidence-rating ROC, even where the
binary answer hardly varies; c describes where the order-averaged dv
places the threshold and is not the criterion of any single prompt,
since the binary answers often depended on the order of the answer
options (Section 3.1). To test whether signal detection theory describes
the models' judgments, we manipulated the response bias in two ways that
should leave the sensitivity unchanged: the order of the two answer
options, and a sentence stating the share of AI texts in the study (20,
50, or 80\%), added to the plain prompt. For each single prompt
condition, we computed d′ and c from the dv of that prompt alone and
estimated the changes between conditions with a grouped bootstrap; where
a model gave the same answer to at least 98\% of the texts, c is close
to its bound and can hardly move. To measure how strongly a model uses
the stated base rate, we computed for every text the shift of dv between
the prompts stating 20\% and 80\% AI texts and compared its mean with
the shift of the prior log-odds, ln(.8/.2) − ln(.2/.8) = ln 16 = 2.77
nats, by which an ideal observer would shift the posterior log-odds of
every text. The base-rate sentence named the share of AI texts first
(``In dieser Studie wurden etwa 20 \% der Texte von einer KI generiert
und etwa 80 \% von Menschen geschrieben.''); for Gemma 3 4B, OLMo 2 and
OLMo 3 Think, the manipulation was repeated with the human share named
first. Confidence intervals (95\%) were obtained by a grouped bootstrap
that resampled whole source groups (1,000--2,000 resamples). The AUC,
and hence d′, as well as hit and false-alarm rates are computed within
each class and therefore do not depend on the unequal numbers of AI and
human texts (chats: 239 vs.~61; newspaper segments: 65 vs.~38); the
class proportions affect only the precision of the estimates, which is
limited by the smaller class and by the number of independent groups.
For the same reason, the probes were fitted with balanced class weights.

To isolate the effect of post-training from the effect of the prompt
format, we compared each post-trained checkpoint with its base model on
the identical raw prompt (``clean'' contrasts), computing the difference
in d′ on the same texts with a paired grouped bootstrap. Comparisons
across formats (base raw versus post-trained chat) are reported but
marked as confounded. p-values were obtained from a z-test with the
bootstrap standard error and corrected for multiple comparisons with the
Holm procedure within the family of the 18 comparisons between final and
base model on the identical prompt, which address our hypothesis; as a
stricter check, we report the Holm correction across all tested
contrasts, and the Benjamini--Hochberg procedure as a secondary
correction (Appendix G). The summary inference rests on the model family
as the unit (Section 3.2).

\hypertarget{probes-and-baselines}{%
\subsection{2.5 Probes and baselines}\label{probes-and-baselines}}

To assess whether the authorship information is available in a model's
internal representation, we trained linear probes on the activations of
the residual stream at the last position of the prompt, i.e., where the
answer is formed, for every layer. Each probe consisted of
standardization, a principal component rotation, and an L2-regularized
logistic regression whose regularization strength was selected by an
inner grouped cross-validation (standardization and rotation were fitted
on the whole training fold, so that the inner validation folds entered
these unsupervised steps; the held-out fold of the outer
cross-validation entered none; a strictly nested version, in which all
steps were refitted within each inner split, changed the probe d′ for
the short chats by −0.03 on average, with a range from −0.20 to +0.12
across the base and final checkpoints of the six families). Performance
was estimated out of fold with a 10-fold cross-validation that held out
whole source groups, and expressed as d′ (AUC). As a check, the same
procedure was repeated with labels shuffled within source groups. For
comparisons across models, we report the probe at a fixed relative depth
of two thirds of the network.

To determine how much of this information is specific to the models, we
compared the probes with four baselines under the same cross-validation
scheme: logistic regressions on character n-grams (2--4) and on word
n-grams (1--2), both with TF-IDF weighting, on 14 surface statistics
(e.g., text length, word and sentence length, type-token ratio,
punctuation), and a probe on the average input embedding of the text
tokens (i.e., before any processing by the network). As an externally
trained reference, we applied EditLens (Thai et al.,
\protect\hyperlink{ref-thai2026editlens}{2025}), a detector based on
Llama 3.2 3B that was trained on English texts to estimate the extent of
AI editing; we re-implemented its published inference procedure and ran
it locally.

\hypertarget{in-context-examples}{%
\subsection{2.6 In-context examples}\label{in-context-examples}}

To test whether the models' judgments reflect a belief that labeled
examples can correct, we placed k = 2, 8, or 16 labeled examples (half
human, half AI) before each text, in the same format as the target
question and with the correct answer filled in. Examples were drawn from
source groups other than that of the target text (for the chats, from
other authors) and truncated to 200 tokens. We used four fixed example
sets per condition; because few-shot answers are sensitive to the label
of the last example, the last label alternated between example sets.

\hypertarget{causal-interventions}{%
\subsection{2.7 Causal interventions}\label{causal-interventions}}

We used three interventions, which are illustrated in Figure 2. All of
them change internal states or weights of the model during a
computation, which can be repeated at will with and without the
intervention.

\emph{Steering.} As a check on how the answer depends on the represented
information at the position where it is produced (Appendix D), we added
a vector to the residual stream at the last prompt position after a
given layer and measured the change in the answer score dv (Section
2.3). The authorship (or truth) direction was the difference between the
mean activations of the two classes, estimated for each text from other
source groups (7-fold cross-fitting). Steps were expressed in units of
the within-class standard deviation along this direction; the primary
test used ±2 SD, and we recorded a dose-response curve (±1, 2, 4, and 6
SD). The directional effect of a direction was half the difference
between the changes in dv for positive and negative steps. As null
controls, we used 50 random directions within the subspace spanned by
the 20 leading principal components of the within-class activations and
50 mean differences between random halves of the texts within each
class, each drawn once and applied to all texts. Controls took steps of
the same length as the authorship direction (norm-matched) or, in a
second run, of ±2 SD along their own direction (scale-matched). The
p-value of a direction was its rank among the absolute control effects
(two-sided; minimum 1/101). As a positive control, we used the direction
along which the model's own answers vary (the decision direction). We
also report, per cell, the share of shifted texts whose projection
stayed within the 1st--99th percentile of the observed texts, and the
effect of a shift by one class-mean distance, extrapolated from the ±1
SD step. To make this number of directions feasible, the prompt up to
the last position was processed once per text and its cached activations
were reused for all steering settings, which reproduced the scores of
the uncached computation in a check on three statements with all
steering settings (102 scores: r = .999, largest difference 0.06 nats).
Steering was run for 25 cells (model × corpus × layer; Table D1 in
Appendix D).

\emph{Model stitching.} Model stitching combines parts of different
networks (Bansal et al.,
\protect\hyperlink{ref-bansal2021stitching}{2021}; Lenc \& Vedaldi,
\protect\hyperlink{ref-lenc2015equivariance}{2015}). For the models with
stage-wise checkpoints, we combined the embedding and the first ℓ layers
of one checkpoint with the remaining layers, final normalization and
output layer of another, passing the residual stream at the junction on
unchanged. Stitching without an adapter is justified because both
checkpoints derive from the same pretrained weights; we verified this by
linear centered kernel alignment (CKA; Kornblith et al.,
\protect\hyperlink{ref-kornblith2019similarity}{2019}), a measure of how
similarly two sets of activations for the same inputs are arranged,
which is 1 for identical arrangements irrespective of a rotation of the
activation space, between the activations of base and post-trained
checkpoints, by the perplexity of each stitched model on 40 texts, and
by checking the two extremes: With ℓ = 0 (only the embedding from the
other checkpoint) and with ℓ = L (all layers from the other checkpoint,
but the own final normalization and output layer), d′ was within 0.05 of
the respective unstitched checkpoint in 15 of 16 cases (largest
difference 0.12). As a control task, we stitched the same checkpoints on
statements of diverse facts (counterfact), which base and SFT models
judged alike; for authorship, stitching and patching were run on the
textbook summaries and the short chats, and patching was repeated with a
prompt that placed the question before the text. We also exchanged only
the attention blocks or only the feed-forward blocks (each with its
normalization) of the first ℓ layers.

\emph{Activation patching.} Activation patching replaces internal states
of a model with those obtained under different conditions (Heimersheim
\& Nanda, \protect\hyperlink{ref-heimersheim2024patching}{2024}; Meng et
al., \protect\hyperlink{ref-meng2022rome}{2023}; Vig et al.,
\protect\hyperlink{ref-vig2020causal}{2020}); here, the states came from
a different checkpoint (cf.~Prakash et al.,
\protect\hyperlink{ref-prakash2024finetuning}{2024}). Stitching without
an adapter at layer ℓ is the special case in which the full residual
stream is exchanged at all positions. To separate the representation of
the text from the processing at the positions that follow it, we ran the
donor checkpoint over every prompt, stored its residual stream after
decoder layer k (i.e., the output of layer k, counted from 1) at all
prompt positions, and then ran the recipient checkpoint with its
residual stream after layer k replaced by the donor's at a selected set
of positions: the tokens of the text (or statement), the positions
before the text (the instruction), the positions after the text
(question, instruction to answer with one word, and answer cue), the
tokens of the question line only, the final position only, or all
positions. The effect of an exchange is reported as the d′ of the
patched recipient or as the share of the difference in d′ between the
two unpatched checkpoints that the exchange transferred. A share above
100\% or below 0\% means that the hybrid lies outside the range of the
two checkpoints; where a checkpoint is at the maximum d′, shares are
compressed and a small disruption cannot be seen.

We first ran the exchange at layers 8 and 16 (textbook summaries, short
chats, and the negated statements with 300 pairs), and for the textbook
summaries with a prompt that placed the question before the text. We
then repeated it across layers: for authorship on the textbook summaries
(all 238 texts; OLMo 2 and Apertus, base vs.~SFT) and for the negated
statements (first 100 pairs; OLMo 2 and Apertus, base vs.~SFT; Gemma 3
4B and 12B, base vs. final model), at 12 layers each (2 to 16 in steps
of 2, then 20, 24, 28, and 32 for the models with 32 layers; 12 layers
between 3 and 34 for Gemma 3 4B and between 4 and 48 for Gemma 3 12B).
For the negated statements, the default prompt poses the question before
the statement (``Is the following statement true or false? Statement:
\ldots{} Answer with one word (True or False). Answer:''); for OLMo 2
and Apertus we therefore also ran the answers and the patching across
layers with the statement placed first (``Statement: \ldots{} Is this
statement true or false? Answer with one word (True or False).
Answer:''). The runs across layers for OLMo 2 used NVIDIA V100 GPUs; in
half precision, their scores differed slightly from those obtained on
A100 GPUs (unpatched condition: largest difference 0.10 nats per item, r
\textgreater{} .998), so that each run across layers is compared only
with its own unpatched conditions.

\hypertarget{two-further-judgments}{%
\subsection{2.8 Two further judgments}\label{two-further-judgments}}

To test whether the results generalize beyond authorship, we applied the
same analyses to two further judgments on publicly available data, for
which the correct answer is known as well.

\emph{Truth judgments.} We used three statement sets from Marks \&
Tegmark (\protect\hyperlink{ref-marks2024geometry}{2024}), each
consisting of 748 matched pairs of a true and a false statement that
differ in a single fact: simple geographic statements (``The city of
Bokaro is in India.''), the same statements negated (``The city of
Bokaro is not in China.''), and statements on diverse facts
(counterfact; one true and one false statement with the same subject and
relation, 748 pairs drawn with a fixed seed). The models were asked ``Is
the following statement true or false?'' and answered with ``True'' or
``False''; dv was the log-odds of the two answers, computed and averaged
over both answer orders as for authorship. In the chat format, the
system prompt was ``You evaluate statements.'' The pair formed the unit
of resampling and cross-validation.

\emph{Deceptive reviews.} The Deceptive Opinion Spam Corpus (Ott et al.,
\protect\hyperlink{ref-ott2011finding}{2011},
\protect\hyperlink{ref-ott2013negative}{2013}) contains 1,600 reviews of
20 hotels in Chicago, half of them written by guests (truthful) and half
by crowd workers who had never stayed at the hotel (deceptive), each in
a positive and a negative version. Human readers judge these reviews
close to chance, whereas classifiers on word frequencies separate them
well (Ott et al., \protect\hyperlink{ref-ott2011finding}{2011}). The
prompt defined both classes (``A truthful review was written by a real
guest about an actual stay; a deceptive review was written by someone
who has never stayed at the hotel.''), and the models answered with
``Truthful'' or ``Deceptive''; the system prompt in the chat format was
``You evaluate hotel reviews.'' The hotel formed the unit of resampling
and cross-validation.

For both judgments, we used the same six base models as for authorship,
with all post-training stages for OLMo 2, Apertus 1.5 and OLMo 3 Think,
and the base and final models for Gemma 3 4B, Gemma 3 12B and Qwen3 8B.
Probes were logistic regressions (L2 penalty, C = 0.01) on the
activations at the last prompt position at two thirds of the network
depth, evaluated out of fold. For the negated statements, we
additionally applied a probe trained on the simple geographic statements
of the same checkpoint (transfer probe), which tests whether one truth
direction also covers negated statements (Marks \& Tegmark,
\protect\hyperlink{ref-marks2024geometry}{2024}). Model stitching and
activation patching were carried out as for authorship, for both further
judgments, but only where there was a change to locate: in models whose
answers changed substantially with post-training and whose base models
placed at least 30\% of their probability on the two answer words, the
criterion used for authorship (Table 2). For the negated statements,
these were OLMo 2 and Apertus (base vs.~SFT, 300 pairs); the Gemma 3
base models fell below this criterion (18\% and 25\%), and the base
models of Qwen3 and OLMo 3 already judged the negated statements
perfectly. Because the two models differed in where the repair appeared
(Section 3.6), we later added the patching across layers for the two
Gemma 3 models as an exception to this criterion; their results are
reported with this caveat. For the reviews, these were Gemma 3 4B, which
lost sensitivity, and Qwen3 8B, which gained (base vs.~final model; 800
pairs for stitching, 400 for patching); OLMo 2, Apertus and OLMo 3
changed little, and Gemma 3 12B, added later for consistency, was not
localized. For the statements on diverse facts, post-training changed
the answers too little in any model.

\emph{Probe transfer between checkpoints.} To test whether post-training
changes the geometry of the representation, we trained a probe at each
layer on the activations of one checkpoint and tested it on those of the
other (base vs.~final model, both directions; grouped five-fold
cross-validation over the same items). Each checkpoint was standardized
with the statistics of its own training folds, so that the test concerns
the direction that separates the classes. The transfer deficit is the
loss in d′ relative to the probe trained and tested on the same
checkpoint, averaged over both directions. For the authorship corpora,
the analysis was carried out both for the last prompt position and for
the mean over the text positions.

\emph{Lens model decomposition.} For the short chats and the reviews, we
decomposed the judgments with the lens model equation, using
interpretable text properties as cues (Appendix B).

\emph{Supervised fine-tuning data.} To examine whether the supervised
fine-tuning data contain valid examples of the three judgments, we
searched the SFT mixtures of OLMo 2 {[}Tülu 3; Lambert et al.
(\protect\hyperlink{ref-lambert2024tulu3}{2025}); all 939,344
conversations{]} and of Apertus 1.5 (a random sample of 373,520 out of
3,707,851 conversations) with regular expressions for requests to judge
the authorship of a text, to judge the authenticity of reviews or
statements, and to judge statements as true or false. Matching
conversations were inspected to determine whether they were actual
judgment tasks.

\hypertarget{training-study}{%
\subsection{2.9 Training study}\label{training-study}}

To test causally whether the training signal sets the direction of the
readout, we fine-tuned the OLMo 2 7B base model under five conditions
that differed only in a small set of added judgments. Every run
contained the same 20,000 conversations, sampled proportionally from all
parts of the model's original supervised fine-tuning mixture (Tülu 3;
Lambert et al., \protect\hyperlink{ref-lambert2024tulu3}{2025}). The
conditions added 1,994 authorship judgments in the format of our
experiments (plain prompt, random answer order) with correct labels
(C1), with labels that followed a formality stereotype (C2), or with
random labels (C3), or no judgments at all (C0). A further condition
(C4) added the same texts with an unrelated judgment in the same layout
(whether the text is about politics, business, or society; answers
``Ja''/``Nein''; labels from the newspaper section), so that it
practiced a binary judgment of a text without the authorship question.
The judged texts were German news articles that are disjoint from our
test corpora: human articles from the online edition of the Austrian
daily DER STANDARD (10kGNAD; Block,
\protect\hyperlink{ref-block2019gnad}{2019}; derived from Schabus et
al., \protect\hyperlink{ref-schabus2017onemillion}{2017}), shortened to
25, 80, or 150 words, and AI articles that Qwen3 8B and Gemma 3 12B
(instruction-tuned versions) wrote on the same topics and sections,
shortened to the same lengths; each topic contributed either its human
or its AI text. For C2, the more formal half of each true class,
according to a score combining mean word length, the share of long
words, and the share of capitalized nouns, was labeled ``KI'', so that
the labels were independent of the true authorship (50\% agreement) and
followed the stereotype alone. Fine-tuning used low-rank adapters
{[}LoRA; Hu et al. (\protect\hyperlink{ref-hu2022lora}{2021}); rank 16,
all attention and feed-forward projections, learning rate
10\textsuperscript{−4}, one epoch, three seeds per condition{]}, with
the loss on the assistant turns of the conversations and on the answer
word of the judgments. We evaluated the final adapters on our three
authorship corpora and on 400 held-out news items (plain prompt, both
answer orders), and we analyzed their representations with probes and
probe transfer as in Section 2.8. The official SFT checkpoint of OLMo 2
served as a reference.

To test whether the answers after training with valid news labels follow
what distinguishes AI from human news, we fitted a probe on the held-out
news in the base model (text positions, layers 20 and 24), applied it
unchanged to the short chats, and computed for every fine-tuned model
the Spearman correlation across the 300 chats between the probe score
and the answer score dv (pooled and within each class; means over the
three seeds; grouped bootstrap over source groups; differences between
conditions on the same resamples). The prediction was a positive
correlation for C1 that exceeds those of C0, C3, and C4. As a control
that uses no activations, we repeated the test with logistic regressions
on word n-grams (1--2) and character n-grams (2--4) of the same held-out
news, and computed partial rank correlations of each predictor with dv
controlling for the other. We also computed the cosine between the
direction that separates AI from human news at the last prompt position
of a fine-tuned model (difference of class means, or weight vector of a
probe) and the direction along which its answers to the chats vary
(difference between chats above and below the median dv, or ridge
regression of dv on the activations), with 1,000 random directions as a
reference; all variants are reported.

\hypertarget{implementation}{%
\subsection{2.10 Implementation}\label{implementation}}

All models were run in bfloat16 precision with the Hugging Face
\texttt{transformers} library (version 5.22.2; for Apertus, the
developers' fork of version 5.18) and PyTorch 2.6, on NVIDIA A100 GPUs
(40 GB) of the university's high-performance computing cluster and on a
local workstation. Gradients for the supplementary readout analysis were
computed with gradient checkpointing, which leaves them unchanged. All
analysis code is available in the project repository
(https://github.com/laubrock-lab/represented-but-misread).

\hypertarget{data-and-code-availability}{%
\subsection{2.11 Data and code
availability}\label{data-and-code-availability}}

The newspaper texts are unpublished and confidential, the human chat
messages were written by students who have not yet been asked for
consent to publication, and the human textbook summaries are protected
by copyright. Therefore, these texts cannot be shared. The two further
judgments (Section 2.8) use publicly available data (Marks \& Tegmark,
\protect\hyperlink{ref-marks2024geometry}{2024}; Ott et al.,
\protect\hyperlink{ref-ott2011finding}{2011},
\protect\hyperlink{ref-ott2013negative}{2013}). For the negated
statements, the repository contains one entry point that reruns the
analyses from the public statement sets and a check against the values
reported here; for the reviews, it contains the code but no such entry
point. The human texts of the training study (Section 2.9) are public
(Block, \protect\hyperlink{ref-block2019gnad}{2019}); the AI texts that
we generated for it are not yet released. All code and the tables of
derived results from which the reported numbers are taken are available
in the project repository. The per-text model outputs for the authorship
corpora are not included, because their identifiers refer to unpublished
texts.

\hypertarget{use-of-ai-tools}{%
\subsection{2.12 Use of AI tools}\label{use-of-ai-tools}}

AI tools assisted at all stages of this work. The analyses were
implemented and run with the assistance of AI coding agents (Claude Code
with Claude Opus and Claude Fable; additional analyses with GLM and
DeepSeek models). Claude also assisted in planning the analyses and in
drafting and revising the manuscript, and, in separate sessions,
reviewed successive drafts in the role of a referee. The author designed
the study, decided on all analyses and interpretations, and is
responsible for the content. The complete history of the analyses is
documented in the repository.

\hypertarget{results}{%
\section{3 Results}\label{results}}

We first report the authorship judgments, for which we used our own
corpora: whether the information is available and whether the answers
use it (Section 3.1), how post-training changed the answers (Section
3.2), how the models used a stated base rate (Section 3.3), and where in
the network the change acts (Sections 3.4 and 3.5). Because these
corpora cannot be shared, Section 3.6 reports a judgment on public data
that can be reproduced in full, the truth of negated statements, for
which post-training changed the answers in the opposite direction.
Section 3.7 reports a training study that tests what sets the direction
of the change. A further judgment, whether a hotel review is genuine,
gave less conclusive results and is reported in Appendix C. Unless
stated otherwise, sensitivity is expressed as d′ estimated from the area
under the ROC curve, with 95\% confidence intervals from a grouped
bootstrap.

\hypertarget{authorship-the-information-is-available-but-the-answers-do-not-use-it}{%
\subsection{3.1 Authorship: the information is available, but the
answers do not use
it}\label{authorship-the-information-is-available-but-the-answers-do-not-use-it}}

We used three corpora of German texts that each pair human-written texts
with AI-generated counterparts: short chat messages truncated to 25
words (300 texts from 61 human originals), segments of local newspaper
articles (103 segments from 14 articles), and textbook summaries (238
texts, 119 matched human--AI pairs; Rosin \& Laubrock, in preparation).
In all analyses, texts derived from the same human original form one
group, so that a model, a probe, or a bootstrap resample never profits
from recognizing a topic. Positive values of d′ indicate that AI texts
are judged as more AI-like than human texts, negative values indicate
the reverse.

Is the information that separates AI from human texts available to the
models at all? We first answered this question with linear probes, i.e.,
simple classifiers that are trained on a model's internal activations to
predict authorship and are evaluated on held-out texts. At a fixed depth
of two thirds of the network, probes separated AI from human texts well
in every model and corpus, with a median d′ of 1.35 for the short chats
(range across model conditions 0.95--1.80), 2.66 for the newspaper
segments (1.65--3.14), and 3.70 for the textbook summaries (2.36--3.73).
When the authorship labels were shuffled within groups, probe
performance dropped to chance (mean AUC = .50), which rules out that the
probes merely memorized the training texts.

However, this information is not specific to the internal machinery of
language models. Classifiers trained on simple word frequencies reached
the same or higher sensitivity than the probes (short chats: d′ = 1.73
{[}1.44, 2.06{]}; newspaper segments and textbook summaries: AUC ≥
.99\footnote{When nearly all pairs of texts are ordered correctly, the
  AUC-based d′ approaches a maximum that depends on the number of texts
  (d′ = 3.73 for the textbook summaries and 3.14 for the newspaper
  segments); for such near-perfect classifiers, we therefore report the
  AUC.}), and a probe on the average input embedding of the text, i.e.,
before any processing by the network, performed about as well as probes
on deep layers (short chats: d′ = 1.48). Holding out whole authors
instead of single originals did not change this picture (word
frequencies: d′ = 1.79). A detector that was trained by others to
recognize AI-edited English text (EditLens; Thai et al.,
\protect\hyperlink{ref-thai2026editlens}{2025}) transferred to our
German material in a graded manner, reaching an AUC of .99 for the
textbook summaries, .73 for the newspaper segments, and .57 for the
short chats (Figure 3). Thus, the cues that distinguish AI from human
texts are present and readily accessible, both in the texts themselves
and in the models' representations; they consist largely of ordinary
lexical and stylistic regularities. In terms of Brunswik's lens model,
the environment offers valid cues.

\begin{figure}
\centering
\includegraphics[width=1\textwidth,height=0.68\textheight]{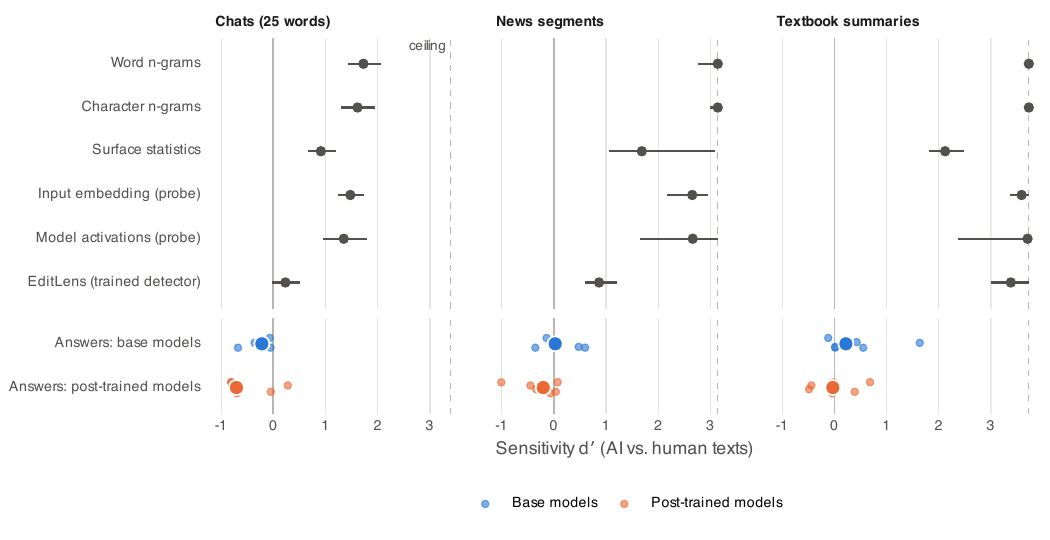}
\caption{The authorship information is available but not used.
Sensitivity (d′) for separating AI from human texts in the three
corpora. Upper rows: classifiers on word and character n-grams and on
surface statistics, probes on the average input embedding and on the
internal activations at two thirds of the network depth, and an
externally trained detector (EditLens). Lines show 95\% bootstrap
confidence intervals, except for the two probe rows, where they show the
range across models. Lower rows: sensitivity of the models' own answers
(raw prompt), one small dot per model, large dot = median. The dashed
line marks the maximum d′ that can be estimated with the given number of
texts.}
\end{figure}

These cues, however, are specific to the corpus. When a probe trained on
the newspaper segments was applied unchanged to the short chats, it
separated them with the wrong sign (Gemma 3 4B and 12B, six model
conditions: median d′ = −0.46 at the last prompt position and −0.48 for
the averaged text positions; Qwen3: −0.48 and −0.60), and the authorship
direction of the news pointed slightly against that of the chats (cosine
about −0.2). In the opposite direction, probes trained on the chats
transferred weakly but with the correct sign (median d′ = 0.37--0.87).
Hence, there is no single authorship direction that holds across text
types: Properties that mark AI texts in one corpus can mark human texts
in another. This is important for reading the answers below, because a
judge that applies one fixed policy to all texts can be above chance on
one corpus and below chance on another.

When the models were asked directly whether a text had been written by a
human or by an AI, their answers did not reflect this information. Table
2 summarizes the order-averaged sensitivity of the answers for the
pretrained base models and for the final post-trained models. Many cells
are close to chance, and a considerable number are clearly below it:
Qwen3-8B after post-training judged the newspaper segments with d′ =
−1.01 (raw prompt) and −1.30 (chat prompt), and Gemma 3 12B judged the
short chats with d′ = −0.81 and −1.09. Several models thus ordered the
texts inversely: What distinguished AI from human texts moved their
answers toward the wrong class. Note that these values were computed
from the graded preference of the model, i.e., the log-odds of ``AI''
versus ``human'' for each text, which plays the role of a confidence
rating in a rating-scale ROC analysis. The binary answer itself often
did not vary: In 23 of the 36 conditions with the plain prompt (Table 2,
base and final models), the model gave the same answer to more than 95\%
of the texts, and in several base models the share of ``AI'' answers
depended strongly on the order in which the two options were named
(e.g., OLMo 2 on the short chats: 89\% vs.~0\%). A negative d′ therefore
does not mean that the models gave many wrong binary answers. It means
that the graded evidence on which the answer is based was ordered the
wrong way round: human texts received a stronger preference for ``AI''
than AI texts did. Such an inverse ordering cannot result from a
response bias, which only shifts the preference for all texts alike;
properties typical of human texts pushed the answer toward ``AI'', and
properties typical of AI texts toward ``human''.

\textbf{Table 2.} Sensitivity (d′) of the models' answers. Base models
were queried with a plain completion prompt (``raw''); post-trained
models with the same prompt and with their chat template (``chat'').
Conditions in which a model placed less than 30\% of its probability on
the two answer words are marked.

\begin{longtable}[]{@{}lllll@{}}
\toprule
Base model & Corpus & Base (raw) & Final (raw) & Final
(chat)\tabularnewline
\midrule
\endhead
Gemma 3 4B & Chats & −0.21 & −0.72 & −0.66\tabularnewline
& News & +0.48 & −0.34 & −0.28\tabularnewline
& Textbook & +0.01 & −0.49 & −0.34\tabularnewline
Gemma 3 12B & Chats & −0.36 & −0.81 & −1.09\tabularnewline
& News & −0.02 & +0.07 & −0.01\tabularnewline
& Textbook & +1.64 & +0.68 & +0.27\tabularnewline
Qwen3 8B & Chats & −0.68 & −0.80 & −0.76\tabularnewline
& News & −0.36 & −1.01 & −1.30\tabularnewline
& Textbook & +0.55 & −0.02 & −0.27\tabularnewline
OLMo 2 7B & Chats & −0.07 & +0.28 & +0.03\tabularnewline
& News & −0.14 & −0.45 & −0.34\tabularnewline
& Textbook & −0.12 & −0.44 & −0.16\tabularnewline
Apertus 1.5 8B & Chats & −0.23 & −0.70 & −0.54\tabularnewline
& News & +0.07 & −0.07 & −0.26\tabularnewline
& Textbook & +0.43 & −0.04 & +0.03\tabularnewline
OLMo 3 7B (Think) & Chats & −0.05 & −0.05 & −0.06\tabularnewline
& News & +0.60 & +0.04 & −0.23\tabularnewline
& Textbook & +0.01 & (+0.39)ᵃ & −0.32\tabularnewline
\bottomrule
\end{longtable}

ᵃ Answer probability ≈ .30; interpret with caution.

The comparison between the columns of Table 2 already shows the pattern
that we test in the next section: On the identical plain prompt, the
post-trained model was less sensitive than its base model in 14 of the
18 comparisons.

\hypertarget{post-training-lowers-the-sensitivity-of-authorship-judgments}{%
\subsection{3.2 Post-training lowers the sensitivity of authorship
judgments}\label{post-training-lowers-the-sensitivity-of-authorship-judgments}}

Base models were not uniformly better, but several of them discriminated
clearly above chance where their post-trained counterparts did not
(e.g., Gemma 3 12B on the textbook summaries, d′ = +1.64; OLMo 3 on the
newspaper segments, +0.60; Qwen3 on the textbook summaries, +0.55). To
test the effect of post-training without confounding it with the prompt
format, we compared each final model with its base model on the
identical plain prompt. Of these 18 comparisons, 14 showed a decrease.
Because each model family contributes three corpora and each corpus six
families, the 18 comparisons are not independent. Our primary test
therefore treats the model family as the unit: A t-test over the six
family means yielded a mean change of Δd′ = −0.34 {[}−0.56, −0.11{]}
(t(5) = −3.82, p = .012); all six families showed a decrease on average,
two of them only slightly (OLMo 2: −0.10; OLMo 3 Think: −0.06). A
multilevel random-effects model with random effects for model family,
corpus, and cell gave the same estimate (−0.33 {[}−0.53, −0.13{]}); its
nominal test (t(17) = −3.54, p = .003) uses residual degrees of freedom
and is likely too liberal with six families. Heterogeneity between cells
was considerable (I² = 83\%). Restricted to the nine comparisons in
which both models placed at least half of their probability on the two
answer words, the estimate was similar but less precise (−0.31 {[}−0.64,
0.01{]}, p = .057; four families). In the three comparisons in which the
binary answers varied in both the base and the final model (Gemma 3 12B
on the short chats and the textbook summaries, Gemma 3 4B on the short
chats), post-training lowered d′ in all three cases (−0.45 to −0.95).
Per corpus, with six families each, the decrease was reliable for the
newspaper segments (−0.42 {[}−0.75, −0.08{]}, p = .024), but not for the
textbook summaries (−0.41 {[}−0.85, 0.04{]}, p = .066) or the short
chats (−0.19 {[}−0.55, 0.16{]}, p = .22).

Did the decline depend on the wording of the prompt? We repeated all
comparisons with a paraphrased instruction and question and with the
question placed before the text. Across the three prompts, the mean
change over the six family means (each the mean of nine comparisons) was
Δd′ = −0.29 {[}−0.54, −0.03{]} (t(5) = −2.88, p = .035), with a decrease
in five of six families and in 39 of the 54 comparisons. A multilevel
model with additional random effects for the prompt gave the same
estimate (−0.29), with variance components near zero for prompt and
corpus, but it is not informative as a test, because the prompt and the
corpus have only three levels each (t(2) = −2.81, p = .11). With the
paraphrased prompt alone, the mean change was −0.23 (13 of 18
comparisons negative; t(5) = −3.63, p = .015), and the changes of the
individual cells correlated with those for the original prompt (r =
.81). With the question placed before the text, the mean change was
similar (−0.30; 12 of 18 negative), but two families changed direction
(OLMo 3 Think: +0.38; Qwen3: +0.04), so that the family-level test was
not significant (p = .16; r = .63 with the original prompt). Note that
the wording moved the individual comparisons considerably: Between the
original prompt and the question-first prompt, Δd′ of a cell changed by
0.30 on average, compared with a mean absolute effect of post-training
of 0.43, and only 8 of the 18 cells showed a decrease with all three
prompts (5 of them reliably each time). The decline is therefore a
property of the average over models and prompts, not of every single
cell. The below-chance judgments of the short chats by Gemma 3 and Qwen3
persisted with all three prompts.

Among the individual comparisons, six decreases survived Holm correction
within this family of 18 tests, five of them on the textbook summaries
(Appendix G). Hence, post-training lowered the sensitivity of the
answers on average, but neither uniformly nor in a single direction
across models and text types.

For OLMo 2, Apertus 1.5, and OLMo 3, the developers released checkpoints
after each stage of post-training, which allowed us to ask when the
change occurs. On the textbook summaries, it occurred in OLMo 2 and
Apertus at the first, supervised stage (supervised fine-tuning, SFT), in
which the model learns to imitate example answers. For OLMo 2 on the
textbook summaries, d′ fell from −0.12 (base) to −0.50 after SFT and
then remained stable (−0.46 after preference optimization, −0.44 after
the final reinforcement-learning stage). For Apertus, d′ fell from +0.43
to −0.04 after SFT and stayed near zero through the subsequent stages
(0.00, −0.03, −0.04). The other corpora and the OLMo 3 lines, for which
stage-wise checkpoints exist as well, mostly gave the same picture: On
the newspaper segments, d′ fell at the supervised stage in OLMo 2 (−0.14
to −0.54), OLMo 3 Think (+0.60 to +0.08), and OLMo 3 Instruct (+0.60 to
−0.28), and on the short chats in Apertus (−0.23 to −0.73). There were
two exceptions: OLMo 3 Instruct on the short chats declined only at the
preference stage (−0.09 after SFT, −0.30 after DPO), and Apertus on the
newspaper segments did not change clearly at any stage (+0.07 to −0.07).
In most cases, the later, preference-based stages thus added little to a
change that was already complete after imitation learning.

Could the post-trained models simply hold a wrong belief about what AI
texts look like, a belief that a few correctly labeled examples would
correct? We tested this by placing 2, 8, or 16 labeled examples from
other sources before each text (four different sets of examples; d′
computed per set and averaged; Figure 4B). With 16 examples, d′ rose
reliably in 8 of 11 model conditions on the short chats and in 4 of 6 on
the textbook summaries, whereas two examples hardly helped. The gains of
base and post-trained models did not differ consistently: The difference
was reliable in two of five comparisons, with opposite signs (OLMo 2 on
the chats, where the base model gained more, +0.54 {[}0.27, 0.83{]};
Qwen3 on the textbook summaries, where the post-trained model gained
more, −0.31 {[}−0.55, −0.09{]}; Table E1). The gap between them
persisted: After 16 examples, the post-trained model was still less
sensitive than its base model in all five comparisons (e.g., Gemma 3 4B
on the short chats: +0.48 vs.~−0.27; Qwen3: +0.15 vs.~−0.17), and the
post-trained models of Gemma 3 4B and Qwen3 remained below chance on the
chats (Appendix E). Labeled examples do not give a model a new ability;
they only show which properties go with which label in the material at
hand. That they raised d′ in base and post-trained models alike
therefore indicates that part of the failure lies in which properties
the models take as signs of AI authorship, i.e., in a mapping that
information can correct, and not in an inability to tell the texts
apart. Because the examples did not remove the difference between base
and post-trained models, however, such a correctable mapping does not
account for the deficit that post-training introduced.

\begin{figure}
\centering
\includegraphics[width=1\textwidth,height=0.68\textheight]{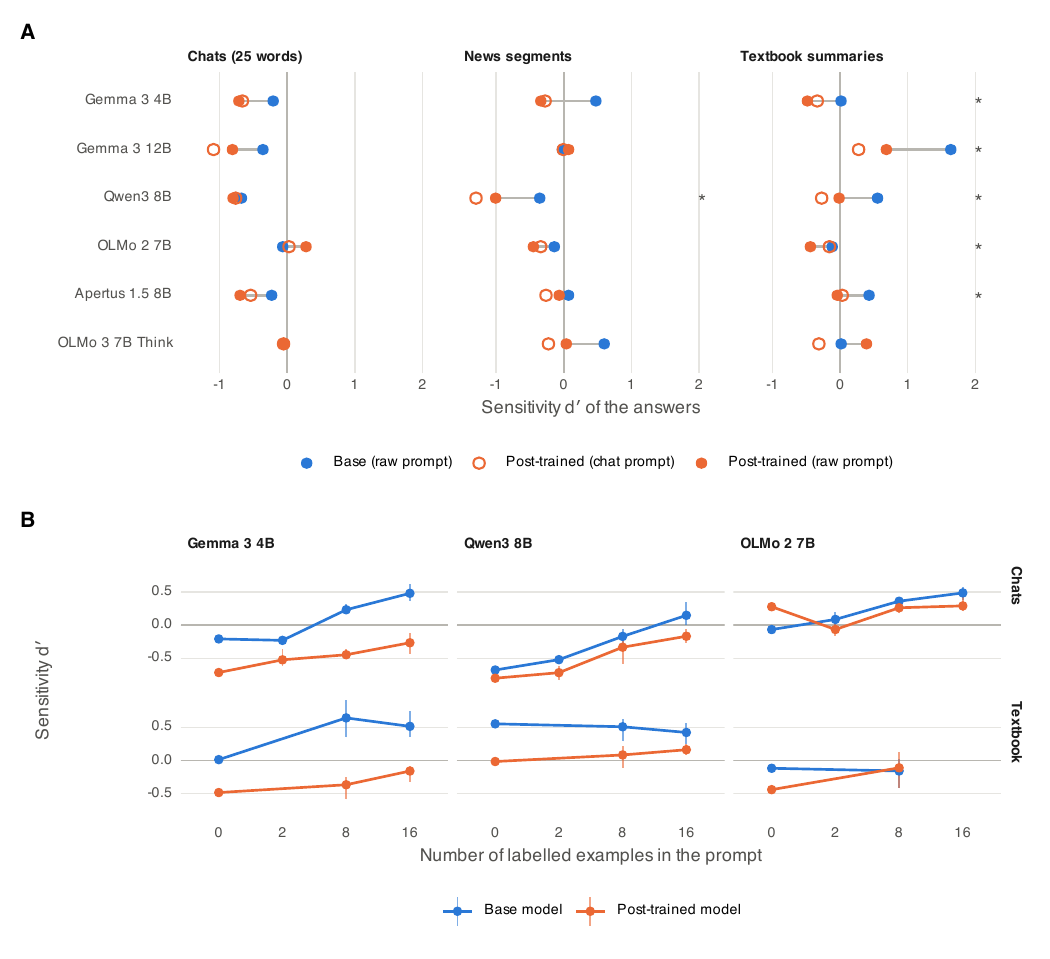}
\caption{Post-training mostly lowers the sensitivity of the answers;
with 16 labelled examples, most models improve. (A) Sensitivity of the
base model (blue) and of the final post-trained model with the identical
raw prompt (filled orange) and with its chat template (open orange), per
model and corpus. Asterisks mark differences between the base and the
final model on the identical raw prompt that survive Holm correction
within the family of 18 final-vs-base comparisons; all of them are
decreases. The increases of OLMo 2 on the short chats and of OLMo 3
Think on the textbook summaries are significant only without correction
or under false-discovery-rate control (the latter in a condition with
low answer probability, ≈ .30). Multilevel random-effects summary across
all 18 comparisons: Δd′ = −0.33 {[}−0.53, −0.13{]}. (B) Sensitivity as a
function of the number of labelled examples placed before each text (raw
prompt; base vs.~final post-trained model). Points show the mean of d′
computed separately for each of four example sets; vertical lines show
their range. Whether base models profit more than post-trained models
(difference of gains, 0 vs.~16 examples) is tested in the supplement:
the base model gained more in four of five comparisons, reliably only
for OLMo 2 on the chats; for Qwen3 on the textbook summaries, the
post-trained model gained more. On the textbook summaries, only 0, 8,
and 16 examples were tested; for OLMo 2 only 0 and 8, because 16
textbook summaries exceed its context window of 4,096 tokens.}
\end{figure}

Which properties of the texts did the answers follow? A lens-model
analysis of the short chats (Appendix B) showed that the answers of most
post-trained models weighted 34 text properties in the direction
opposite to their validity (cross-validated matching index G: Gemma 3 4B
−.49, Gemma 3 12B −.58, Qwen3 −.61, Apertus after SFT −.57). They took
long words, adjectives, nouns, and low readability as signs of AI
authorship, whereas the AI messages in this corpus were more verbal and
lexically more diverse than the human ones. In Qwen3, this weighting was
already present in the base model (−.53); OLMo 2, whose answers on the
chats improved slightly, weighted the properties in the opposite way
after SFT (+.37). The properties that the students had named as typical
of AI text (e.g., dashes, perfect capitalization) were barely valid in
the truncated messages and barely used by the models, and style-matched
messages that had not been revised according to the students' feedback
were judged like the revised ones (change in AUC between −.03 and +.04
for every model).

\hypertarget{criterion-sensitivity-and-the-use-of-a-stated-base-rate}{%
\subsection{3.3 Criterion, sensitivity, and the use of a stated base
rate}\label{criterion-sensitivity-and-the-use-of-a-stated-base-rate}}

Is signal detection theory an appropriate description of these
judgments? Its central assumption is that manipulations of the response
bias move the criterion but leave the sensitivity unchanged. This held
approximately. Raising the stated share of AI texts from 20\% to 80\%
moved the criterion in 48 of 72 comparisons (model × checkpoint × corpus
× answer order), in all of them toward ``AI'' and reliably so in 42
(mean \textbar Δc\textbar{} across all comparisons = 1.19); in the
remaining 24 comparisons, c did not change, because the model gave the
same answer to nearly every text at either share. The sensitivity was
not strictly invariant, however: d′ changed by 0.14 on average, without
a consistent direction, and the bootstrap interval of the change
excluded zero in 28 of the 72 comparisons. In the 14 comparisons in
which no condition was saturated, the criterion shifted reliably in all
14 and d′ in 3 (mean \textbar Δd′\textbar{} = 0.10). The order of the
answer options and the mere presence of the base-rate sentence also
acted mainly on the criterion (mean \textbar Δc\textbar{} = 0.78 and
0.42), and changed d′ by similar amounts (0.16 and 0.13, again without a
consistent direction; Appendix F). Thus, manipulations of the response
bias acted mainly on the criterion where the criterion could move, and
the criterion is a property of the model together with the prompt rather
than of the model alone. Because the accompanying changes of d′ were not
negligible relative to the effects reported below, we average d′ over
both answer orders and compare checkpoints only on identical prompts.

How strongly did the models use the stated base rate? Any observer who
takes the stated share into account should answer ``AI'' more readily
when it is higher. A reference point for the size of this effect is an
ideal observer in a stationary environment, for whom the stated share is
the prior and changes nothing else: The log-odds of the answer then
shift by the same amount for every text, ln 16 = 2.77 nats between 20\%
and 80\%. We use this value as a reference, not as a norm that a
sensible judge must meet (in a changing environment, for instance,
weighting a stated base rate less can be adequate), and it applies to dv
only insofar as dv is read as the model's posterior log-odds. The base
models shifted far less (mean shift of dv per model between 0.18 and
1.40 nats, i.e., 6--51\% of the benchmark), and the shift was nearly
uniform across texts (between-text SD within a condition 0.08--0.33).
After post-training, five of the six models shifted more than the
benchmark (3.67--10.93 nats, i.e., 1.3--3.9 times the benchmark), and
the shift differed considerably between texts (SD up to 3.14); OLMo 3
Think was the exception (1.21 nats, as in its base model, 1.17). These
values were obtained with a base-rate sentence that named the share of
AI texts first. For the three families that fit the available GPUs, we
repeated the manipulation with the human share named first. The pattern
was the same, but the size of the shift depended on the wording in the
post-trained models: The base models again shifted too little (Gemma 3
4B: 0.16; OLMo 2: 1.35; OLMo 3: 1.07 nats), the post-trained models of
Gemma 3 4B and OLMo 2 again shifted more than the benchmark, but less
than with the first wording (5.61 vs.~10.65 and 4.37 vs.~5.39 nats), and
OLMo 3 Think again stayed below it (1.48). Post-training thus changed
the use of a stated base rate from underweighting to overweighting in
five of six model families, by a factor that depended on the model and
on the wording (Table F2); the second wording was not run for Apertus,
Qwen3, and Gemma 3 12B.

Note that a shift in nats also depends on the overall scale of a model's
log-odds, and this scale differed between checkpoints: Across texts of
the same class, dv varied far more in the post-trained models
(within-class SD per model 0.61--2.89 nats) than in the base models
(0.21--0.53). Relative to this spread, i.e., to the differences that the
models make between texts, the stated base rate already moved the base
models considerably (by 0.9--3.3 SD per model) and the post-trained
models somewhat more (2.0--5.3 SD); the relative shift increased with
post-training in five families and decreased in OLMo 3 Think (3.3 to 2.0
SD), and with the second wording it increased only in Gemma 3 4B (0.8 to
4.2 SD), not in OLMo 2 (2.5 to 2.4 SD). Thus, the post-trained models
gave more extreme log-odds in general, to the base rate as well as to
the texts; that the base rate gained weight relative to the texts holds
for most, but not all, models and wordings.

\hypertarget{where-the-change-acts-model-stitching-and-activation-patching}{%
\subsection{3.4 Where the change acts: model stitching and activation
patching}\label{where-the-change-acts-model-stitching-and-activation-patching}}

\hypertarget{model-stitching}{%
\subsubsection{Model stitching}\label{model-stitching}}

Where in the network does post-training make the difference? Because the
base and the post-trained checkpoints of a model share the same
architecture and start from the same weights, we could combine them: the
lower layers from one checkpoint and the upper layers from the other,
passing the internal state at the junction on unchanged (``model
stitching''). This test requires no probe and no statistic derived from
the answers; it only measures how well the combined model discriminates.
Two observations justified the procedure. First, the representations of
the texts themselves were highly similar in the base and the
post-trained checkpoints of the three models with stage-wise checkpoints
(linear CKA across layers 0.97--0.999 for OLMo 2, Apertus and OLMo 3;
textbook summaries), whereas the representations at the position where
the answer is formed differed considerably (0.44--0.99, median
0.65--0.85). Note that linear CKA without bias correction, computed here
on 238 texts in about 4,000 dimensions, is biased upward, so that values
near 1 are weak evidence of identity; the contrast between the text
positions and the answer position is the more informative result. Note
also that this was less so for the other families: In Gemma 3, even the
text representations changed noticeably, most at mid-depth (0.42--0.99,
median 0.80--0.83), and the answer position differed strongly (median
0.20--0.31), so that for Gemma 3 the claim that the text positions
changed little rests on the probe transfer alone (Section 3.5). Second,
every stitched model produced fluent text (perplexity 6.0--6.7 for
Apertus, 8.2--9.4 for OLMo 2), and, as a check of the implementation,
stitching at the very bottom or the very top reproduced the respective
parent model.

If post-training changed only the final readout stage, the upper layers
should decide the outcome. This was not the case (Figure 5A). When the
first 8 of 32 layers of Apertus came from the base model and all layers
above from the SFT model, the combined model discriminated at least as
well as the base model (d′ = +0.59 {[}0.41, 0.80{]}; base +0.43).
Conversely, the first 8 layers of the SFT model sufficed to abolish the
base model's discrimination, although all 24 layers above came from the
base model (−0.50 {[}−0.74, −0.25{]}). For OLMo 2, the first 16 base
layers restored base-level performance under SFT upper layers (−0.01
{[}−0.15, 0.14{]}; base −0.12, SFT −0.50), and the first 6 SFT layers
were sufficient to produce the full deficit (−0.48). Finer steps showed
that the change builds up gradually over several early layers rather
than in a single one (Apertus: layers 3--8; OLMo 2: layers 1--14).

\begin{figure}
\centering
\includegraphics[width=1\textwidth,height=0.68\textheight]{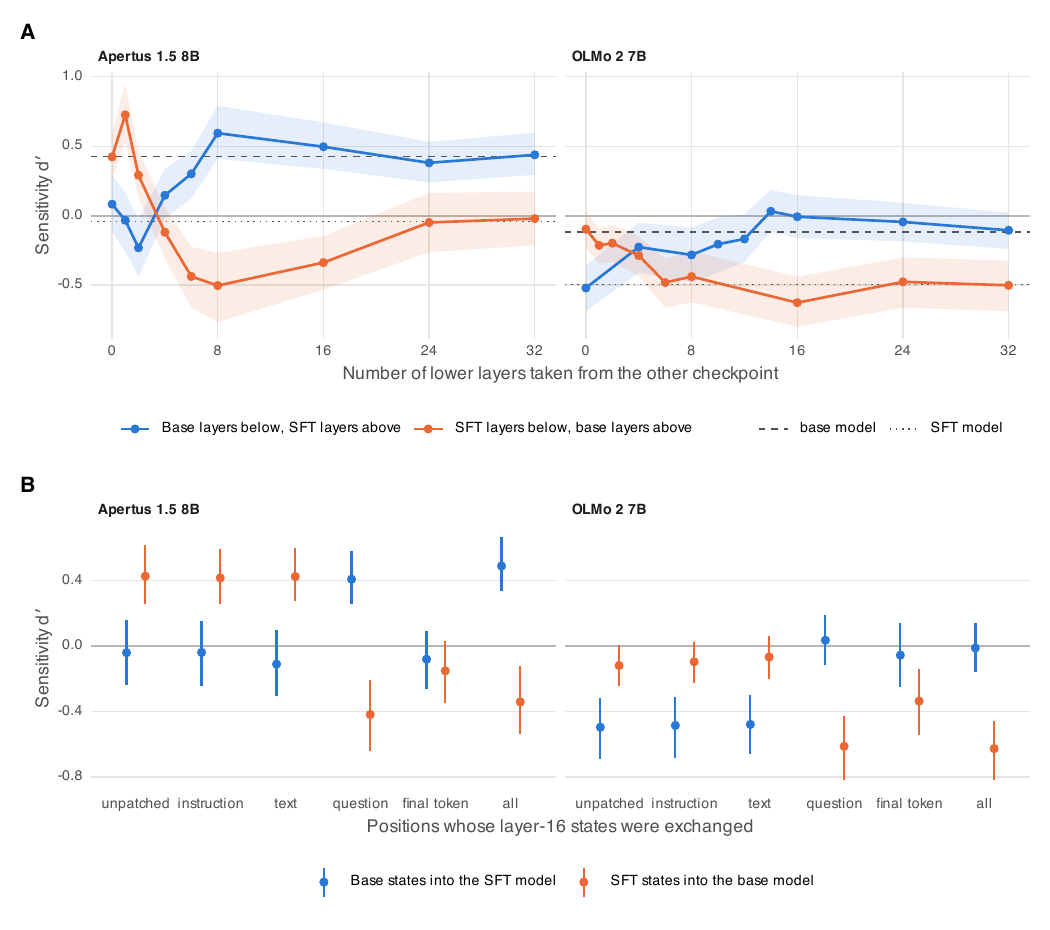}
\caption{Authorship: model stitching and activation patching at layer
16. (A) Model stitching between base and SFT checkpoints (textbook
summaries; the numbers of layers that were tested differ between the two
directions and between the models). The x-axis gives the number of lower
layers taken from the other checkpoint. Blue: base layers below, SFT
layers above; orange: SFT layers below, base layers above. Dashed and
dotted lines mark the sensitivity of the unmodified base and SFT models;
bands are 95\% bootstrap confidence intervals. (B) Activation patching
at layer 16: sensitivity after exchanging the internal states at the
indicated positions. Blue: base model's states inserted into the SFT
model; orange: SFT model's states inserted into the base model.
``Unpatched'' shows the recipient model without any exchange. Error bars
are 95\% bootstrap confidence intervals. Patching at layer 8 (more
mixed; text positions contribute somewhat) is shown in Supplementary
Figure S1.}
\end{figure}

Note, however, that the hybrids did not always lie between their
parents. In Apertus, the first 8 SFT layers under base upper layers
judged worse than the SFT model itself (−0.50 vs.~−0.04), with a reduced
probability on the two answer words (.39 vs.~.68 for the base and .83
for the SFT model), and the first SFT layer alone under base upper
layers judged better than either parent (+0.73). Were these effects
artifacts of stitching as such? As a control, we stitched the same
checkpoints on a judgment that post-training had changed little, the
truth of statements on diverse facts (counterfact; Apertus: d′ = 1.37
base vs.~1.60 SFT; OLMo 2: 1.78 vs.~1.74, on the 300 pairs used for
stitching). There, all hybrids stayed within or close to the range of
their parents (largest deviation from the unstitched parents 0.02 for
Apertus and 0.08 for OLMo 2; none of 14 Apertus hybrids and one of 14
OLMo 2 hybrids reliably outside), whereas on the textbook summaries 6 of
18 Apertus hybrids and 4 of 18 OLMo 2 hybrids lay reliably outside
(largest deviation 0.46 and 0.15). Stitching thus does not disturb the
models' judgments in general. The control does not show, however, that
the overshoot is specific to the judgment that post-training changed:
The control judgment was far from chance, with high answer probability,
whereas the textbook judgments were close to chance, where small
perturbations can reorder the texts. A judgment near chance that
post-training left unchanged would be the appropriate control, which we
did not run. The stitching results show that the early-layer weights of
a checkpoint suffice to carry the difference into an otherwise foreign
network; they do not by themselves show that the difference is computed
there.

On the short chats, on which post-training lowered the sensitivity of
Apertus (−0.23 to −0.73) and raised that of OLMo 2 (−0.07 to +0.38), the
early locus replicated in one direction only. The first 8 base layers of
Apertus restored the base level under SFT upper layers (−0.27), and the
first 4 base layers of OLMo 2 removed most of its gain (+0.10). The
first 8 SFT layers under base upper layers, however, did not produce the
effect of SFT in either model (Apertus: −0.31; OLMo 2: −0.07); it
appeared only with 16 SFT layers (−0.70 and +0.27). On the chats, the
early base layers were thus sufficient to remove the effect of SFT, but
the early SFT layers were not sufficient to produce it, in contrast to
the textbook summaries. No hybrid lay outside the range of its parents
on the chats. Note also that the effect to be localized is small for
OLMo 2 on this corpus.

Both sublayer types, attention and feed-forward blocks, contributed to
the effect of the early layers (Appendix H).

\hypertarget{activation-patching-at-layers-8-and-16}{%
\subsubsection{Activation patching at layers 8 and
16}\label{activation-patching-at-layers-8-and-16}}

The stitching results implicate the early layers, but do not yet say
where in the prompt the change acts. In a language model, information
about the text reaches the answer through the positions that follow it:
the question ``Was this text written by a human or generated by an
AI?'', the answer cue, and the last position, from which the answer is
produced and which itself reads the text at every layer. We therefore
exchanged the internal states between base and SFT checkpoints
selectively at different positions of the prompt (``activation
patching''): at the tokens of the text, at the instruction before the
text, at the question after the text, or at the final position only.

At layer 16, the pattern was unambiguous and identical for both models
(Figure 5B, which shows the positions exchanged at this layer).
Exchanging the states of the text tokens changed nothing: Apertus after
SFT remained at chance when it received the base model's text
representation (−0.11; SFT alone −0.04), and the base model kept its
sensitivity when it received the SFT model's text representation (+0.43;
base alone +0.43). Exchanging only the states of the question positions
transferred the full difference in both directions: the base model's
question states restored discrimination in the SFT model (Apertus:
+0.41; OLMo 2: +0.04, from −0.50), and the SFT model's question states
abolished it in the base model (Apertus: −0.42; OLMo 2: −0.61, from
−0.12). The final position alone carried part of this effect in Apertus
(−0.08 and −0.15) and in OLMo 2 most of it in one direction (−0.06 from
−0.50) and about half in the other (−0.34 from −0.12). The same held for
the short chats, on which exchanging the text states changed nothing
(all changes ≤ 0.04), whereas the states after the text transferred the
difference: The SFT states made the base models judge like the SFT
models (Apertus: −0.23 → −0.75, SFT −0.73; OLMo 2: −0.07 → +0.34, SFT
+0.38), and the base states restored the base level in Apertus (−0.73 →
−0.27) and partly in OLMo 2 (+0.38 → +0.20). Within the positions after
the text, the tokens of the question itself carried little and the
answer cue most of the difference (e.g., Apertus base with SFT states at
the question tokens: −0.46; at all positions after the text: −0.75); the
final position alone carried between a third and two thirds. Because
post-training improved the judgments of OLMo 2 on the chats, the change
resided at the positions after the text regardless of its direction. At
layer 8, the picture was mixed (Figure S1, which adds this layer): in
Apertus, the text positions still had a partial effect, and in OLMo 2
even exchanging all positions transferred only part of the difference,
consistent with the stitching result that in OLMo 2 the change
accumulates up to about layer 14. Thus, at the middle of the network,
exchanging the states at the positions after the text (question and
answer cue) transferred the difference, and exchanging the text states
did not.

\hypertarget{activation-patching-across-layers}{%
\subsubsection{Activation patching across
layers}\label{activation-patching-across-layers}}

Patching at two layers can miss a contribution that is confined to other
layers, as the negated statements show (Section 3.6). We therefore
repeated the patching on the textbook summaries at 12 layers between 2
and 32 (Figure 6). In OLMo 2, the text states did not transfer the
difference at any layer, in either direction: The base model with the
SFT model's text states stayed between −0.17 and −0.05 (unpatched
−0.12), and the SFT model with the base model's text states between
−0.61 and −0.44 (unpatched −0.50). The states after the text transferred
the difference from layer 12 onward (base model with SFT states: −0.50
at layer 12 and −0.59 at layer 14; SFT model with base states: −0.02 at
layer 14). Below layer 12, the difference was distributed: Exchanging
all positions had the clearest effect (e.g., −0.48 for the base model at
layer 6), and the positions of the instruction before the text
transferred a reliable part by themselves (34--48\% at layers 6 to 10
for the base model with SFT states; 32--51\% at layers 8 to 12 in the
other direction), about as much as the positions after the text at these
layers. In Apertus, in contrast, the text states carried part of the
difference in the lower half of the network: With the SFT model's text
states, the base model fell from 0.43 to 0.21--0.33 at layers 2 to 12
(22--46\% of the difference between the checkpoints), and with the base
model's text states, the SFT model rose from −0.04 to 0.10--0.26 at
layers 4 to 12 (30--64\%); the bootstrap interval of the transferred
share excluded zero at layers 6 to 12 in the first direction and at
layers 4, 8, 10, and 12 in the second. From layer 14 onward, the text
states had no effect in either direction. The instruction positions
again carried a part in the lower layers (41--47\% at layers 4 to 8 for
the SFT model with base states; inconsistent in sign in the other
direction). The SFT model's states after the text lowered the base model
from layer 4 onward, by more than the whole difference between the
checkpoints (e.g., to −0.48 at layer 12, compared with −0.04 for the SFT
model), which again shows that hybrids produce effects of their own; in
the other direction, the base model's states after the text restored the
SFT model partly at layers 6 to 12 (0.09--0.31), fully at layers 14 to
20 (0.41--0.61), and again only partly above (0.24--0.31 at layers 24 to
32, where exchanging all positions gave hardly more, 0.24--0.37; the
final normalization and output layer of the two checkpoints differ).
Note that the instruction before the text already names the two
alternatives (human or AI), so that the text is read with the judgment
in view, although before the question; a contribution of the text or
instruction states can therefore be specific to the judgment, and our
data do not decide whether the Apertus text-state effect is specific to
it or reflects a general change in how the lower layers encode text.
Thus, from layer 14 onward the difference between the checkpoints was
carried by the positions after the text in both models; below, the text
states contributed in Apertus but not in OLMo 2, and the instruction
positions contributed in both.

\begin{figure}
\centering
\includegraphics[width=1\textwidth,height=0.68\textheight]{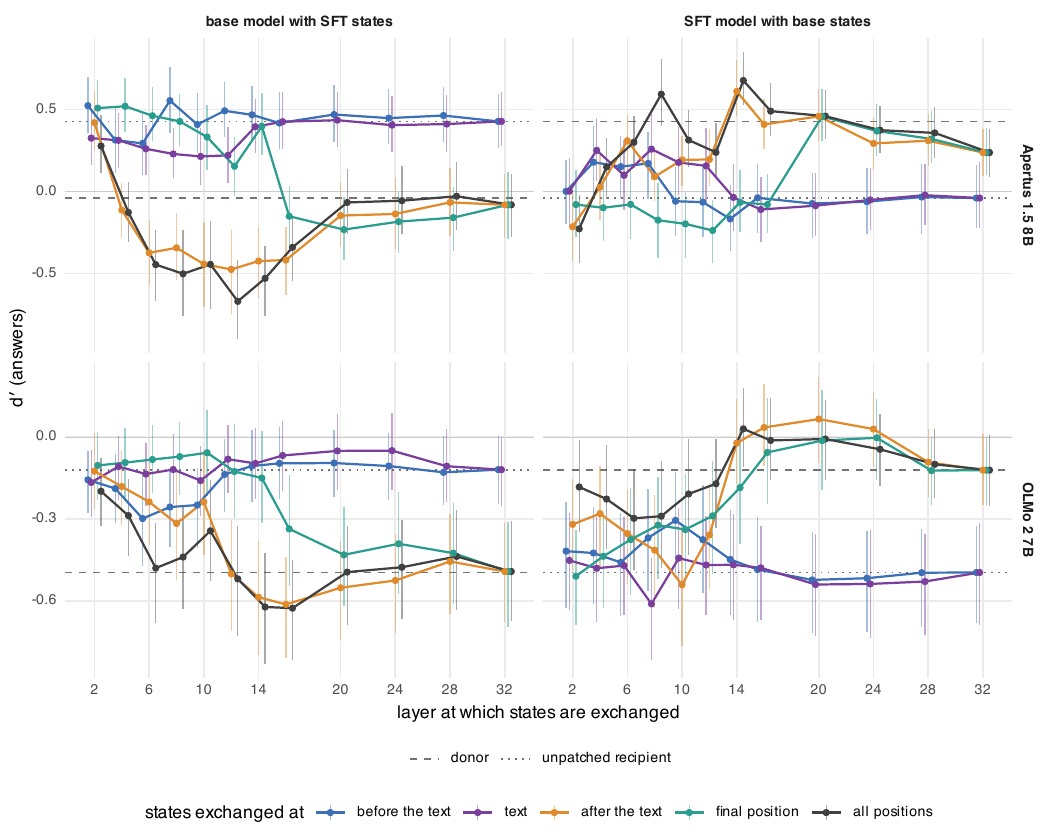}
\caption{Authorship: activation patching across layers (textbook
summaries, 238 texts; error bars: 95\% grouped-bootstrap confidence
intervals). Sensitivity (d′) of the recipient checkpoint after its
residual stream was replaced by the donor's at one set of positions and
one layer: the instruction before the text, the tokens of the text, the
question and answer cue after the text, the final position only, or all
positions. Note that the two rows have different scales. Horizontal
lines show the unpatched recipient (dotted) and the donor (dashed). In
OLMo 2, the text states had no effect at any layer; in Apertus, they
carried part of the difference between the checkpoints up to layer 12.
The instruction positions carried part of it in the lower layers of both
models. From layer 14 onward, the states after the text carried it in
both models. Hybrids can lie outside the range of the two checkpoints.}
\end{figure}

\hypertarget{question-placed-before-the-text}{%
\subsubsection{Question placed before the
text}\label{question-placed-before-the-text}}

In our default prompt, the question follows the text (the instruction
before the text names the two alternatives, but does not pose the
question). We also repeated the patching at layer 16 with the question
placed before the text, so that the text was read with the question
known. The decline persisted with this prompt (Apertus: +0.36 base
vs.~−0.19 SFT; OLMo 2: +0.09 vs.~−0.42). Table H1 (Appendix H) shows the
change of d′ for each set of positions (layer 16 only). Exchanging the
text states changed d′ by at most 0.12 (3--22\% of the difference
between the checkpoints). The few positions after the text
(``Antwort:'') transferred the whole difference in one of the four cases
(Apertus SFT with base states: −0.19 → +0.44, more than the difference
itself) and 42--49\% in the other three; the question before the text
transferred 18--53\% in three cases and moved the Apertus base model
away from the SFT model (+0.36 → +0.55). The effects of the single sets
did not add up to the effect of exchanging all positions, so that the
states before, within, and after the text contributed jointly. Note also
that a missing effect of the text states at layer 16 does not show that
they are irrelevant, because the later positions have by then read the
recipient's own text states in the layers below; with the default prompt
at layer 8, the text states did have an effect in Apertus. We therefore
conclude only that, at mid-depth, the difference between the checkpoints
was not carried by the text states alone, also when the text is read
with the question known. In summary, exchanging the text states at
mid-depth transferred little of the change, whereas exchanging the
states at the positions that follow the text transferred much or all of
it with the default prompt; in this sense, post-training changed how the
question takes up the text.

\hypertarget{probes-across-checkpoints}{%
\subsection{3.5 Probes across
checkpoints}\label{probes-across-checkpoints}}

If post-training leaves the representation intact and changes how it is
read out, a probe trained on the base model should work on the
post-trained model at the text positions, but less so at the position
where the answer is formed. This was the case wherever we could test it
(Figure S2), with qualifications. The comparison of the two sites exists
only for the authorship corpora (all six families); for the other
judgments, the activations were stored at the last position only, for
OLMo 2 and Apertus (all three sets) and for Gemma 3 4B and Qwen3
(reviews). At the text positions of the authorship corpora, probes
transferred between the base and the final checkpoints with little loss:
Averaged across models, the transfer deficit stayed at or below 0.15 d′
at every depth, and 94\% of the single model × layer values were at or
below 0.10 (largest value 0.54, for OLMo 2 on the textbook summaries in
the top layers). Because the probes at the text positions were at their
maximum in 89\% of these values for the textbook summaries and in 67\%
for the newspaper segments, the test can only fail where they are not,
i.e., mainly for the short chats (within-checkpoint d′ = 1.30--2.48);
there, the largest deficit of any model was 0.16. At the last prompt
position, the deficit grew with depth, reaching on average 0.7 to 1.9 d′
in the upper third of the network for the authorship corpora and 0.5 to
1.1 d′ for the other judgments (with one exception: for OLMo 2 on the
negated statements, the upper layers showed no deficit). Probes still
transferred partially at this position (mean d′ in the upper third up to
2.0, averaged over both directions of transfer, but near zero for Gemma
3 4B on the newspaper segments), so the readout position was reorganized
rather than replaced. Probes that were trained and tested on the same
model version (base or post-trained) continued to decode the information
at the last position, but in four of the six families less well after
post-training on at least one corpus (upper third of the network, final
vs.~base model: Gemma 3 4B, −0.52 on the newspaper segments and −0.90 on
the textbook summaries; Gemma 3 12B, −0.57 and −0.47; OLMo 3 Think,
−0.41 and −0.70; Qwen3, −0.38 on the textbook summaries). In Apertus,
the change was mixed (chats −0.26, newspaper segments +0.43), and in
OLMo 2 there was no decline. Post-training thus left the text positions
essentially unchanged where the test is informative, and both
reorganized and, in some models, degraded the information available at
the answer position.

In summary, for authorship the information was represented, the probes
at the text positions changed little with post-training, and from the
middle of the network onward the states at the positions after the text
transferred the change between checkpoints. In the lower layers, the
text states contributed to the change in Apertus, but not in OLMo 2.

\hypertarget{a-reproducible-case-with-the-opposite-direction-negated-statements}{%
\subsection{3.6 A reproducible case with the opposite direction: negated
statements}\label{a-reproducible-case-with-the-opposite-direction-negated-statements}}

The authorship corpora cannot be shared. Is the dissociation between
representation and answer, and its modification by post-training,
specific to them? We repeated the central analyses with truth judgments
on public statement sets (Section 2.8), whose main analyses (answers,
probes, stitching, and patching at two layers) can be rerun with one
script. Here, post-training acted in the opposite direction: It repaired
answers that the base models gave poorly. Table 3 summarizes the
results.

\textbf{Table 3.} Truth judgments: sensitivity (d′) of the probes (last
prompt position, two thirds of the network depth) and of the answers
(plain prompt) for statements on diverse facts (counterfact) and for
negated geographic statements. SFT: answers after supervised
fine-tuning; ---: no intermediate checkpoint released. The maximum d′
(AUC = 1) is 4.54 for the 748 statement pairs.

\begin{longtable}[]{@{}llrrrrr@{}}
\toprule
Judgment & Model & Probe base & Probe final & Answer base & Answer SFT &
Answer final\tabularnewline
\midrule
\endhead
Truth & OLMo 2 7B & 2.15 & 2.05 & 1.79 & 1.77 & 1.84\tabularnewline
(diverse facts) & Apertus 1.5 8B & 1.95 & 2.09 & 1.41 & 1.64 &
1.68\tabularnewline
& OLMo 3 7B Think & 1.89 & 1.64 & 1.61 & 1.35 & 1.34\tabularnewline
& Gemma 3 4B & 1.82 & 1.93 & 0.81ᵃ & --- & 1.46\tabularnewline
& Gemma 3 12B & 2.24 & 2.26 & 1.64ᵃ & --- & 1.97\tabularnewline
& Qwen3 8B & 2.28 & 2.15 & 1.94 & --- & 1.77\tabularnewline
Truth & OLMo 2 7B & 4.54 & 4.54 & 0.80 & 3.36 & 3.10\tabularnewline
(negated statements) & Apertus 1.5 8B & 4.54 & 4.54 & −0.88 & 4.54 &
4.54\tabularnewline
& OLMo 3 7B Think & 4.54 & 4.31 & 4.54 & 3.77 & 3.80\tabularnewline
& Gemma 3 4B & 4.54 & 4.34 & 0.06ᵃ & --- & 3.46\tabularnewline
& Gemma 3 12B & 4.54 & 4.54 & 1.97ᵃ & --- & 4.54\tabularnewline
& Qwen3 8B & 4.54 & 4.54 & 4.54 & --- & 4.54\tabularnewline
\bottomrule
\end{longtable}

ᵃ The Gemma 3 base models placed only 18--32\% of their probability on
the two answer words; interpret with caution.

Table 3 lists, for all six model families, the base model and the final
post-trained (instruction-tuned) model, and for the three families with
stage-wise checkpoints also the model after supervised fine-tuning. The
simple geographic statements, which we ran for OLMo 2 and Apertus only,
were judged at or near ceiling by all their checkpoints (d′ ≥ 4.38), so
that this set served only as a check. For the negated statements,
however, two base models showed a clear dissociation between
representation and answer: Probes decoded the truth of every statement
(AUC = 1), but the answers of the OLMo 2 base model used this
information only weakly (d′ = 0.80), and those of the Apertus base model
ordered the statements inversely (d′ = −0.88). Both base models answered
``False'' almost always (criterion c = 2.6 and 3.0). Supervised
fine-tuning repaired the answers of both models, whereas the probes
remained at ceiling, and the final post-trained models kept this level.
The Gemma 3 models showed the same pattern between their base and final
versions, although their base models rarely answered with the two answer
words. The base models of Qwen3 and OLMo 3, in contrast, already judged
the negated statements perfectly; post-training left this unchanged in
Qwen3 and lowered it slightly in OLMo 3. Thus, post-training repaired
the readout of negation where the base model read it wrongly, but did
not improve it further where the base model read it correctly. A probe
trained on the simple statements of the same checkpoint suggests what
the base models' answers followed: Applied to the negated statements, it
separated them with the wrong sign (OLMo 2: d′ = −1.28; Apertus: −1.43),
i.e., the general truth direction of the base models encoded whether
city and country belong together and ignored the negation; whether the
answers followed this direction item by item was not tested. After SFT,
the same probe separated the negated statements in the correct direction
(+0.69; +3.05). For the statements on diverse facts, post-training
changed the answers only slightly, except in OLMo 3, where answers and
probes declined together (answers 1.61 → 1.34; probes 1.89 → 1.64).

Why did post-training change the answers for the negated statements but
hardly for the statements on diverse facts? The two sets differ in what
limits the answers. The statements on diverse facts are affirmative, and
judging them requires only that the fact is known: The probes were far
from their maximum, and the answers of the base models were already
close to the probes (e.g., OLMo 2: d′ = 1.79 for the answers vs.~2.15
for the probe), so that there was little represented information that
the answers failed to use. The negated statements require that the fact
is combined with the negation. The probes show that this combination is
available at the last prompt position of every model, but the answers of
several base models did not reflect it. Post-training thus changed the
answers where they fell short of what was represented, not where they
were limited by what the model knows.

Where does the repair arise? We combined the lower layers of one
checkpoint with the upper layers of the other (model stitching; Section
2.7, Figure 2C; results in Figure 7A). Stitching located the repair in
the middle of the network rather than in the early layers. In OLMo 2,
the first 8 base layers combined with all upper SFT layers still judged
at the level of the SFT model (d′ = 3.76 {[}3.47, 4.15{]}, i.e., above
it, another case of a hybrid outside the range of its parents; base
0.63, SFT 3.20 in this subset of 300 pairs), whereas the first 16 base
layers abolished the repair (0.38). In Apertus, the first 8 base layers
under SFT upper layers likewise left the repair intact (4.15), the first
12 removed most of it (1.33), and the first 16 all of it (−0.94);
conversely, the first 8 SFT layers under base upper layers produced part
of the repair (1.21) and the first 12 all of it (4.15). The repair thus
involved layers 9 to 16, with a partial contribution of the layers
below. To ask where in the prompt the repair acts, we exchanged the
internal states between the checkpoints at selected positions of the
prompt (activation patching; Figure 2D). Note that in the prompt for the
truth judgments the question precedes the statement, so that the
statement is read with the task known, and the positions after the
statement contain no question, only the instruction to answer with one
word and the answer cue (``Answer:''). At layer 16, the SFT model's
states at the positions after the statement repaired the base model
completely (OLMo 2: 0.63 → 3.26; Apertus: −0.87 → 4.15), and the base
model's states at these positions removed most or all of the repair in
the SFT model (3.20 → 1.07, the base model being at 0.63; 4.15 → −1.09).
In Apertus, the final position alone carried the full effect in one
direction (base model with the SFT model's final state: 4.15).
Exchanging the states of the statement itself had no effect in Apertus
and in the SFT model of OLMo 2, but a partial one in the OLMo 2 base
model (0.63 → 1.51). At layer 8, the SFT model's states after the
statement repaired the base models only partly (Apertus: 1.61; OLMo 2:
1.07; Figure 7B), in line with the stitching result that the repair is
not complete at this depth. Exchanging the states of the instruction and
question before the statement had no effect at either layer.

\begin{figure}
\centering
\includegraphics[width=1\textwidth,height=0.68\textheight]{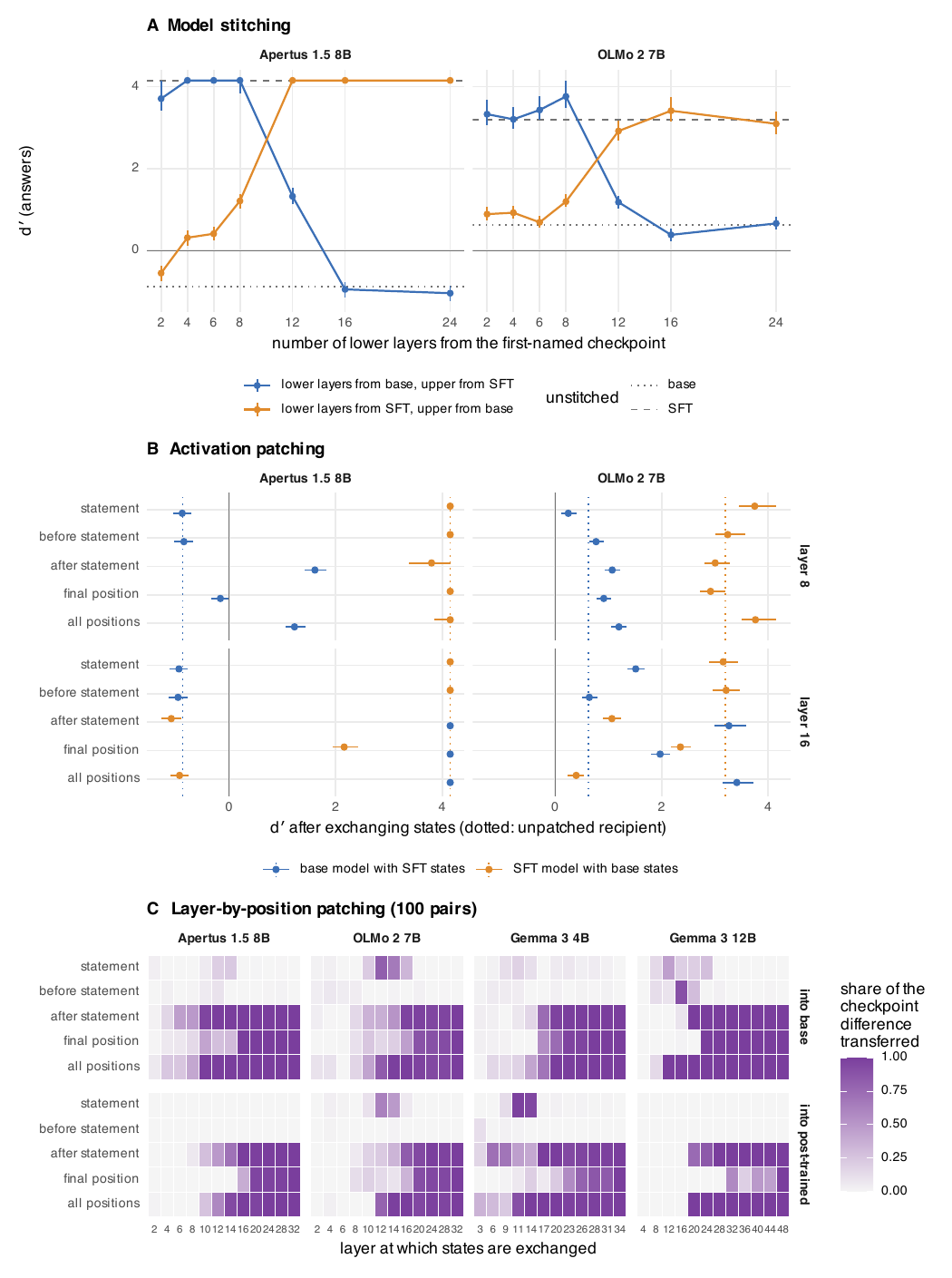}
\caption{Negated statements: where supervised fine-tuning repairs the
readout (300 statement pairs; error bars: 95\% grouped-bootstrap
confidence intervals). (A) Model stitching: sensitivity (d′) of the
answers of hybrids that take the first ℓ layers from one checkpoint and
all layers above from the other; horizontal lines show the unstitched
base (dotted) and SFT (dashed) checkpoints. (B) Activation patching at
layers 8 and 16: d′ of the recipient checkpoint after its residual
stream was replaced by the donor's at one set of positions (the
statement, the instruction before it, the question and answer cue after
it, the final position only, or all positions); dotted lines show the
unpatched recipients. Note that in the truth prompt the question
precedes the statement, so that the positions after the statement
contain only the answer instruction and cue. (C) The same exchange at 12
layers (100 statement pairs), for base and SFT model of OLMo 2 and
Apertus and for base and final post-trained model of Gemma 3 4B and 12B:
share of the difference between the two checkpoints that the exchange
transferred (values below 0 and above 1 are clipped). Upper row: the
base model receives the states of the post-trained model (``into
base''); lower row: the post-trained model receives the states of the
base model.}
\end{figure}

Patching at two layers cannot show where the repair first appears. We
therefore repeated the patching at 12 layers between 2 and 32, with 100
statement pairs (Figure 7C; the values are in the same summary file as
for authorship; on this subset, d′ = 0.34 for the OLMo 2 base model and
3.09 for its SFT model, and −0.86 and 3.64, the maximum for 100 pairs,
for Apertus). We express each effect as the share of the difference
between the two checkpoints that the patch transferred. The two models
differed. In OLMo 2, the repair appeared first at the statement
positions: The SFT model's statement states alone transferred 22\% of
the repair at layer 10, 82\% at layer 12 (d′ = 2.59), and 67\% at layer
14, although all layers above and all other positions were those of the
base model. The share then fell to 33\% at layer 16 and to zero from
layer 20 onward. The positions after the statement showed the
complementary course (36\% at layer 10, 37\% at layer 12, 52\% at layer
14, 97\% at layer 16), and the final position alone carried the repair
only from layer 20 onward (89\%). The reverse direction gave the same
picture: The base model's statement states lowered the SFT model from
3.09 to 1.38 at layer 12 and to 1.75 at layer 14, and had little effect
at layer 16 (2.92) and none above. In OLMo 2, supervised fine-tuning
thus changed how the statement itself is represented at mid-depth, in a
way that the upper layers of the base model could read correctly; the
information then passed to the answer cue and to the final position. In
Apertus, in contrast, the SFT model's statement states transferred at
most 23\% (layers 12 and 14), whereas its states after the statement
transferred 46\% at layer 6 and 96\% at layer 10. In the reverse
direction, the base model's states after the statement lowered the
Apertus SFT model progressively from layer 10 onward, and its statement
states did not lower it at any layer; note that the Apertus SFT model
was at the maximum d′, so that a small disruption would not be visible.
Hence, the two models agree in that the repair has reached the positions
after the statement by the middle of the network, but they differ in
where it arises: at the statement positions in OLMo 2, and at the answer
cue in Apertus.

To see which of the two patterns is more typical, we ran the same sweep
for the two Gemma 3 models, comparing the base model with the final
post-trained model at 12 layers each. These models had not been
localized before, because their base models placed only 19\% and 26\% of
their probability on the two answer words (Section 2.8); the results
should be read with this in mind. Note also that the post-trained Gemma
3 12B, like the Apertus SFT model, was at the maximum d′, so that a
disruption of these models by base states cannot be measured unless it
is large. In both Gemma 3 models, the states after the statement carried
the repair completely from about half of the depth onward (Gemma 3 4B,
34 layers: 75\% at layer 17 and 97\% at layer 20; Gemma 3 12B, 48
layers: 100\% from layer 20). Below that depth, other positions
contributed. In Gemma 3 4B, the post-trained model's statement states
repaired the base model only slightly (at most 19\% {[}15, 26{]}, at
layer 11), whereas the base model's statement states abolished the
repair in the post-trained model almost completely at layers 11 and 14
(d′ = 0.09 and 0.20, from 3.42). In Gemma 3 12B, the post-trained
model's statement states transferred up to 45\% {[}28, 77{]} of the
repair (layer 12), and at layer 16, one third of the depth, the repair
was carried mainly by the positions before the statement, i.e., by the
instruction and question (90\% {[}63, 100{]}, and still 30\% {[}17,
49{]} at layer 20), when the positions after the statement transferred
only 9\%.

Across the four models, the states after the statement thus carried the
repair from about the middle of the network onward. Below that, the
share that the post-trained model's statement states alone transferred
to the base model peaked at 82\% {[}73, 91{]} in OLMo 2 (layer 12), 45\%
{[}28, 77{]} in Gemma 3 12B (layer 12), 23\% {[}20, 27{]} in Apertus
(layer 14), and 19\% {[}15, 26{]} in Gemma 3 4B (layer 11). The
statement representation therefore contributed in all four models,
strongly in OLMo 2 and to a smaller, but reliable extent in the others;
in the reverse direction, which is informative only for OLMo 2 and Gemma
3 4B, the base model's statement states removed about 60\% and nearly
all of the repair, respectively. Note that a patch of the positions
after the statement at a given layer transfers everything that these
positions have gathered from the statement in the layers below; it shows
where the difference between the checkpoints sits at that layer, not
which computation produced it. Exchanging the statement states at all
layers up to a given depth at once would separate the two, and was not
done.

Because the question precedes the statement in this prompt, we also
posed the judgment with the statement placed before the question. The
pattern of the answers was the same, and more pronounced: The base
models ordered the negated statements at or below chance (OLMo 2: d′ =
−0.14 {[}−0.24, −0.05{]}; Apertus: −2.13 {[}−2.29, −1.99{]}), and the
SFT models judged them nearly as with the default prompt (3.42 and
4.54). The repair by supervised fine-tuning thus does not depend on the
question being known when the statement is read. Patching across layers
with this prompt (same 12 layers, 100 statement pairs) gave the same
localization as with the default prompt. In OLMo 2, the SFT model's
statement states alone transferred 36\%, 75\%, and 61\% of the repair at
layers 10, 12, and 14 (base model: d′ = 0.85, 2.14, and 1.69, from
−0.36; SFT model 2.99), and the base model's statement states lowered
the SFT model at the same layers (to 2.13, 1.25, and 1.71); the states
after the statement carried the repair completely from layer 16 onward
(3.09), and the question line by itself transferred nothing in either
direction. In Apertus, the SFT model's statement states did not repair
the base model at any layer (at layers 8 to 12 they even lowered it, to
between −3.01 and −2.64, from −2.11), whereas its states after the
statement transferred the repair partly from layer 4 and completely from
layer 12 onward. Because the statement is read before the question in
this prompt, the contribution of the statement states in OLMo 2 cannot
reflect a reading of the statement in the light of the task: Supervised
fine-tuning changed how this model encodes a negated statement as such.

Steering at the answer position was consistent with a readout that uses
the represented truth direction after SFT; as explained in Appendix D,
this is a consistency check rather than an independent test. Shifting
the internal state at the answer position along the direction that
separates true from false negated statements (±2 SD, layer 21; Appendix
D) moved the answers of both SFT models toward the correct answer, more
than the control directions (Apertus: +0.36 nats, p = .02; OLMo 2:
+1.07, p ≤ .04), and in these models the truth direction coincided with
the decision direction (cosine ≈ 1.0). In the base models, it did not
(Apertus: −0.14, p = .13--.33; OLMo 2: +0.09, p ≥ .54), although in both
base models the positive control, too, exceeded only the norm-matched
controls. Supervised fine-tuning thus aligned the answer with the
represented truth direction; whether the Apertus base model used this
direction with the wrong sign, as its negative sensitivity would
suggest, could not be established.

In summary, the truth of negated statements could be decoded at ceiling
from the last prompt position before and after post-training, the
answers of several base models did not use this information, and
supervised fine-tuning repaired the answers, through weights in the
middle layers. In all four models in which we could trace the repair, it
resided at the positions after the statement from about the middle of
the network onward. Below that, the representation of the statement
itself carried most of it in OLMo 2 and a smaller part in the other
three models (19--45\%), and in Gemma 3 12B the positions before the
statement carried it at one third of the depth. Note that the Gemma 3
base models answered with low answer probability, and that two base
models had nothing to repair.

What do the training data teach? The direction of the change differed
between the judgments: Post-training impaired the readout of authorship,
repaired that of negated statements, and reshaped that of the reviews.
To examine whether this corresponds to the supervised fine-tuning data,
we searched the SFT mixtures of OLMo 2 and Apertus. Both contained
thousands of requests to judge statements as true or false (Tülu 3:
2,685; Apertus, extrapolated: about 2,300), largely factual claims in
Tülu 3 and mathematical statements with explicit reasoning, often
including negations, in Apertus. In contrast, we found no request to
judge whether a text had been written by a human or by an AI: The search
pattern matched five conversations in Tülu 3 and none in the Apertus
sample, and on inspection none of the five posed this judgment (they
concerned, e.g., a training script for an audio classifier and summaries
of research papers; the matches are listed in the repository). The
matches for fake or deceptive reviews were sentiment tasks or summaries
of research papers. Because our judgments were posed in German, we
repeated the search with German patterns in the conversations written in
German, which made up only 0.1\% of Tülu 3 (992 conversations) and 1.7\%
of the Apertus sample (6,330). They contained no request to judge
authorship and no deception judgment, two mentions of AI-generated text
(Apertus), and six true/false questions. Note that a search with regular
expressions can miss requests that are phrased differently, and that the
German subsets are so small that their search has little power. The
census is thus consistent with the idea that post-training improved the
one judgment for which the training data contain examples, where the
base model read it wrongly, and impaired or merely reshaped the
judgments for which we found none; on its own, it cannot establish this.

\hypertarget{the-training-signal-sets-the-direction-of-the-readout}{%
\subsection{3.7 The training signal sets the direction of the
readout}\label{the-training-signal-sets-the-direction-of-the-readout}}

The analyses so far show that post-training changed the readout and that
the direction of the change differed between judgments; whether it
depends on the training data could only be inferred. We therefore
manipulated the training data directly (Section 2.9). Table 4 summarizes
the answers.

\textbf{Table 4.} Training study: sensitivity (d′) of the answers of
OLMo 2 7B after fine-tuning with background conversations only (C0) or
with additional authorship judgments with correct (C1), stereotyped
(C2), or random labels (C3), or with an unrelated topic judgment in the
same layout (C4), and of the official SFT checkpoint. Held-out news: 400
items from the distribution of the training texts. d′, mean over three
seeds (per-seed d′, both answer orders averaged). C0 and C4 did not
learn the answer words of the authorship question (probability of the
two answer words .65--.76, vs. ≈ 1.0 in C1--C3); C3 therefore serves as
a further baseline (see text).

\begin{longtable}[]{@{}lrrrrrr@{}}
\toprule
\begin{minipage}[b]{0.09\columnwidth}\raggedright
Test corpus\strut
\end{minipage} & \begin{minipage}[b]{0.12\columnwidth}\raggedleft
C0 background\strut
\end{minipage} & \begin{minipage}[b]{0.12\columnwidth}\raggedleft
C1 valid\strut
\end{minipage} & \begin{minipage}[b]{0.12\columnwidth}\raggedleft
C2 stereotype\strut
\end{minipage} & \begin{minipage}[b]{0.12\columnwidth}\raggedleft
C3 random\strut
\end{minipage} & \begin{minipage}[b]{0.12\columnwidth}\raggedleft
C4 other task\strut
\end{minipage} & \begin{minipage}[b]{0.12\columnwidth}\raggedleft
Official SFT\strut
\end{minipage}\tabularnewline
\midrule
\endhead
\begin{minipage}[t]{0.09\columnwidth}\raggedright
Held-out news\strut
\end{minipage} & \begin{minipage}[t]{0.12\columnwidth}\raggedleft
−0.30\strut
\end{minipage} & \begin{minipage}[t]{0.12\columnwidth}\raggedleft
4.28\strut
\end{minipage} & \begin{minipage}[t]{0.12\columnwidth}\raggedleft
0.11\strut
\end{minipage} & \begin{minipage}[t]{0.12\columnwidth}\raggedleft
0.13\strut
\end{minipage} & \begin{minipage}[t]{0.12\columnwidth}\raggedleft
0.04\strut
\end{minipage} & \begin{minipage}[t]{0.12\columnwidth}\raggedleft
−0.14\strut
\end{minipage}\tabularnewline
\begin{minipage}[t]{0.09\columnwidth}\raggedright
Newspaper segments\strut
\end{minipage} & \begin{minipage}[t]{0.12\columnwidth}\raggedleft
−0.06\strut
\end{minipage} & \begin{minipage}[t]{0.12\columnwidth}\raggedleft
2.18\strut
\end{minipage} & \begin{minipage}[t]{0.12\columnwidth}\raggedleft
−0.35\strut
\end{minipage} & \begin{minipage}[t]{0.12\columnwidth}\raggedleft
−0.06\strut
\end{minipage} & \begin{minipage}[t]{0.12\columnwidth}\raggedleft
−0.19\strut
\end{minipage} & \begin{minipage}[t]{0.12\columnwidth}\raggedleft
−0.54\strut
\end{minipage}\tabularnewline
\begin{minipage}[t]{0.09\columnwidth}\raggedright
Textbook summaries\strut
\end{minipage} & \begin{minipage}[t]{0.12\columnwidth}\raggedleft
−0.12\strut
\end{minipage} & \begin{minipage}[t]{0.12\columnwidth}\raggedleft
1.39\strut
\end{minipage} & \begin{minipage}[t]{0.12\columnwidth}\raggedleft
0.32\strut
\end{minipage} & \begin{minipage}[t]{0.12\columnwidth}\raggedleft
0.03\strut
\end{minipage} & \begin{minipage}[t]{0.12\columnwidth}\raggedleft
−0.03\strut
\end{minipage} & \begin{minipage}[t]{0.12\columnwidth}\raggedleft
−0.50\strut
\end{minipage}\tabularnewline
\begin{minipage}[t]{0.09\columnwidth}\raggedright
Short chats\strut
\end{minipage} & \begin{minipage}[t]{0.12\columnwidth}\raggedleft
0.04\strut
\end{minipage} & \begin{minipage}[t]{0.12\columnwidth}\raggedleft
−0.58\strut
\end{minipage} & \begin{minipage}[t]{0.12\columnwidth}\raggedleft
−0.29\strut
\end{minipage} & \begin{minipage}[t]{0.12\columnwidth}\raggedleft
−0.21\strut
\end{minipage} & \begin{minipage}[t]{0.12\columnwidth}\raggedleft
0.04\strut
\end{minipage} & \begin{minipage}[t]{0.12\columnwidth}\raggedleft
0.38\strut
\end{minipage}\tabularnewline
\bottomrule
\end{longtable}

Because the background condition C0 did not learn to answer with the two
answer words, whereas the three conditions with authorship examples did,
we compared each condition with three baselines: C0; C3, the condition
with random labels, which learned the question and the answer words but
no valid mapping; and C4, which practiced an unrelated judgment in the
same layout (differences in d′ with 95\% CIs from a grouped bootstrap
that resamples texts and is conditional on the three seeds per
condition). Practicing the unrelated judgment left the authorship
answers where C0 left them (C4 − C0 between −0.13 and +0.10 on the three
test corpora, all CIs including zero; on the held-out news, C4 was
higher than C0, +0.34 {[}0.26, 0.43{]}), so that practice with a
judgment as such did not explain the effects of the authorship labels.
Note that the C4 models never answered ``AI'' (``KI'') in any run, so
that their d′ is a ranking of graded preferences under a saturated
criterion, and that C3 was unstable across seeds (criterion c between
−1.10 and +2.33). Two thousand correctly labeled judgments of news texts
sufficed to raise the sensitivity of the answers far above both
baselines, not only on held-out news but also on texts from other
generators and genres: the newspaper segments, written by GPT-5.4 (C1 −
C3: +2.23 {[}1.56, 3.17{]}; C1 − C4: +2.37 {[}1.74, 3.29{]}), and the
textbook summaries (+1.36 {[}1.11, 1.64{]}; +1.42 {[}1.15, 1.73{]}). On
the held-out news, C1 was at the maximum for both generators and all
text lengths (d′ = 3.8--4.0 are the ceilings of the subsets), so that
its success there was not due to recognizing one generator or one
length; the breakdown cannot exclude that the models detect the
generation procedure that the training and held-out texts share. The
information had been represented all along; a small amount of valid
training made the answer read it. On the short chats, however, the same
training made the answers worse than chance (−0.58), also relative to
random labels (C1 − C3: −0.36 {[}−0.69, −0.06{]}) and to the unrelated
judgment (C1 − C4: −0.62 {[}−1.01, −0.25{]}), because the properties
that distinguished generated from human news did not distinguish
generated from human chat messages in the same way.

The stereotype acted according to its validity, but only partly
reliably. Measured with the formality score used for the labels, AI
texts were less formal than human texts in the chats and newspaper
segments (AUC = .40 {[}.32, .48{]} and .39 {[}.28, .49{]}) but more
formal in the textbook summaries (.70 {[}.65, .75{]}). Relative to
random labels, the stereotype improved the answers on the textbook
summaries (C2 − C3: +0.29 {[}0.11, 0.49{]}; C2 − C4: +0.35 {[}0.15,
0.57{]}) and lowered them on the newspaper segments, though not reliably
(−0.29 {[}−0.61, 0.07{]}; −0.15 {[}−0.57, 0.25{]}); on the chats, the
result depended on the baseline: The stereotype did not differ from
random labels (−0.08 {[}−0.39, 0.24{]}) and was lower than the unrelated
judgment, though not reliably (C2 − C4: −0.33 {[}−0.77, 0.08{]}),
because random labels themselves lowered the answers on the chats. Note
also that the stereotype labels were not independent of other text
properties: The more formal half contained more short texts, more texts
on business and domestic politics than on sports, and more texts from
Gemma 3 12B than from Qwen3 8B. Random labels lowered the answers
slightly on the chats relative to C0 (−0.26 {[}−0.52, 0.00{]}). Note
that the background condition did not reproduce the effect of the
official fine-tuning. The official SFT, which updated all weights on the
full mixture of about 940,000 conversations, lowered the sensitivity on
the newspaper segments and textbook summaries (d′ about −0.5) and
shifted the criterion strongly toward ``human''; our background
condition, with 2\% of the conversations, one epoch, and low-rank
adapters, left the judgments close to those of the base model. The
conditions therefore show that a training signal can set the direction
of the readout against a neutral background; they do not yet show that
the decline after the official fine-tuning has the same origin.
Fine-tuning with a larger share of the original mixture, or removing and
enriching specific parts of it (e.g., conversations in which the
assistant talks about AI), could identify which training data install
the misreading.

Did the training change the representation of the texts or the way it is
read out? Probes trained within each fine-tuned model decoded authorship
at the last prompt position as well as in the base model (conditions C0
to C3; C4 was not included in this analysis; e.g., newspaper segments,
mean of layers 20 and 24, i.e., about two thirds of the depth: d′ = 3.14
in the base model vs. 2.93--3.14 after fine-tuning; chats: 1.42
vs.~1.32--1.45), and probes transferred between the base and the
fine-tuned models without loss at the text positions in every condition
(transfer deficit ≤ 0.02 d′, with one exception of 0.15 in the upper
layers for C1 on the textbook summaries; note that the probes for the
newspaper segments and textbook summaries were near their maximum). At
the last prompt position, in contrast, the deficit grew with depth, most
strongly after training with valid labels (upper third, newspaper
segments: C0 0.22, C1 2.09, C2 0.77, C3 0.86; textbook summaries: 0.25,
1.19, 0.72, 0.97). The training thus left the representation of the text
essentially unchanged, as far as the probes can tell, and reorganized
the position at which the answer is formed. That random labels also
reorganized this position, while barely changing the answers, shows that
what mattered for the answers was which readout was learned, not how
much the answer position changed. The chats illustrate this most
clearly: After training with valid news labels, the model represented
the authorship of the chats as well as before (probe d′ = 1.32) but
judged them below chance (−0.58), consistent with the reversed transfer
of news-trained probes to the chats (Section 3.1). The dissociation that
we observed after the original post-training was thus reproduced
experimentally, by training data whose cues were valid for one genre and
misleading for another.

The account that the valid news labels installed a readout of the cues
of the news, which mislead on the chats, makes a prediction that can be
tested item by item (Section 2.9). We fitted a probe for authorship on
the held-out news in the base model, at the text positions, so that it
is defined independently of any fine-tuned readout, and applied it
unchanged to the chats (mean of layers 20 and 24). This news probe
separated the chats with the wrong sign (d′ = −0.44 {[}−0.70, −0.15{]}).
If training with valid news labels makes the answer follow what
distinguishes AI from human news, the answers of the C1 models to the
individual chats should follow the probe's scores, and those of the
other conditions should not. This was the case. Across the 300 chats,
the answers of the C1 models correlated with the news-probe score
(Spearman ρ = .68 {[}.60, .73{]}, mean of three seeds; within the human
messages .70, within the AI messages .63), whereas those of the control
conditions hardly did (C0: −.11; C3: .04; C4: −.12 {[}−.23, −.01{]});
the difference between C1 and the random-label condition was .63 {[}.51,
.75{]}. Stereotype labels produced a weaker correlation (C2: .34 {[}.22,
.46{]}). With a news probe fitted on the 14 articles of the newspaper
segments, the pattern was the same but weaker (C1: .51; C3: .04), and
this probe did not separate the chats reliably in the base model (−0.11
{[}−0.36, 0.14{]}).

Any classifier that has learned to tell AI from human news might agree
with the C1 models on the chats, whatever happens inside the models. We
therefore repeated the test with two classifiers that use no
activations, logistic regressions on word and on character n-grams of
the same held-out news. They, too, separated the chats with the wrong
sign (d′ = −0.53 {[}−0.80, −0.28{]} and −0.55 {[}−0.84, −0.27{]}), so
that the reversal between the genres is a property of the texts and not
of the model's representation. Their scores also correlated with the
answers of the C1 models (ρ = .32 {[}.21, .42{]} and .43 {[}.33, .53{]})
and not with those of the random-label condition (.07 and .03), but
clearly less than the scores of the news probe. When both predictors
were entered together, the news probe kept its relation to the C1
answers (partial ρ = .63 {[}.54, .70{]} controlling for the word n-gram
score; .59 {[}.49, .67{]} controlling for the character n-gram score),
whereas the n-gram scores added little (.06 {[}−.06, .18{]} and .16
{[}.04, .28{]}). The C1 answers thus followed what the base model's
representation of the text says about the authorship of news, beyond
what surface n-grams of the news predict.

At the last prompt position, we also compared the direction along which
the C1 models' answers to the chats varied with the direction that
separated AI from human news in the same model. The result depended on
how the two directions were estimated: The cosine was .85 (range across
seeds .82--.87) between the difference of the class means and the
difference between chats with high and low dv, .50 (.41--.59) when the
answer direction was estimated by ridge regression, and .28--.36 when
the weight vector of the news probe was used in place of the class-mean
difference, compared with a bound of .43 for 95\% of random directions
and values near zero for C3 and C4. Note also that this comparison is
partly circular, because the last position of a model that answers the
news correctly carries its decision. We therefore do not rest a
conclusion on it. The statistics of the item-level test were specified
in the analysis brief before it was run, which is not documented in the
repository, and the inversion of the chats was known at that time.
Within these limits, the results support the following reading:
Fine-tuning on valid labels made the answers follow the authorship
signal for news that the base model already represents, and the answers
to texts of another genre followed this signal also where it pointed the
wrong way. They do not show along which direction of the fine-tuned
model's internal states this readout runs.

\hypertarget{discussion}{%
\section{4 Discussion}\label{discussion}}

In the present study we used authorship judgments as a test case for the
question of whether post-training changes what a language model
represents or how it reads this representation out. Across six
open-weight base models and three corpora of German texts, the
information that distinguishes AI from human texts was readily
available, both in the texts themselves and in the models' internal
representations, but the answers did not use it, and often ordered the
texts inversely. Post-training lowered the sensitivity of these answers
on average (Δd′ = −0.34 across families; −0.29 across three prompt
wordings), with considerable variation between models, text types, and
prompts, and in the two models with stage-wise checkpoints the change
occurred during supervised fine-tuning. Post-training also changed how
strongly a stated base rate moved the log-odds of the answers, from less
than the change in prior odds to more in most models, although relative
to the larger spread of their log-odds this held less generally. Model
stitching implicated the weights of the early layers, with the caveat
that stitched hybrids produce effects of their own, and activation
patching across layers showed that, from the middle of the network
onward, the states at the positions after the text transferred the
change between checkpoints and the states of the text did not; in the
lower layers, the text states carried part of the change in Apertus and
none of it in OLMo 2, and the positions of the instruction before the
text carried part of it in both. A judgment on public data, the truth of
negated statements, showed the complementary case: The truth could be
decoded at or near ceiling in all six model families, the answers of
several base models did not use it, and post-training repaired the
answers while the probes stayed at or near ceiling. Sweeps over layers
in four models located this repair at the positions after the statement
from the middle of the network onward; below that, the representation of
the statement itself contributed in all four, strongly in OLMo 2 (up to
82\% of the repair) and to a smaller extent in the others (19--45\%). In
a training study, a small number of labeled examples shifted the readout
in the direction their labels implied, and the item-level answers on
another genre followed a probe for the authorship of the training texts
that was fitted in the base model, more closely than they followed
surface classifiers of the same texts. In short, post-training changed
how a judgment is read out of information that the models represent, in
a direction that depended on the judgment. The stronger statement, that
this change leaves the representation of the text untouched and resides
only in the positions that follow the text, held for the probes at the
text positions and, in the patching, for authorship in OLMo 2. In the
other cases, the representation of the text changed as well: in the
lower layers of Apertus for authorship, and below mid-depth for
negation, strongly in OLMo 2 and to a smaller extent in Apertus and the
two Gemma 3 models. What all cases share is that, by the middle of the
network, the change resided at the positions after the text. Note that
the localization rests on two models for authorship and on four models
from three families for negation, and that for the deceptive reviews
(Appendix C) it was inconclusive.

\hypertarget{relation-to-previous-work}{%
\subsection{Relation to previous work}\label{relation-to-previous-work}}

Our findings are consistent with the growing evidence that language
models represent more than they say (Burns et al.,
\protect\hyperlink{ref-burns2023discovering}{2024}; Hu \& Levy,
\protect\hyperlink{ref-hu2023prompting}{2023}; Orgad et al.,
\protect\hyperlink{ref-orgad2025know}{2025}), including for authorship
(Ackerman \& Panickssery,
\protect\hyperlink{ref-ackerman2025inspection}{2025}; Y. Zhou et al.,
\protect\hyperlink{ref-zhou2025cosur}{2026}). They add three points.
First, the dissociation is not only a gap but frequently an inverse
ordering. A miscalibrated decision threshold, which has been proposed as
the main source of probe--answer gaps (Villuri et al.,
\protect\hyperlink{ref-villuri2026probes}{2026}), shifts the criterion
in terms of signal detection theory; it can lower accuracy but cannot
produce sensitivity below chance. Reversed readouts of a discriminative
direction have been described for steering vectors (Torop et al.,
\protect\hyperlink{ref-torop2026inverted}{2026}); here, the inverse
ordering occurred in the graded preferences behind the models'
unmodified answers, and in two model families the class-mean difference
at the answer position carried this inverse ordering (Appendix D). Note,
however, that the cues themselves were specific to the corpus: A probe
trained on the newspaper segments separated the chats with the wrong
sign (Section 3.1). A judgment below chance therefore need not reflect
an inverted reading of a single authorship representation; it can also
reflect a fixed policy whose cues are valid for one kind of text and
invalid for another. Second, the decline is linked to post-training.
Interestingly, in the closest related study, post-training had the
opposite effect: It created rather than abolished the ability of a model
to recognize its own texts (Ackerman \& Panickssery,
\protect\hyperlink{ref-ackerman2025inspection}{2025}). Recognizing one's
own output and judging whether an arbitrary text was written by an AI
are different tasks, and it is conceivable that post-training sharpens
the first while it distorts the second. Third, our results bear on the
superficial alignment hypothesis, according to which a model's knowledge
is acquired during pretraining, whereas post-training mainly teaches it
which formats to use (Zhou et al.,
\protect\hyperlink{ref-zhou2023lima}{2023}). This hypothesis has been
supported by behavioral evidence (Lin et al.,
\protect\hyperlink{ref-lin2023urial}{2023}) and by comparisons of
representations between base and post-trained models (Kissane et al.,
\protect\hyperlink{ref-kissane2024saetransfer}{2024}; Lee et al.,
\protect\hyperlink{ref-lee2024dpo}{2024}; Prakash et al.,
\protect\hyperlink{ref-prakash2024finetuning}{2024}; Waldis et al.,
\protect\hyperlink{ref-waldis2026instructions}{2026}), and it has been
qualified by the observation that post-training gains continue to grow
with the amount of fine-tuning data (Raghavendra et al.,
\protect\hyperlink{ref-raghavendra2024revisiting}{2024}; see also
Vergara-Browne et al.,
\protect\hyperlink{ref-vergarabrowne2026sah}{2026}). To our knowledge,
it has not yet been tested with causal interventions across checkpoints.
Our results provide partial support for it: Probes transferred between
base and post-trained checkpoints at the text positions wherever this
test was not limited by ceiling effects, and from the middle of the
network onward the states of the text could be exchanged between
checkpoints without changing the judgment. Below mid-depth, however,
exchanging the text states did change the judgment in several cases
(Sections 3.4 and 3.6), so that the representation of the text was not
left intact throughout. Note, however, that the information available at
the answer position declined after post-training in some models (Gemma
3, OLMo 3 Think), so that the hypothesis holds for the representation of
the text rather than for the model as a whole. In another respect, our
results qualify it. The change was not superficial in the sense of
affecting only the final stages of processing or the output format; it
involved weights in the early layers and altered how the question takes
up the text, and it had substantive consequences for the quality of the
judgment. This is consistent with the observation that a property
tracked across the stages of post-training is shaped mainly during
supervised fine-tuning (Heidari et al.,
\protect\hyperlink{ref-heidari2026evalawareness}{2026}).

Our localization agrees with previous comparisons of base and
post-trained models in one respect and departs from them in another.
That, from mid-depth onward, the effect of post-training resided in the
positions after the text rather than in those of the text agrees with
model diffing and cache patching, in which much of what fine-tuning adds
was concentrated at chat-template and system-prompt tokens (Leong et
al., \protect\hyperlink{ref-leong2025template}{2025}; Loos \& Eberle,
\protect\hyperlink{ref-loos2026kvpatching}{2026}; Minder et al.,
\protect\hyperlink{ref-minder2025crosscoders}{2026}). Because our
prompts used no chat template, our results suggest that the relevant
positions are those at which the task is posed, whether in template
tokens or in a question phrased in natural language. In contrast, the
stitching results, which implicated the early layers for authorship and
the middle layers for negated statements, depart from the common picture
that fine-tuning mainly alters the upper layers (Merchant et al.,
\protect\hyperlink{ref-merchant2020bert}{2020}; Zhao et al.,
\protect\hyperlink{ref-zhao2026layerwise}{2026}; X. Zhou,
\protect\hyperlink{ref-zhou2026fearcaut}{2026}). It is compatible,
however, with evidence that changes in the late layers can depend on
altered earlier computation (Y. Zhou,
\protect\hyperlink{ref-zhou2026convergence}{2026}), that post-training
retunes routing mechanisms formed in pretraining (Gupta et al.,
\protect\hyperlink{ref-gupta2026cuerouting}{2026}), and that task
representations formed early select later processing pathways
(Bigoulaeva et al.,
\protect\hyperlink{ref-bigoulaeva2026instruction}{2026}). This
conclusion is tempered by the observation that stitched hybrids can
judge worse or better than both parents (Section 3.4), so that stitching
shows where the weights that carry the difference reside rather than
where the difference is computed; a control judgment far from chance
showed no such overshoot, so that stitching does not disturb the models
in general, but whether the overshoot is specific to the changed
judgment or to judgments near chance remains open. Note also that
stitching without an adapter at layer ℓ is equivalent to exchanging the
full residual stream at all positions at that layer, so that our
stitching and patching results are two views of one family of
interventions rather than independent tests. Steering at the answer
position does not add an independent test either, because it largely
restates the difference between the classes in the answers (Appendix D).
Two further cautions apply. A probe shows that information can be
decoded, not that the model uses it (Elazar et al.,
\protect\hyperlink{ref-elazar2021amnesic}{2021}; Hewitt \& Liang,
\protect\hyperlink{ref-hewitt2019control}{2019}); this is why we rest
the claim that post-training changed the readout on the stitching and
patching results rather than on the probes. Conversely, interventions
along a subspace can change the output through a pathway other than the
one the subspace was meant to capture (Makelov et al.,
\protect\hyperlink{ref-makelov2023subspace}{2023}); our steering results
therefore show which way the answer pathway reads a direction, and we
compared them with control directions drawn from the same high-variance
subspace and with the decision direction, but they do not prove that the
model uses this direction under natural conditions.

\hypertarget{the-direction-of-the-change-depends-on-the-training-signal}{%
\subsection{The direction of the change depends on the training
signal}\label{the-direction-of-the-change-depends-on-the-training-signal}}

Post-training impaired the readout of authorship on average, and
repaired the readout of negated statements in the base models that
misread them, whereas it hardly changed the judgments of statements on
diverse facts, for which the answers of the base models were already
close to what could be decoded. A common account of these outcomes
follows from the second lens in Figure 1. Post-training does not add
information to the representation; it installs a way of reading the
representation out. Whether this improves or impairs a judgment then
depends on whether the training signal specifies a valid way of reading
the question. For truth judgments, the supervised fine-tuning data of
both models with stage-wise checkpoints contained thousands of
statements to be judged as true or false; for authorship, we found none.
In terms of the lens model, post-training changes the weights of the
second lens, and these weights are valid only to the extent that the
training data contain valid examples. Two observations qualify this
account. First, the base models of Qwen3 and OLMo 3 already read negated
statements correctly, presumably because their pretraining data
contained instruction-like material; for them, post-training had nothing
to repair. Second, the account runs the risk of circularity: If the
presence of a valid training signal were inferred only from the
improvement of a judgment, it would explain every outcome and predict
none. The inspection of the training data provides an independent
measurement, but a correlational one that rests on two data mixtures and
on a pattern search with limited power; it cannot show which examples
produced the repair (the truth judgments themselves, or, for instance,
the many mathematical and reasoning tasks that teach a model to read a
question precisely).

We therefore manipulated the training signal directly (Section 3.7).
Adding two thousand correctly labeled authorship judgments to otherwise
identical fine-tuning data improved the readout of texts whose cues
resembled those of the training texts and impaired it where they did
not, and the probes at the text positions changed little. On the chats,
the item-level answers followed a probe for the authorship of news that
was fitted in the base model, which turns the post-hoc explanation of
the inversion into a tested prediction; they followed it more closely
than they followed surface n-gram classifiers of the same news, which
suggests that fine-tuning made the answer read features that the base
model already represents. The training signal can thus set the direction
of the readout, as the account assumes. The training study does not
show, however, that the decline after the official fine-tuning of OLMo 2
has the same origin: Our background condition, with a small part of the
original data and low-rank adapters, did not reproduce this decline, and
identifying the parts of real post-training data that install a
misreading remains a task for future work. The deceptive reviews
(Appendix C) fit the account only loosely: Post-training removed a
polarity heuristic that the base models shared, without improving the
judgments.

\hypertarget{sensitivity-criterion-and-the-stated-base-rate}{%
\subsection{Sensitivity, criterion, and the stated base
rate}\label{sensitivity-criterion-and-the-stated-base-rate}}

Signal detection theory was useful here in two ways. First, it separates
what a threshold can explain from what it cannot. Many of the models'
binary answers were constant, which by itself would suggest a
miscalibrated threshold; the graded preferences behind them, however,
were ordered inversely in several models, which no placement of the
criterion can produce. Second, the theory offers a reference point for a
manipulation of the response bias. An ideal observer in a stationary
environment who is told the proportion of signal trials shifts the
criterion by the change in prior log-odds and leaves the evidence for
each stimulus untouched. This is a reference and not a norm for every
situation: Where the environment changes, weighting a stated base rate
less, or sampling answers in proportion to their probability, can be
adequate. The qualitative prediction, that the answers should move with
the stated odds, holds in either case. Human observers typically adjust
less than this (Macmillan \& Creelman,
\protect\hyperlink{ref-macmillan2005detection}{2004}; Phillips \&
Edwards, \protect\hyperlink{ref-phillips1966conservatism}{1966}), and in
some tasks neglect stated base rates in favor of the individual case
(Kahneman \& Tversky,
\protect\hyperlink{ref-kahneman1973prediction}{1973}). The base models
behaved like conservative observers: Their log-odds shifted in the
correct direction, uniformly across texts, but by 6--51\% of the
reference amount. After post-training, the log-odds shifted by more than
the reference amount in five of six model families, and by different
amounts for different texts, so that the stated share no longer acted as
a pure shift of the criterion. Speculatively, a post-trained model
treats a sentence about the share of AI texts less as prior information
to be combined with the evidence and more as a cue about the expected
answer. Two qualifications apply. The post-trained models gave more
extreme log-odds in general, to the texts as well as to the base rate,
and relative to the spread of their own judgments the weight of the base
rate increased in most, but not in all, models and wordings (Section
3.3). Moreover, the benchmark presupposes that the log-odds of the two
answer words are read as a posterior, which is an idealization for base
models that place only about half of their probability on these words.
Within these limits, the comparison shows that post-training changed not
only which texts receive more evidence for ``AI'', but also how the
models combine this evidence with what the prompt states about the
situation, a second aspect of the readout.

\hypertarget{what-reads-the-text}{%
\subsection{What reads the text}\label{what-reads-the-text}}

What does it mean that the change resides at the positions after the
text? In a language model, the answer is produced from the state at the
last position of the prompt. Every position after the text, i.e., the
question, the cue for the answer, and the last position itself,
retrieves information from the text through attention at each layer and
transforms it through the feed-forward blocks. Post-training could
intervene at either step, by changing what is retrieved or by changing
what is made of it, and the stitching results suggest that both are
involved. Speculatively, a psychological description of such a change is
a different construal of the task. A model after post-training may
understand ``written by a human'' differently than its base model did,
much as a task set determines which features of a stimulus an observer
selects and how they are mapped onto a response. In accounts of biased
competition, a task set acts early, at the stage of selection, and not
only at a late decision stage (Desimone \& Duncan,
\protect\hyperlink{ref-desimone1995biased}{1995}). It is instructive to
see that the effect of post-training on the authorship judgments, too,
involved weights in the early layers, and positions as early as the
instruction that precedes the text. Note that our data do not single out
the question as the place of the change: Where we could separate the
tokens of the question from the cue for the answer and the last
position, the latter carried most of the difference (Sections 3.4 and
3.6).

The sweeps over layers show that the positions after the text are not
the whole story. For the negated statements in Apertus, most of the
repair appeared at the positions after the statement (the answer
instruction and cue), as the picture of a changed reader suggests,
although the statement states carried a reliable part of it (up to
23\%). In OLMo 2, most of it appeared in the states of the statement
itself at layers 10 to 14, and the upper layers of the base model read
these states correctly; here, supervised fine-tuning changed what is
read, and the reader of the base model was adequate. The same held when
the statement was placed before the question, so that it was processed
without the task being known: The change of the statement states in OLMo
2 is therefore a change in how this model encodes a negated statement as
such, not in how it reads the statement in the light of the question.
Why the two models differ is open. It is conceivable that the difference
reflects their post-training data, or the degree to which the base model
already encodes negation at the statement positions; our data do not
distinguish these possibilities.

The parallel to human judges is equally instructive. Human readers fail
to recognize AI text because they rely on cues that are plausible but
invalid {[}Jakesch et al.
(\protect\hyperlink{ref-jakesch2023human}{2023}); see also Rosin and
Laubrock, in preparation{]}, which in terms of the lens model is a
failure of cue utilization rather than of cue availability (Brunswik,
\protect\hyperlink{ref-brunswik1956perception}{1956}; Karelaia \&
Hogarth, \protect\hyperlink{ref-karelaia2008lens}{2008}). The models in
our study fail in a similar way: Valid cues are present in the texts and
in the representations, and the answers do not use them.

\hypertarget{why-would-human-oriented-post-training-impair-the-detection-of-ai-text}{%
\subsection{Why would human-oriented post-training impair the detection
of AI
text?}\label{why-would-human-oriented-post-training-impair-the-detection-of-ai-text}}

Our data do not answer this question conclusively, but they constrain
the possible answers. The lens analysis of the chat messages (Appendix
B) showed what most post-trained models took as signs of AI authorship:
long words, adjectives, nouns, and text that is hard to read, i.e., a
formal and nominal style. In our corpus, this was the wrong way round:
The AI messages, written by current models instructed to sound like a
private group chat, were more verbal and lexically more diverse than the
students' messages. Except for Qwen3, whose base model already weighted
these properties inversely (cross-validated G = −.53, after
post-training −.61), the base models used them less. One possible
explanation, which our data cannot test directly, is an outdated
stereotype of AI text: The picture of AI writing as formal, elaborate,
and impersonal fits the output of earlier assistant models and is
widespread in texts about AI, and current models that imitate a group
chat no longer fit it. If this explanation is correct, the error is a
moving target that changes with each generation of models. Two
alternatives are less consistent with our data. Familiarity with the
model's own style would predict that text in a formal assistant register
is judged human, which is the opposite of what we observed, and removing
the surface markers that the students had identified as typical of AI
text did not change the models' judgments. OLMo 2 does not fit the
picture either: After supervised fine-tuning, it weighted the same
properties in the opposite direction and judged the chats slightly
better than its base model. Note also that the lens analysis is
correlational and that many of the text properties are correlated with
each other; it shows which properties the answers follow, not which ones
cause them.

\hypertarget{implications}{%
\subsection{Implications}\label{implications}}

Three implications follow, each of which requires further testing.
First, the side effects of post-training can be localized, and possibly
repaired. A stitched model that combined the first eight layers of the
Apertus base model with all upper layers of the fine-tuned model
discriminated better than either parent, although such hybrids also
produce effects of their own (Section 3.4). Whether such hybrids retain
the intended effects of post-training, such as instruction following, is
an open question, and the approach is related to combining fine-tuned
weights by interpolation or averaging (Wortsman, Ilharco, Kim, et al.,
\protect\hyperlink{ref-wortsman2022wiseft}{2022}; Wortsman, Ilharco,
Gadre, et al., \protect\hyperlink{ref-wortsman2022soups}{2022}). Second,
knowing where a side effect arises makes it possible to protect the
affected components during fine-tuning or to correct them with small,
targeted interventions instead of broader training; in our training
study, two thousand examples sufficed to change the readout in either
direction. That such interventions are practical has been shown, albeit
with the opposite intention, for the refusal behavior that post-training
installs: Refusal is mediated by a single direction in the residual
stream, and removing this direction abolishes it (Arditi et al.,
\protect\hyperlink{ref-arditi2024refusal}{2024}), and tools that
optimize such edits automatically remove refusals while changing the
model's other outputs only minimally (Weidmann,
\protect\hyperlink{ref-weidmann2025heretic}{2025}). Note, however, that
the same ease of editing that makes targeted corrections possible also
shows how fragile safety-related post-training can be. Third, and in our
view most important, post-training can change how a question is read
while the knowledge that the answer would require remains in the model.
The answer then no longer reflects what the model represents. In the
terms of memory research, the information is available but not
accessible (Tulving \& Pearlstone,
\protect\hyperlink{ref-tulving1966availability}{1966}); in the terms of
the second lens in Figure 1, the readout weights the internal cues as if
their validity were different from what it is, and post-training can
make this misweighting worse. Whether language models can access their
own internal states at all is the question of introspection, for which
they show some, but limited and unreliable, abilities (Binder et al.,
\protect\hyperlink{ref-binder2024looking}{2024}; Gurnee et al.,
\protect\hyperlink{ref-gurnee2026verbalizable}{2026}; Kadavath et al.,
\protect\hyperlink{ref-kadavath2022know}{2022}; Lindsey,
\protect\hyperlink{ref-lindsey2025introspection}{2025}). Our case is a
concrete and measurable instance of a model that represents something
but says something else, and it suggests that comparing the
representation at the question positions with the answer might be a
useful tool for auditing models, in the spirit of distinguishing the
presence of a feature from the behavior it supports (Templeton et al.,
\protect\hyperlink{ref-templeton2024scaling}{2024}) and of examining
which representations a model can verbalize (Gurnee et al.,
\protect\hyperlink{ref-gurnee2026verbalizable}{2026}). From the
perspective of self-regulation, in which post-training is often seen as
a top-down control of the dispositions acquired in pretraining, our
results add a cautionary note: A top-down task set can bias perception
and not only correct a decision.

\hypertarget{limitations}{%
\subsection{Limitations}\label{limitations}}

Several limitations qualify these conclusions. We studied models with 4
to 12 billion parameters, and the below-chance judgments might be weaker
or absent in larger models. We derived the sensitivity of the answers
from the graded log-odds of the two answer words; because the binary
answer hardly varied in most conditions, our d′ characterizes the
evidence on which the answer is based rather than the binary decisions
themselves. The information that the models failed to use was largely
lexical and stylistic, as shown by simple n-gram classifiers, so our
results concern the readout of ordinary cues rather than of any deep
representation of authorship. The stitching analyses and the patching
for authorship were restricted to two models, because stage-wise
checkpoints with a clear change were available for these two only. For
the negated statements, the patching across layers additionally included
the two Gemma 3 models, whose base models answer with low probability.
These analyses covered two corpora for authorship and used the plain
prompt format, because base models cannot be queried with a chat
template. The patching across layers was run for one authorship corpus
and with 100 statement pairs, which was what the available computing
time allowed. The two further judgments were studied with four to six
models, steering for them with two models each and on 300 items, and the
effects of post-training on the reviews were small; steering did not
explain them. Steering itself is limited in two ways: It tests whether a
direction is read more than arbitrary directions, which a partly aligned
direction can fail even if it contributes to the answer, and it did not
reveal the change during SFT for authorship, which rests on stitching
and patching. The training study used one base model, low-rank adapters,
and training texts from other generators and of another genre than the
test corpora, and its background condition did not reproduce the effect
of the original fine-tuning; the stereotype labels were moreover
correlated with text length, topic, and generator, a confound that we
noticed only after the training runs and that a stratified assignment of
the labels would avoid. The interventions were applied to partly
different combinations of models and corpora, because the analyses were
added step by step and not all could be repeated for every combination;
only Apertus on the textbook summaries was examined with steering,
stitching, and patching alike. The newspaper segments stem from only 14
articles, so that bootstrap intervals that resample articles are likely
too narrow for this corpus. The authorship judgments used German prompts
(the truth and review judgments used English ones); a paraphrased prompt
reproduced the decline, a prompt with the question before the text
reproduced it on average but not in every family, and we did not test
English prompts for authorship. Models from different developers also
differ in their post-training recipes, including distillation from
larger models, the emphasis on reasoning, and the share of human
preference data, and our design cannot separate these factors; they are
a plausible source of the heterogeneity between models and text types.
Labeled examples improved the judgments of base and post-trained models
alike without closing the gap between them; they show that a correctable
belief contributes to the failure, but not how much of it. Finally, we
defined the family of hypothesis tests for the post-training contrasts
after inspecting the full set of comparisons; we therefore report the
result under a stricter correction across all contrasts as well, which
left five of six decreases significant.

\hypertarget{conclusion}{%
\subsection{Conclusion}\label{conclusion}}

Language models can represent what an answer would require and still
answer otherwise. For authorship, post-training impaired such a readout
on average, and for negated statements it repaired one: Models that
write like humans failed to recognize their own kind of text, not
because they lacked the information but because of how they read it out.
In both judgments, the change between checkpoints was carried by weights
in the middle or early layers and had reached the positions after the
text by the middle of the network; whether it arose there or in the
representation of the text differed between models. To know what a model
knows, it is therefore not enough to look at its answers; one has to
look at how the question is read. A training study showed that the
training data can set the direction of this readout with little change
at the text positions. Future studies with larger models may show how
general this mechanism is, which parts of real post-training data
install a misreading, and whether such side effects can be prevented
without giving up the benefits of post-training.

\hypertarget{references}{%
\section{References}\label{references}}

\hypertarget{refs}{}
\begin{cslreferences}
\leavevmode\hypertarget{ref-ackerman2025inspection}{}%
Ackerman, C., \& Panickssery, N. (2025). \emph{Inspection and control of
self-generated-text recognition ability in llama3-8b-instruct}.
\url{https://arxiv.org/abs/2410.02064}

\leavevmode\hypertarget{ref-arditi2024refusal}{}%
Arditi, A., Obeso, O., Syed, A., Paleka, D., Panickssery, N., Gurnee,
W., \& Nanda, N. (2024). \emph{Refusal in language models is mediated by
a single direction}. \url{https://arxiv.org/abs/2406.11717}

\leavevmode\hypertarget{ref-bansal2021stitching}{}%
Bansal, Y., Nakkiran, P., \& Barak, B. (2021). \emph{Revisiting model
stitching to compare neural representations}.
\url{https://arxiv.org/abs/2106.07682}

\leavevmode\hypertarget{ref-bhattacharjee2024fighting}{}%
Bhattacharjee, A., \& Liu, H. (2024). Fighting fire with fire: Can
chatgpt detect ai-generated text? \emph{ACM SIGKDD Explorations
Newsletter}, \emph{25}(2), 14--21.
\url{https://doi.org/10.1145/3655103.3655106}

\leavevmode\hypertarget{ref-bigoulaeva2026instruction}{}%
Bigoulaeva, I., Rohweder, J., Dutta, S., \& Gurevych, I. (2026).
\emph{Patches of nonlinearity: Instruction vectors in large language
models}. \url{https://arxiv.org/abs/2602.07930}

\leavevmode\hypertarget{ref-binder2024looking}{}%
Binder, F. J., Chua, J., Korbak, T., Sleight, H., Hughes, J., Long, R.,
Perez, E., Turpin, M., \& Evans, O. (2024). \emph{Looking inward:
Language models can learn about themselves by introspection}. arXiv.
\url{https://doi.org/10.48550/ARXIV.2410.13787}

\leavevmode\hypertarget{ref-block2019gnad}{}%
Block, T. (2019). \emph{Ten thousand german news articles dataset
(10kGNAD)}. GitHub repository. \url{https://github.com/tblock/10kGNAD}

\leavevmode\hypertarget{ref-brunswik1956perception}{}%
Brunswik, E. (1956). \emph{Perception and the representative design of
psychological experiments}. University of California Press.
\url{https://doi.org/10.1525/9780520350519}

\leavevmode\hypertarget{ref-burns2023discovering}{}%
Burns, C., Ye, H., Klein, D., \& Steinhardt, J. (2024).
\emph{Discovering latent knowledge in language models without
supervision}. \url{https://arxiv.org/abs/2212.03827}

\leavevmode\hypertarget{ref-clark2021human}{}%
Clark, E., August, T., Serrano, S., Haduong, N., Gururangan, S., \&
Smith, N. A. (2021). All that's ``human'' is not gold: Evaluating human
evaluation of generated text. In C. Zong; F. Xia; W. Li; \& R. Navigli
(Eds.), \emph{Proceedings of the 59th annual meeting of the Association
for Computational Linguistics and the 11th international joint
conference on natural language processing (volume 1: Long papers)} (pp.
7282--7296). Association for Computational Linguistics.
\url{https://doi.org/10.18653/v1/2021.acl-long.565}

\leavevmode\hypertarget{ref-desimone1995biased}{}%
Desimone, R., \& Duncan, J. (1995). Neural mechanisms of selective
visual attention. \emph{Annual Review of Neuroscience}, \emph{18}(1),
193--222. \url{https://doi.org/10.1146/annurev.ne.18.030195.001205}

\leavevmode\hypertarget{ref-elazar2021amnesic}{}%
Elazar, Y., Ravfogel, S., Jacovi, A., \& Goldberg, Y. (2021). Amnesic
probing: Behavioral explanation with amnesic counterfactuals.
\emph{Transactions of the Association for Computational Linguistics},
\emph{9}, 160--175. \url{https://doi.org/10.1162/tacl_a_00359}

\leavevmode\hypertarget{ref-funder1995accuracy}{}%
Funder, D. C. (1995). On the accuracy of personality judgment: A
realistic approach. \emph{Psychological Review}, \emph{102}(4),
652--670. \url{https://doi.org/10.1037/0033-295x.102.4.652}

\leavevmode\hypertarget{ref-gemma2025gemma3}{}%
Gemma Team, Kamath, A., Ferret, J., Pathak, S., Vieillard, N., Merhej,
R., Perrin, S., Matejovicova, T., Ramé, A., Rivière, M., Rouillard, L.,
Mesnard, T., Cideron, G., Grill, J.-b., Ramos, S., Yvinec, E., Casbon,
M., Pot, E., Penchev, I., \ldots{} Hussenot, L. (2025). \emph{Gemma 3
technical report}. \url{https://arxiv.org/abs/2503.19786}

\leavevmode\hypertarget{ref-gupta2026cuerouting}{}%
Gupta, P., Zhang, T. J., \& Jin, Z. (2026). \emph{Many biases, one
circuit: Post-training tunes a pretrained cue-routing mechanism}.
\url{https://arxiv.org/abs/2610.22937}

\leavevmode\hypertarget{ref-gurnee2026verbalizable}{}%
Gurnee, W., Sofroniew, N., Pearce, A., Piotrowski, M., Kauvar, I., Chen,
R., Soligo, A., Bogdan, P., Ong, E., Wang, R., Thompson, B., Abrahams,
D., Kantamneni, S., Ameisen, E., Batson, J., \& Lindsey, J. (2026).
\emph{Verbalizable representations form a global workspace in language
models}. \url{https://arxiv.org/abs/2607.15495}

\leavevmode\hypertarget{ref-hautus1995corrections}{}%
Hautus, M. J. (1995). Corrections for extreme proportions and their
biasing effects on estimated values of d′. \emph{Behavior Research
Methods, Instruments, \& Computers}, \emph{27}(1), 46--51.
\url{https://doi.org/10.3758/bf03203619}

\leavevmode\hypertarget{ref-heidari2026evalawareness}{}%
Heidari, F., Memarian, A., \& Rabusseau, G. (2026). \emph{Evaluation
awareness in language models: Representation, verbalization, and
control}. \url{https://arxiv.org/abs/2608.21766}

\leavevmode\hypertarget{ref-heimersheim2024patching}{}%
Heimersheim, S., \& Nanda, N. (2024). \emph{How to use and interpret
activation patching}. \url{https://arxiv.org/abs/2404.15255}

\leavevmode\hypertarget{ref-apertus2025}{}%
Hernández-Cano, A., Hägele, A., Huang, A. H., Romanou, A., Solergibert,
A.-J., Pásztor, B., Messmer, B., Garbaya, D., Ďurech, E. F., Hakimi, I.,
Giraldo, J. G., Ismayilzada, M., Foroutan, N., Moalla, S., Chen, T.,
Sabolčec, V., Xu, Y., Aerni, M., AlKhamissi, B., \ldots{} Schlag, I.
(2026). Apertus: Democratizing open and compliant LLMs for global
language environments. \emph{Proceedings of the 64th Annual Meeting of
the Association for Computational Linguistics (Volume 1: Long Papers)},
46877--46955. \url{https://doi.org/10.18653/v1/2026.acl-long.2172}

\leavevmode\hypertarget{ref-hewitt2019control}{}%
Hewitt, J., \& Liang, P. (2019). Designing and interpreting probes with
control tasks. In K. Inui; J. Jiang; V. Ng; \& X. Wan (Eds.),
\emph{Proceedings of the 2019 conference on empirical methods in natural
language processing and the 9th international joint conference on
natural language processing (EMNLP-IJCNLP)} (pp. 2733--2743).
Association for Computational Linguistics.
\url{https://doi.org/10.18653/v1/D19-1275}

\leavevmode\hypertarget{ref-hu2022lora}{}%
Hu, E. J., Shen, Y., Wallis, P., Allen-Zhu, Z., Li, Y., Wang, S., Wang,
L., \& Chen, W. (2021). \emph{LoRA: Low-rank adaptation of large
language models}. arXiv. \url{https://doi.org/10.48550/ARXIV.2106.09685}

\leavevmode\hypertarget{ref-hu2023prompting}{}%
Hu, J., \& Levy, R. (2023). Prompting is not a substitute for
probability measurements in large language models. In H. Bouamor; J.
Pino; \& K. Bali (Eds.), \emph{Proceedings of the 2023 conference on
empirical methods in natural language processing} (pp. 5040--5060).
Association for Computational Linguistics.
\url{https://doi.org/10.18653/v1/2023.emnlp-main.306}

\leavevmode\hypertarget{ref-jakesch2023human}{}%
Jakesch, M., Hancock, J. T., \& Naaman, M. (2023). Human heuristics for
ai-generated language are flawed. \emph{Proceedings of the National
Academy of Sciences}, \emph{120}(11).
\url{https://doi.org/10.1073/pnas.2208839120}

\leavevmode\hypertarget{ref-jones2025turing}{}%
Jones, C. R., \& Bergen, B. K. (2025). \emph{Large language models pass
the turing test}. \url{https://arxiv.org/abs/2503.23674}

\leavevmode\hypertarget{ref-kadavath2022know}{}%
Kadavath, S., Conerly, T., Askell, A., Henighan, T., Drain, D., Perez,
E., Schiefer, N., Hatfield-Dodds, Z., DasSarma, N., Tran-Johnson, E.,
Johnston, S., El-Showk, S., Jones, A., Elhage, N., Hume, T., Chen, A.,
Bai, Y., Bowman, S., Fort, S., \ldots{} Kaplan, J. (2022).
\emph{Language models (mostly) know what they know}.
\url{https://arxiv.org/abs/2207.05221}

\leavevmode\hypertarget{ref-kahneman1973prediction}{}%
Kahneman, D., \& Tversky, A. (1973). On the psychology of prediction.
\emph{Psychological Review}, \emph{80}(4), 237--251.
\url{https://doi.org/10.1037/h0034747}

\leavevmode\hypertarget{ref-karelaia2008lens}{}%
Karelaia, N., \& Hogarth, R. M. (2008). Determinants of linear judgment:
A meta-analysis of lens model studies. \emph{Psychological Bulletin},
\emph{134}(3), 404--426.
\url{https://doi.org/10.1037/0033-2909.134.3.404}

\leavevmode\hypertarget{ref-kissane2024saetransfer}{}%
Kissane, C., Krzyzanowski, R., Conmy, A., \& Nanda, N. (2024).
\emph{SAEs (usually) transfer between base and chat models}. Alignment
Forum.
\url{https://www.alignmentforum.org/posts/fmwk6qxrpW8d4jvbd/saes-usually-transfer-between-base-and-chat-models}

\leavevmode\hypertarget{ref-kornblith2019similarity}{}%
Kornblith, S., Norouzi, M., Lee, H., \& Hinton, G. (2019). Similarity of
neural network representations revisited. In K. Chaudhuri \& R.
Salakhutdinov (Eds.), \emph{Proceedings of the 36th international
conference on machine learning} (Vol. 97, pp. 3519--3529). PMLR.
\url{https://proceedings.mlr.press/v97/kornblith19a.html}

\leavevmode\hypertarget{ref-kobis2021angelou}{}%
Köbis, N., \& Mossink, L. D. (2021). Artificial intelligence versus maya
angelou: Experimental evidence that people cannot differentiate
ai-generated from human-written poetry. \emph{Computers in Human
Behavior}, \emph{114}, 106553.
\url{https://doi.org/10.1016/j.chb.2020.106553}

\leavevmode\hypertarget{ref-lambert2024tulu3}{}%
Lambert, N., Morrison, J., Pyatkin, V., Huang, S., Ivison, H., Brahman,
F., Miranda, L. J. V., Liu, A., Dziri, N., Lyu, S., Gu, Y., Malik, S.,
Graf, V., Hwang, J. D., Yang, J., Bras, R. L., Tafjord, O., Wilhelm, C.,
Soldaini, L., \ldots{} Hajishirzi, H. (2025). \emph{Tulu 3: Pushing
frontiers in open language model post-training}.
\url{https://arxiv.org/abs/2411.15124}

\leavevmode\hypertarget{ref-lee2024dpo}{}%
Lee, A., Bai, X., Pres, I., Wattenberg, M., Kummerfeld, J. K., \&
Mihalcea, R. (2024). \emph{A mechanistic understanding of alignment
algorithms: A case study on dpo and toxicity}.
\url{https://arxiv.org/abs/2401.01967}

\leavevmode\hypertarget{ref-lenc2015equivariance}{}%
Lenc, K., \& Vedaldi, A. (2015). Understanding image representations by
measuring their equivariance and equivalence. \emph{2015 Ieee Conference
on Computer Vision and Pattern Recognition (Cvpr)}, 991--999.
\url{https://doi.org/10.1109/cvpr.2015.7298701}

\leavevmode\hypertarget{ref-leong2025template}{}%
Leong, C. T., Yin, Q., Wang, J., \& Li, W. (2025). \emph{Why safeguarded
ships run aground? Aligned large language models' safety mechanisms tend
to be anchored in the template region}.
\url{https://arxiv.org/abs/2502.13946}

\leavevmode\hypertarget{ref-lin2023urial}{}%
Lin, B. Y., Ravichander, A., Lu, X., Dziri, N., Sclar, M., Chandu, K.,
Bhagavatula, C., \& Choi, Y. (2023). \emph{The unlocking spell on base
llms: Rethinking alignment via in-context learning}.
\url{https://arxiv.org/abs/2312.01552}

\leavevmode\hypertarget{ref-lindsey2025introspection}{}%
Lindsey, J. (2025). \emph{Emergent introspective awareness in large
language models}. Transformer Circuits Thread.
\url{https://transformer-circuits.pub/2025/introspection/index.html}

\leavevmode\hypertarget{ref-lindsey2024crosscoders}{}%
Lindsey, J., Templeton, A., Marcus, J., Conerly, T., Batson, J., \&
Olah, C. (2024). \emph{Sparse crosscoders for cross-layer features and
model diffing}. Transformer Circuits Thread.
\url{https://transformer-circuits.pub/2024/crosscoders/index.html}

\leavevmode\hypertarget{ref-loos2026kvpatching}{}%
Loos, J., \& Eberle, O. (2026). \emph{KV-cache patching: Tracing how and
where fine-tuned behaviors arise}.
\url{https://arxiv.org/abs/2610.41480}

\leavevmode\hypertarget{ref-macmillan2005detection}{}%
Macmillan, N. A., \& Creelman, C. D. (2004). \emph{Detection theory}.
Psychology Press. \url{https://doi.org/10.4324/9781410611147}

\leavevmode\hypertarget{ref-makelov2023subspace}{}%
Makelov, A., Lange, G., \& Nanda, N. (2023). \emph{Is this the subspace
you are looking for? An interpretability illusion for subspace
activation patching}. \url{https://arxiv.org/abs/2311.17030}

\leavevmode\hypertarget{ref-marks2024geometry}{}%
Marks, S., \& Tegmark, M. (2024). \emph{The geometry of truth: Emergent
linear structure in large language model representations of true/false
datasets}. \url{https://arxiv.org/abs/2310.06824}

\leavevmode\hypertarget{ref-meng2022rome}{}%
Meng, K., Bau, D., Andonian, A., \& Belinkov, Y. (2023). \emph{Locating
and editing factual associations in gpt}.
\url{https://arxiv.org/abs/2202.05262}

\leavevmode\hypertarget{ref-merchant2020bert}{}%
Merchant, A., Rahimtoroghi, E., Pavlick, E., \& Tenney, I. (2020).
\emph{What happens to bert embeddings during fine-tuning?}
\url{https://arxiv.org/abs/2004.14448}

\leavevmode\hypertarget{ref-minder2025crosscoders}{}%
Minder, J., Dumas, C., Juang, C., Chugtai, B., \& Nanda, N. (2026).
\emph{Overcoming sparsity artifacts in crosscoders to interpret
chat-tuning}. \url{https://arxiv.org/abs/2504.02922}

\leavevmode\hypertarget{ref-orgad2025know}{}%
Orgad, H., Toker, M., Gekhman, Z., Reichart, R., Szpektor, I., Kotek,
H., \& Belinkov, Y. (2025). \emph{LLMs know more than they show: On the
intrinsic representation of llm hallucinations}.
\url{https://arxiv.org/abs/2410.02707}

\leavevmode\hypertarget{ref-ott2013negative}{}%
Ott, M., Cardie, C., \& Hancock, J. T. (2013). Negative deceptive
opinion spam. In L. Vanderwende; H. Daumé III; \& K. Kirchhoff (Eds.),
\emph{Proceedings of the 2013 conference of the North American chapter
of the Association for Computational Linguistics: Human language
technologies} (pp. 497--501). Association for Computational Linguistics.
\url{https://aclanthology.org/N13-1053/}

\leavevmode\hypertarget{ref-ott2011finding}{}%
Ott, M., Choi, Y., Cardie, C., \& Hancock, J. T. (2011). Finding
deceptive opinion spam by any stretch of the imagination. In D. Lin; Y.
Matsumoto; \& R. Mihalcea (Eds.), \emph{Proceedings of the 49th annual
meeting of the Association for Computational Linguistics: Human language
technologies} (pp. 309--319). Association for Computational Linguistics.
\url{https://aclanthology.org/P11-1032/}

\leavevmode\hypertarget{ref-panickssery2024evaluators}{}%
Panickssery, A., Bowman, S. R., \& Feng, S. (2024). \emph{LLM evaluators
recognize and favor their own generations}.
\url{https://arxiv.org/abs/2404.13076}

\leavevmode\hypertarget{ref-panzeri2017cracking}{}%
Panzeri, S., Harvey, C. D., Piasini, E., Latham, P. E., \& Fellin, T.
(2017). Cracking the neural code for sensory perception by combining
statistics, intervention, and behavior. \emph{Neuron}, \emph{93}(3),
491--507. \url{https://doi.org/10.1016/j.neuron.2016.12.036}

\leavevmode\hypertarget{ref-phillips1966conservatism}{}%
Phillips, L. D., \& Edwards, W. (1966). Conservatism in a simple
probability inference task. \emph{Journal of Experimental Psychology},
\emph{72}(3), 346--354. \url{https://doi.org/10.1037/h0023653}

\leavevmode\hypertarget{ref-pimentel2024compute}{}%
Pimentel, T., \& Meister, C. (2024). How to compute the probability of a
word. In Y. Al-Onaizan; M. Bansal; \& Y.-N. Chen (Eds.),
\emph{Proceedings of the 2024 conference on empirical methods in natural
language processing} (pp. 18358--18375). Association for Computational
Linguistics. \url{https://doi.org/10.18653/v1/2024.emnlp-main.1020}

\leavevmode\hypertarget{ref-porter2024poetry}{}%
Porter, B., \& Machery, E. (2024). AI-generated poetry is
indistinguishable from human-written poetry and is rated more favorably.
\emph{Scientific Reports}, \emph{14}(1).
\url{https://doi.org/10.1038/s41598-024-76900-1}

\leavevmode\hypertarget{ref-prakash2024finetuning}{}%
Prakash, N., Shaham, T. R., Haklay, T., Belinkov, Y., \& Bau, D. (2024).
\emph{Fine-tuning enhances existing mechanisms: A case study on entity
tracking}. \url{https://arxiv.org/abs/2402.14811}

\leavevmode\hypertarget{ref-apertus2026v15}{}%
Project Apertus, Xu, Y., Turki, Y., Sabolčec, V., Pyatkin, V., Chu, T.,
Najem-Meyer, S., Carroll, S. F., Smith, R., Kreft, R., Tazza, O. A.,
Lavrovsky, O., Irmiger, N., Santelmo, M., Schneider, M., Giraldo, J. G.,
Yukhymenko, H., Ďurech, E., Melikidze, D., \ldots{} Schlag, I. (2026).
\emph{Apertus 1.5: From open text models to open multimodal reasoners}.
\url{https://arxiv.org/abs/2610.15542}

\leavevmode\hypertarget{ref-raghavendra2024revisiting}{}%
Raghavendra, M., Nath, V., \& Hendryx, S. (2024). \emph{Revisiting the
superficial alignment hypothesis}.
\url{https://arxiv.org/abs/2410.03717}

\leavevmode\hypertarget{ref-rathi2024gpt4}{}%
Rathi, I., Taylor, S., Bergen, B. K., \& Jones, C. R. (2024).
\emph{GPT-4 is judged more human than humans in displaced and inverted
turing tests}. \url{https://arxiv.org/abs/2407.08853}

\leavevmode\hypertarget{ref-russell2025frequent}{}%
Russell, J., Karpinska, M., \& Iyyer, M. (2025). People who frequently
use ChatGPT for writing tasks are accurate and robust detectors of
AI-generated text. In W. Che; J. Nabende; E. Shutova; \& M. T. Pilehvar
(Eds.), \emph{Proceedings of the 63rd annual meeting of the Association
for Computational Linguistics (volume 1: Long papers)} (pp. 5342--5373).
Association for Computational Linguistics.
\url{https://doi.org/10.18653/v1/2025.acl-long.267}

\leavevmode\hypertarget{ref-schabus2017onemillion}{}%
Schabus, D., Skowron, M., \& Trapp, M. (2017). One million posts: A data
set of german online discussions. \emph{Proceedings of the 40th
International Acm Sigir Conference on Research and Development in
Information Retrieval, SIGIR '17}, 1241--1244.
\url{https://doi.org/10.1145/3077136.3080711}

\leavevmode\hypertarget{ref-olmo2025olmo3}{}%
Team Olmo, Ettinger, A., Bertsch, A., Kuehl, B., Graham, D., Heineman,
D., Groeneveld, D., Brahman, F., Timbers, F., Ivison, H., Morrison, J.,
Poznanski, J., Lo, K., Soldaini, L., Jordan, M., Chen, M., Noukhovitch,
M., Lambert, N., Walsh, P., \ldots{} Hajishirzi, H. (2026). \emph{Olmo
3}. \url{https://arxiv.org/abs/2512.13961}

\leavevmode\hypertarget{ref-olmo2024olmo2}{}%
Team OLMo, Walsh, P., Soldaini, L., Groeneveld, D., Lo, K., Arora, S.,
Bhagia, A., Gu, Y., Huang, S., Jordan, M., Lambert, N., Schwenk, D.,
Tafjord, O., Anderson, T., Atkinson, D., Brahman, F., Clark, C., Dasigi,
P., Dziri, N., \ldots{} Hajishirzi, H. (2025). \emph{2 OLMo 2 Furious}.
\url{https://arxiv.org/abs/2501.00656}

\leavevmode\hypertarget{ref-templeton2024scaling}{}%
Templeton, A., Conerly, T., Marcus, J., Lindsey, J., Bricken, T., Chen,
B., Pearce, A., Citro, C., Ameisen, E., Jones, A., Cunningham, H.,
Turner, N. L., McDougall, C., MacDiarmid, M., Tamkin, A., Durmus, E.,
Hume, T., Mosconi, F., Freeman, C. D., \ldots{} Henighan, T. (2024).
\emph{Scaling monosemanticity: Extracting interpretable features from
claude 3 sonnet}. \url{https://arxiv.org/abs/2406.04093}

\leavevmode\hypertarget{ref-thai2026editlens}{}%
Thai, K., Emi, B., Masrour, E., \& Iyyer, M. (2025). \emph{EditLens:
Quantifying the extent of ai editing in text}.
\url{https://arxiv.org/abs/2510.03154}

\leavevmode\hypertarget{ref-torop2026inverted}{}%
Torop, M., Masoomi, A., \& Dy, J. (2026). \emph{Inverted detection and
control in steering vectors}. \url{https://arxiv.org/abs/2608.02957}

\leavevmode\hypertarget{ref-tucker1964lme}{}%
Tucker, L. R. (1964). A suggested alternative formulation in the
developments by hursch, hammond, and hursch, and by hammond, hursch, and
todd. \emph{Psychological Review}, \emph{71}(6), 528--530.
\url{https://doi.org/10.1037/h0047061}

\leavevmode\hypertarget{ref-tulving1966availability}{}%
Tulving, E., \& Pearlstone, Z. (1966). Availability versus accessibility
of information in memory for words. \emph{Journal of Verbal Learning and
Verbal Behavior}, \emph{5}(4), 381--391.
\url{https://doi.org/10.1016/s0022-5371(66)80048-8}

\leavevmode\hypertarget{ref-vergarabrowne2026sah}{}%
Vergara-Browne, T., Patil, D., Titov, I., Reddy, S., Pimentel, T., \&
Mosbach, M. (2026). \emph{Operationalising the superficial alignment
hypothesis via task complexity}. \url{https://arxiv.org/abs/2602.15829}

\leavevmode\hypertarget{ref-vig2020causal}{}%
Vig, J., Gehrmann, S., Belinkov, Y., Qian, S., Nevo, D., Sakenis, S.,
Huang, J., Singer, Y., \& Shieber, S. (2020). \emph{Causal mediation
analysis for interpreting neural nlp: The case of gender bias}.
\url{https://arxiv.org/abs/2004.12265}

\leavevmode\hypertarget{ref-villuri2026probes}{}%
Villuri, G., Shaik, H., \& Doboli, A. (2026). \emph{When do internal
probes beat reading the answer? Miscalibrated readouts and
behavior-concealed knowledge in language models}.
\url{https://arxiv.org/abs/2609.04582}

\leavevmode\hypertarget{ref-waldis2026instructions}{}%
Waldis, A., Choshen, L., Hou, Y., \& Perlitz, Y. (2026).
\emph{Instructions shape production of language, not processing}.
\url{https://arxiv.org/abs/2605.11206}

\leavevmode\hypertarget{ref-weidmann2025heretic}{}%
Weidmann, P. E. (2025). Heretic: Training-free language model
modification. In \emph{GitHub repository}.
\url{https://github.com/p-e-w/heretic}; GitHub.

\leavevmode\hypertarget{ref-wortsman2022soups}{}%
Wortsman, M., Ilharco, G., Gadre, S. Y., Roelofs, R., Gontijo-Lopes, R.,
Morcos, A. S., Namkoong, H., Farhadi, A., Carmon, Y., Kornblith, S., \&
Schmidt, L. (2022). \emph{Model soups: Averaging weights of multiple
fine-tuned models improves accuracy without increasing inference time}.
\url{https://arxiv.org/abs/2203.05482}

\leavevmode\hypertarget{ref-wortsman2022wiseft}{}%
Wortsman, M., Ilharco, G., Kim, J. W., Li, M., Kornblith, S., Roelofs,
R., Gontijo-Lopes, R., Hajishirzi, H., Farhadi, A., Namkoong, H., \&
Schmidt, L. (2022). \emph{Robust fine-tuning of zero-shot models}.
\url{https://arxiv.org/abs/2109.01903}

\leavevmode\hypertarget{ref-yang2025qwen3}{}%
Yang, A., Li, A., Yang, B., Zhang, B., Hui, B., Zheng, B., Yu, B., Gao,
C., Huang, C., Lv, C., Zheng, C., Liu, D., Zhou, F., Huang, F., Hu, F.,
Ge, H., Wei, H., Lin, H., Tang, J., \ldots{} Qiu, Z. (2025). \emph{Qwen3
technical report}. \url{https://arxiv.org/abs/2505.09388}

\leavevmode\hypertarget{ref-yuan2026crossgen}{}%
Yuan, H., Guo, J., Qiu, W., Huang, Z., Li, R., \& Li, S. (2026).
\emph{Using llms to detect llm-generated texts: A cross-generation
analysis}. \url{https://arxiv.org/abs/2609.34691}

\leavevmode\hypertarget{ref-zhao2026layerwise}{}%
Zhao, Q., Gong, X., Chen, X., Kang, Z., \& Li, X. (2026). \emph{A
layer-wise analysis of supervised fine-tuning}.
\url{https://arxiv.org/abs/2604.11838}

\leavevmode\hypertarget{ref-zhou2023lima}{}%
Zhou, C., Liu, P., Xu, P., Iyer, S., Sun, J., Mao, Y., Ma, X., Efrat,
A., Yu, P., Yu, L., Zhang, S., Ghosh, G., Lewis, M., Zettlemoyer, L., \&
Levy, O. (2023). \emph{LIMA: Less is more for alignment}.
\url{https://arxiv.org/abs/2305.11206}

\leavevmode\hypertarget{ref-zhou2026fearcaut}{}%
Zhou, X. (2026). \emph{FearCaut-qwen: Affective steering in a
vision-language model shifts the decision criterion for hazard
assessment}. \url{https://arxiv.org/abs/2610.11986}

\leavevmode\hypertarget{ref-zhou2026convergence}{}%
Zhou, Y. (2026). \emph{The convergence gap: Instruction-tuned language
models stabilize later in the forward pass}.
\url{https://arxiv.org/abs/2605.07282}

\leavevmode\hypertarget{ref-zhou2025cosur}{}%
Zhou, Y., Zhu, W., Wen, J., Peng, W., Wu, Z., \& Xue, Y. (2026).
\emph{From implicit to explicit: Enhancing self-recognition in large
language models}. \url{https://arxiv.org/abs/2508.14408}
\end{cslreferences}

\hypertarget{appendix}{%
\section{Appendix}\label{appendix}}

\hypertarget{a-terminology-across-disciplines}{%
\subsection{A Terminology across
disciplines}\label{a-terminology-across-disciplines}}

This study combines concepts from the psychology of judgment with
methods from machine learning and interpretability research. The two
fields often describe the same idea with different words, and they
sometimes use the same word for different ideas. Table A1 lists the
correspondences that are relevant here, ordered by the two lenses of
Figure 1 (terms marked † are used in this paper); Table A2 lists terms
that are used in both fields with different meanings and states how we
use them in this paper.

\textbf{Table A1. Correspondences between the terminology of psychology
and statistics and that of machine learning and interpretability
research.}

\begin{longtable}[]{@{}lll@{}}
\toprule
\begin{minipage}[b]{0.30\columnwidth}\raggedright
Psychology / statistics\strut
\end{minipage} & \begin{minipage}[b]{0.30\columnwidth}\raggedright
Machine learning / interpretability\strut
\end{minipage} & \begin{minipage}[b]{0.30\columnwidth}\raggedright
Meaning in this study\strut
\end{minipage}\tabularnewline
\midrule
\endhead
\begin{minipage}[t]{0.30\columnwidth}\raggedright
\emph{The state of the world}\strut
\end{minipage} & \begin{minipage}[t]{0.30\columnwidth}\raggedright
\strut
\end{minipage} & \begin{minipage}[t]{0.30\columnwidth}\raggedright
\strut
\end{minipage}\tabularnewline
\begin{minipage}[t]{0.30\columnwidth}\raggedright
state of the world, distal criterion (Brunswik)†\strut
\end{minipage} & \begin{minipage}[t]{0.30\columnwidth}\raggedright
ground truth, label, target\strut
\end{minipage} & \begin{minipage}[t]{0.30\columnwidth}\raggedright
whether a text was written by an AI, whether a statement is true,
whether a review is genuine\strut
\end{minipage}\tabularnewline
\begin{minipage}[t]{0.30\columnwidth}\raggedright
proximal cue†\strut
\end{minipage} & \begin{minipage}[t]{0.30\columnwidth}\raggedright
input feature\strut
\end{minipage} & \begin{minipage}[t]{0.30\columnwidth}\raggedright
an observable property of the text, e.g., word choice, first-person
pronouns, length\strut
\end{minipage}\tabularnewline
\begin{minipage}[t]{0.30\columnwidth}\raggedright
ecological (cue) validity†\strut
\end{minipage} & \begin{minipage}[t]{0.30\columnwidth}\raggedright
feature--label correlation\strut
\end{minipage} & \begin{minipage}[t]{0.30\columnwidth}\raggedright
correlation of a text property with the ground truth\strut
\end{minipage}\tabularnewline
\begin{minipage}[t]{0.30\columnwidth}\raggedright
environmental predictability (R\textsubscript{e})\strut
\end{minipage} & \begin{minipage}[t]{0.30\columnwidth}\raggedright
performance of a (linear) model on the features\strut
\end{minipage} & \begin{minipage}[t]{0.30\columnwidth}\raggedright
how well the cues jointly predict the ground truth\strut
\end{minipage}\tabularnewline
\begin{minipage}[t]{0.30\columnwidth}\raggedright
representative design (Brunswik)\strut
\end{minipage} & \begin{minipage}[t]{0.30\columnwidth}\raggedright
natural, in-distribution test data\strut
\end{minipage} & \begin{minipage}[t]{0.30\columnwidth}\raggedright
real texts rather than constructed stimuli\strut
\end{minipage}\tabularnewline
\begin{minipage}[t]{0.30\columnwidth}\raggedright
\emph{Signal detection}\strut
\end{minipage} & \begin{minipage}[t]{0.30\columnwidth}\raggedright
\strut
\end{minipage} & \begin{minipage}[t]{0.30\columnwidth}\raggedright
\strut
\end{minipage}\tabularnewline
\begin{minipage}[t]{0.30\columnwidth}\raggedright
sensitivity d′†\strut
\end{minipage} & \begin{minipage}[t]{0.30\columnwidth}\raggedright
separability; AUC (d′ = √2 · z(AUC))\strut
\end{minipage} & \begin{minipage}[t]{0.30\columnwidth}\raggedright
how well the answers separate the two classes\strut
\end{minipage}\tabularnewline
\begin{minipage}[t]{0.30\columnwidth}\raggedright
response criterion c, response bias†\strut
\end{minipage} & \begin{minipage}[t]{0.30\columnwidth}\raggedright
decision threshold; prior or calibration offset\strut
\end{minipage} & \begin{minipage}[t]{0.30\columnwidth}\raggedright
tendency to give one answer rather than the other, e.g., ``human''\strut
\end{minipage}\tabularnewline
\begin{minipage}[t]{0.30\columnwidth}\raggedright
hit rate, false-alarm rate\strut
\end{minipage} & \begin{minipage}[t]{0.30\columnwidth}\raggedright
true-positive rate (recall), false-positive rate\strut
\end{minipage} & \begin{minipage}[t]{0.30\columnwidth}\raggedright
as in signal detection theory\strut
\end{minipage}\tabularnewline
\begin{minipage}[t]{0.30\columnwidth}\raggedright
ROC curve†\strut
\end{minipage} & \begin{minipage}[t]{0.30\columnwidth}\raggedright
ROC curve\strut
\end{minipage} & \begin{minipage}[t]{0.30\columnwidth}\raggedright
identical\strut
\end{minipage}\tabularnewline
\begin{minipage}[t]{0.30\columnwidth}\raggedright
\emph{Lens 1: encoding (internal representation)}\strut
\end{minipage} & \begin{minipage}[t]{0.30\columnwidth}\raggedright
\strut
\end{minipage} & \begin{minipage}[t]{0.30\columnwidth}\raggedright
\strut
\end{minipage}\tabularnewline
\begin{minipage}[t]{0.30\columnwidth}\raggedright
internal representation\strut
\end{minipage} & \begin{minipage}[t]{0.30\columnwidth}\raggedright
hidden states, activations, residual stream†\strut
\end{minipage} & \begin{minipage}[t]{0.30\columnwidth}\raggedright
the model's internal vectors at a given layer and position\strut
\end{minipage}\tabularnewline
\begin{minipage}[t]{0.30\columnwidth}\raggedright
decoding analysis, multivariate pattern analysis (neuroscience)\strut
\end{minipage} & \begin{minipage}[t]{0.30\columnwidth}\raggedright
linear probe†\strut
\end{minipage} & \begin{minipage}[t]{0.30\columnwidth}\raggedright
logistic regression that predicts the ground truth from
activations\strut
\end{minipage}\tabularnewline
\begin{minipage}[t]{0.30\columnwidth}\raggedright
representational similarity analysis\strut
\end{minipage} & \begin{minipage}[t]{0.30\columnwidth}\raggedright
centered kernel alignment (CKA)†\strut
\end{minipage} & \begin{minipage}[t]{0.30\columnwidth}\raggedright
similarity of the representations of two checkpoints\strut
\end{minipage}\tabularnewline
\begin{minipage}[t]{0.30\columnwidth}\raggedright
cross-decoding, generalization across conditions\strut
\end{minipage} & \begin{minipage}[t]{0.30\columnwidth}\raggedright
probe transfer†\strut
\end{minipage} & \begin{minipage}[t]{0.30\columnwidth}\raggedright
a probe fitted on one checkpoint (or set) and tested on another\strut
\end{minipage}\tabularnewline
\begin{minipage}[t]{0.30\columnwidth}\raggedright
\emph{Lens 2: readout (utilization and answer)}\strut
\end{minipage} & \begin{minipage}[t]{0.30\columnwidth}\raggedright
\strut
\end{minipage} & \begin{minipage}[t]{0.30\columnwidth}\raggedright
\strut
\end{minipage}\tabularnewline
\begin{minipage}[t]{0.30\columnwidth}\raggedright
task set, instruction\strut
\end{minipage} & \begin{minipage}[t]{0.30\columnwidth}\raggedright
prompt; system prompt; question†\strut
\end{minipage} & \begin{minipage}[t]{0.30\columnwidth}\raggedright
the question that asks for the judgment\strut
\end{minipage}\tabularnewline
\begin{minipage}[t]{0.30\columnwidth}\raggedright
cue utilization (r\textsubscript{s}, β\textsubscript{s})†\strut
\end{minipage} & \begin{minipage}[t]{0.30\columnwidth}\raggedright
feature attribution; feature importance\strut
\end{minipage} & \begin{minipage}[t]{0.30\columnwidth}\raggedright
how strongly a cue influences the answer\strut
\end{minipage}\tabularnewline
\begin{minipage}[t]{0.30\columnwidth}\raggedright
policy capturing, judgment analysis\strut
\end{minipage} & \begin{minipage}[t]{0.30\columnwidth}\raggedright
behavioral surrogate model\strut
\end{minipage} & \begin{minipage}[t]{0.30\columnwidth}\raggedright
regression of the answers on the cues\strut
\end{minipage}\tabularnewline
\begin{minipage}[t]{0.30\columnwidth}\raggedright
matching index G (lens model equation)†\strut
\end{minipage} & \begin{minipage}[t]{0.30\columnwidth}\raggedright
cosine similarity of weight vectors; agreement of decision rules\strut
\end{minipage} & \begin{minipage}[t]{0.30\columnwidth}\raggedright
whether the answer weights the cues as the environment does\strut
\end{minipage}\tabularnewline
\begin{minipage}[t]{0.30\columnwidth}\raggedright
achievement r\textsubscript{a}†\strut
\end{minipage} & \begin{minipage}[t]{0.30\columnwidth}\raggedright
accuracy; correlation of predictions with labels\strut
\end{minipage} & \begin{minipage}[t]{0.30\columnwidth}\raggedright
how well the answers track the ground truth\strut
\end{minipage}\tabularnewline
\begin{minipage}[t]{0.30\columnwidth}\raggedright
practice trials, worked examples\strut
\end{minipage} & \begin{minipage}[t]{0.30\columnwidth}\raggedright
few-shot examples, in-context learning†\strut
\end{minipage} & \begin{minipage}[t]{0.30\columnwidth}\raggedright
labeled examples given in the prompt\strut
\end{minipage}\tabularnewline
\begin{minipage}[t]{0.30\columnwidth}\raggedright
\emph{Causal tests and learning history}\strut
\end{minipage} & \begin{minipage}[t]{0.30\columnwidth}\raggedright
\strut
\end{minipage} & \begin{minipage}[t]{0.30\columnwidth}\raggedright
\strut
\end{minipage}\tabularnewline
\begin{minipage}[t]{0.30\columnwidth}\raggedright
experimental manipulation; microstimulation (neurophysiology)\strut
\end{minipage} & \begin{minipage}[t]{0.30\columnwidth}\raggedright
activation steering†\strut
\end{minipage} & \begin{minipage}[t]{0.30\columnwidth}\raggedright
shifting the internal state along a direction and observing the
answer\strut
\end{minipage}\tabularnewline
\begin{minipage}[t]{0.30\columnwidth}\raggedright
transplantation, lesion and rescue\strut
\end{minipage} & \begin{minipage}[t]{0.30\columnwidth}\raggedright
model stitching†; activation patching†\strut
\end{minipage} & \begin{minipage}[t]{0.30\columnwidth}\raggedright
combining layers or internal states of two checkpoints\strut
\end{minipage}\tabularnewline
\begin{minipage}[t]{0.30\columnwidth}\raggedright
developmental or training stages; longitudinal measurement\strut
\end{minipage} & \begin{minipage}[t]{0.30\columnwidth}\raggedright
checkpoints†; pretraining, supervised fine-tuning, preference
optimization†\strut
\end{minipage} & \begin{minipage}[t]{0.30\columnwidth}\raggedright
the published states of a model after each stage of training\strut
\end{minipage}\tabularnewline
\begin{minipage}[t]{0.30\columnwidth}\raggedright
\emph{General statistics}\strut
\end{minipage} & \begin{minipage}[t]{0.30\columnwidth}\raggedright
\strut
\end{minipage} & \begin{minipage}[t]{0.30\columnwidth}\raggedright
\strut
\end{minipage}\tabularnewline
\begin{minipage}[t]{0.30\columnwidth}\raggedright
fitting, estimation\strut
\end{minipage} & \begin{minipage}[t]{0.30\columnwidth}\raggedright
training, learning\strut
\end{minipage} & \begin{minipage}[t]{0.30\columnwidth}\raggedright
\strut
\end{minipage}\tabularnewline
\begin{minipage}[t]{0.30\columnwidth}\raggedright
predictor, independent variable\strut
\end{minipage} & \begin{minipage}[t]{0.30\columnwidth}\raggedright
feature, input\strut
\end{minipage} & \begin{minipage}[t]{0.30\columnwidth}\raggedright
\strut
\end{minipage}\tabularnewline
\begin{minipage}[t]{0.30\columnwidth}\raggedright
outcome, dependent variable\strut
\end{minipage} & \begin{minipage}[t]{0.30\columnwidth}\raggedright
target, label\strut
\end{minipage} & \begin{minipage}[t]{0.30\columnwidth}\raggedright
\strut
\end{minipage}\tabularnewline
\begin{minipage}[t]{0.30\columnwidth}\raggedright
regression coefficient; intercept\strut
\end{minipage} & \begin{minipage}[t]{0.30\columnwidth}\raggedright
weight; bias term\strut
\end{minipage} & \begin{minipage}[t]{0.30\columnwidth}\raggedright
\strut
\end{minipage}\tabularnewline
\begin{minipage}[t]{0.30\columnwidth}\raggedright
ridge (penalized) regression\strut
\end{minipage} & \begin{minipage}[t]{0.30\columnwidth}\raggedright
L2 regularization, weight decay\strut
\end{minipage} & \begin{minipage}[t]{0.30\columnwidth}\raggedright
used in the probes\strut
\end{minipage}\tabularnewline
\begin{minipage}[t]{0.30\columnwidth}\raggedright
clustered (grouped) resampling\strut
\end{minipage} & \begin{minipage}[t]{0.30\columnwidth}\raggedright
grouped cross-validation; avoiding leakage†\strut
\end{minipage} & \begin{minipage}[t]{0.30\columnwidth}\raggedright
texts from the same source never in training and test, or resampled
together\strut
\end{minipage}\tabularnewline
\begin{minipage}[t]{0.30\columnwidth}\raggedright
random-effects meta-analysis†\strut
\end{minipage} & \begin{minipage}[t]{0.30\columnwidth}\raggedright
---\strut
\end{minipage} & \begin{minipage}[t]{0.30\columnwidth}\raggedright
summary of the post-training contrasts across models and corpora\strut
\end{minipage}\tabularnewline
\bottomrule
\end{longtable}

\textbf{Table A2. False friends: terms with different meanings in the
two fields, and how we use them.}

\begin{longtable}[]{@{}llll@{}}
\toprule
\begin{minipage}[b]{0.22\columnwidth}\raggedright
Term\strut
\end{minipage} & \begin{minipage}[b]{0.22\columnwidth}\raggedright
Psychology / statistics\strut
\end{minipage} & \begin{minipage}[b]{0.22\columnwidth}\raggedright
Machine learning / interpretability\strut
\end{minipage} & \begin{minipage}[b]{0.22\columnwidth}\raggedright
Use in this paper\strut
\end{minipage}\tabularnewline
\midrule
\endhead
\begin{minipage}[t]{0.22\columnwidth}\raggedright
criterion\strut
\end{minipage} & \begin{minipage}[t]{0.22\columnwidth}\raggedright
response criterion c (signal detection); distal criterion
(Brunswik)\strut
\end{minipage} & \begin{minipage}[t]{0.22\columnwidth}\raggedright
loss function (e.g., \texttt{criterion} in PyTorch code)\strut
\end{minipage} & \begin{minipage}[t]{0.22\columnwidth}\raggedright
only the response criterion; the distal variable is called the state of
the world (ground truth)\strut
\end{minipage}\tabularnewline
\begin{minipage}[t]{0.22\columnwidth}\raggedright
bias\strut
\end{minipage} & \begin{minipage}[t]{0.22\columnwidth}\raggedright
response bias (signal detection); social bias\strut
\end{minipage} & \begin{minipage}[t]{0.22\columnwidth}\raggedright
bias term (intercept) of a layer; social bias, fairness\strut
\end{minipage} & \begin{minipage}[t]{0.22\columnwidth}\raggedright
``response bias'' for the signal-detection sense\strut
\end{minipage}\tabularnewline
\begin{minipage}[t]{0.22\columnwidth}\raggedright
sensitivity\strut
\end{minipage} & \begin{minipage}[t]{0.22\columnwidth}\raggedright
d′, discrimination ability; also ``sensitivity analysis'' (robustness
check of an analysis)\strut
\end{minipage} & \begin{minipage}[t]{0.22\columnwidth}\raggedright
true-positive rate (recall)\strut
\end{minipage} & \begin{minipage}[t]{0.22\columnwidth}\raggedright
d′; ``sensitivity analysis'' only in the statistical sense, where
unambiguous\strut
\end{minipage}\tabularnewline
\begin{minipage}[t]{0.22\columnwidth}\raggedright
validity\strut
\end{minipage} & \begin{minipage}[t]{0.22\columnwidth}\raggedright
ecological, construct, or predictive validity\strut
\end{minipage} & \begin{minipage}[t]{0.22\columnwidth}\raggedright
validation set\strut
\end{minipage} & \begin{minipage}[t]{0.22\columnwidth}\raggedright
cue validity: correlation of a cue with the ground truth\strut
\end{minipage}\tabularnewline
\begin{minipage}[t]{0.22\columnwidth}\raggedright
alignment\strut
\end{minipage} & \begin{minipage}[t]{0.22\columnwidth}\raggedright
--- (lens model: matching, G)\strut
\end{minipage} & \begin{minipage}[t]{0.22\columnwidth}\raggedright
AI alignment (post-training toward human preferences); representational
alignment\strut
\end{minipage} & \begin{minipage}[t]{0.22\columnwidth}\raggedright
``post-training'' for the training stages; ``matching'' for G; fixed
names kept (superficial alignment hypothesis, centered kernel
alignment)\strut
\end{minipage}\tabularnewline
\begin{minipage}[t]{0.22\columnwidth}\raggedright
attention\strut
\end{minipage} & \begin{minipage}[t]{0.22\columnwidth}\raggedright
selective attention\strut
\end{minipage} & \begin{minipage}[t]{0.22\columnwidth}\raggedright
attention mechanism of the transformer\strut
\end{minipage} & \begin{minipage}[t]{0.22\columnwidth}\raggedright
only the transformer mechanism; ``selection'' for the psychological
sense\strut
\end{minipage}\tabularnewline
\begin{minipage}[t]{0.22\columnwidth}\raggedright
representation\strut
\end{minipage} & \begin{minipage}[t]{0.22\columnwidth}\raggedright
mental representation\strut
\end{minipage} & \begin{minipage}[t]{0.22\columnwidth}\raggedright
vector in activation space\strut
\end{minipage} & \begin{minipage}[t]{0.22\columnwidth}\raggedright
activations, read with probes\strut
\end{minipage}\tabularnewline
\begin{minipage}[t]{0.22\columnwidth}\raggedright
probe\strut
\end{minipage} & \begin{minipage}[t]{0.22\columnwidth}\raggedright
probe stimulus, probe trial\strut
\end{minipage} & \begin{minipage}[t]{0.22\columnwidth}\raggedright
linear classifier on activations\strut
\end{minipage} & \begin{minipage}[t]{0.22\columnwidth}\raggedright
the classifier\strut
\end{minipage}\tabularnewline
\begin{minipage}[t]{0.22\columnwidth}\raggedright
policy\strut
\end{minipage} & \begin{minipage}[t]{0.22\columnwidth}\raggedright
judgment policy (policy capturing)\strut
\end{minipage} & \begin{minipage}[t]{0.22\columnwidth}\raggedright
policy π of a reinforcement-learning agent\strut
\end{minipage} & \begin{minipage}[t]{0.22\columnwidth}\raggedright
``cue weighting'' or ``judgment policy'', defined where used\strut
\end{minipage}\tabularnewline
\begin{minipage}[t]{0.22\columnwidth}\raggedright
model\strut
\end{minipage} & \begin{minipage}[t]{0.22\columnwidth}\raggedright
statistical or computational model of a judge\strut
\end{minipage} & \begin{minipage}[t]{0.22\columnwidth}\raggedright
the neural network itself\strut
\end{minipage} & \begin{minipage}[t]{0.22\columnwidth}\raggedright
``model'' for the language model; statistical models named
explicitly\strut
\end{minipage}\tabularnewline
\begin{minipage}[t]{0.22\columnwidth}\raggedright
feature\strut
\end{minipage} & \begin{minipage}[t]{0.22\columnwidth}\raggedright
stimulus feature\strut
\end{minipage} & \begin{minipage}[t]{0.22\columnwidth}\raggedright
input feature; learned feature (e.g., of a sparse autoencoder)\strut
\end{minipage} & \begin{minipage}[t]{0.22\columnwidth}\raggedright
text properties are ``cues''\strut
\end{minipage}\tabularnewline
\begin{minipage}[t]{0.22\columnwidth}\raggedright
readout\strut
\end{minipage} & \begin{minipage}[t]{0.22\columnwidth}\raggedright
read-out of a neural population by a downstream area\strut
\end{minipage} & \begin{minipage}[t]{0.22\columnwidth}\raggedright
output layer; decoding from activations\strut
\end{minipage} & \begin{minipage}[t]{0.22\columnwidth}\raggedright
how the question turns the representation into an answer\strut
\end{minipage}\tabularnewline
\bottomrule
\end{longtable}

\hypertarget{b-lens-model-analyses}{%
\subsection{B Lens-model analyses}\label{b-lens-model-analyses}}

\emph{Method.} For each judge, i.e., the answers of a checkpoint or the
out-of-fold scores of a probe on its representation, we computed the
components of the lens model equation (Karelaia \& Hogarth,
\protect\hyperlink{ref-karelaia2008lens}{2008}; Tucker,
\protect\hyperlink{ref-tucker1964lme}{1964}) from linear regressions of
the ground truth and of the judgments on a set of interpretable cues:
the environmental predictability R\textsubscript{e}, the consistency of
the judge R\textsubscript{s}, the matching index G (the correlation of
the two predictions), the unmodeled agreement C, and the achievement
r\textsubscript{a}, which decomposes as r\textsubscript{a} = G
R\textsubscript{e} R\textsubscript{s} + C √(1 − R\_e²) √(1 − R\_s²).
Because the regressions are fitted in sample, we also report
cross-validated R\textsubscript{e}, R\textsubscript{s} and G (10-fold,
grouped by source). For the short chats, the cues were 34 text
properties from five groups (length and lexical diversity, readability,
syntax, cohesion, and word frequency) and properties that the students
had named as typical of AI text (e.g., dashes, capitalization), computed
with a feature extractor written for the lab course; for the reviews, 15
cues motivated by Ott et al.
(\protect\hyperlink{ref-ott2011finding}{2011}), such as the rates of
first-person pronouns, of references to family and to the hotel or city,
of spatial details and of emotion words, and the length of the review
and of its words and sentences, all centered within the positive and the
negative reviews.

\emph{Short chats.} On the 300 messages for which probes were available,
the cues predicted authorship moderately well (R\textsubscript{e} = .56;
cross-validated .40). The answers of most post-trained models weighted
the cues in the direction opposite to their validity (cross-validated G:
Gemma 3 4B −.49, Gemma 3 12B −.58, Qwen3 −.61, Apertus after SFT −.57);
for Qwen3, this was already the case in the base model (−.53), whereas
the other base models were closer to zero (−.28 to −.04). The AI
messages in this corpus were richer in adverbs and verbs, lexically more
diverse, and contained longer sentences than the human messages (r = .16
to .23 with authorship), whereas the human messages contained more
adjectives, proper nouns, and long words; the post-trained models of
Gemma 3 12B, Qwen3, and Apertus took long words, adjectives, nouns, and
low readability as signs of AI authorship (e.g., Gemma 3 12B, mean word
length: r = −.04 before and +.24 after post-training; adjectives: −.06
and +.27). OLMo 2 showed the opposite weighting after SFT (G = +.37).
Probes weighted the cues in line with their validity in every checkpoint
(cross-validated G = .69--.78); because probes are trained on the ground
truth, this is expected by construction and only shows that the cues
were represented. In-sample values and the 422-message set including the
style-matched messages gave the same picture.

\emph{Deceptive reviews.} Fifteen cues predicted the ground truth
moderately well (R\textsubscript{e} = .56; cross-validated .55). Probes
again matched the environment closely in all 13 checkpoints of the four
models analyzed (G = .97--.99), as expected by construction. The answers
were poorly matched (G = .15--.83), and post-training changed this
matching in opposite directions: from .83 to .26 in Gemma 3 4B and from
.17 to .53 in Qwen3. After post-training, all models treated long,
detailed reviews as truthful, a cue without validity (r = .03 with the
ground truth), and ignored first-person pronouns, the most valid single
cue (r = −.33).

\hypertarget{c-deceptive-reviews}{%
\subsection{C Deceptive reviews}\label{c-deceptive-reviews}}

\textbf{Table C1.} Deceptive reviews: sensitivity (d′) of the probes
(last prompt position, two thirds of the network depth) and of the
answers (plain prompt). SFT: answers after supervised fine-tuning; ---:
no intermediate checkpoint released.

\begin{longtable}[]{@{}llrrrrr@{}}
\toprule
Judgment & Model & Probe base & Probe final & Answer base & SFT &
Final\tabularnewline
\midrule
\endhead
Authenticity & OLMo 2 7B & 2.93 & 2.79 & 0.16 & 0.20 &
0.17\tabularnewline
(hotel reviews) & Apertus 1.5 8B & 2.62 & 2.49 & 0.14 & 0.32 &
0.23\tabularnewline
& OLMo 3 7B Think & 2.80 & 2.69 & 0.13 & −0.01 & 0.00\tabularnewline
& Gemma 3 4B & 2.82 & 2.51 & 0.45 & --- & 0.08\tabularnewline
& Gemma 3 12B & 3.01 & 2.57 & 0.52 & --- & 0.86\tabularnewline
& Qwen3 8B & 2.56 & 2.59 & 0.06 & --- & 0.50\tabularnewline
\bottomrule
\end{longtable}

\emph{Answers.} All models judged the reviews close to chance (d′
between 0.00 and 0.86), although probes separated truthful from
deceptive reviews well (d′ = 2.5--3.0), at least as well as a classifier
on word frequencies (d′ = 2.44). Post-training changed the probes little
and the answers by small amounts in a direction that depended on the
model: Gemma 3 4B lost sensitivity (d′ 0.45 → 0.08), Qwen3 (0.06 → 0.50)
and Gemma 3 12B (0.52 → 0.86) gained, and the others changed little
(Table C1). What post-training changed consistently was a shared
heuristic. The base models judged positive reviews as more truthful than
negative ones, and review polarity accounted for 41--90\% of the
variance of their judgments; post-training largely removed this
heuristic (2--32\%), except in OLMo 3 Think (76\%), and all final models
showed a strong bias toward ``Truthful'' (83--92\% of the answers). The
localization and steering analyses were less conclusive than for the
other two judgments: Patching located the change at the question
positions in two of four cases, and steering did not explain it (see
below). The reviews are thus a case in which post-training reshaped the
readout without improving it.

\emph{Convergence.} Within each review polarity, the item-level
judgments of the five models post-trained mainly for conversation became
more similar: The mean correlation between models rose from .36 (range
.16--.50) for the base models to .52 (.45--.58) for the final models;
OLMo 3 Think did not join this convergence (.18--.33 with the other
final models). Without separating the polarities, the correlations fell
(from .75 to .48), because the shared polarity heuristic of the base
models disappeared.

\emph{Localization.} Stitching and patching were run for Gemma 3 4B
(which lost sensitivity) and Qwen3 (which gained). In Qwen3, the gain
arose gradually over layers 2--18 in stitching; at layer 18, the base
model's question states removed the gain from the final model (0.53 →
0.17; text states: 0.29), whereas the final model's question states
raised the base model only partly (0.08 → 0.22; all positions: 0.49). In
Gemma 3 4B, the final model's question states lowered the base model
(0.53 → 0.23; text states: no effect), but the base model's states did
not restore the final model at any set of positions (≤ 0.25), and
stitching produced non-additive effects in the middle and upper layers
with strongly varying answer probability. Thus, the question positions
carried the change in two of the four directions tested.

\emph{Steering.} In none of the four cells did the review direction move
the answer more than the control directions (Table D1).

These results are less conclusive than those for authorship and negated
statements: The effects of post-training were small
(\textbar Δd′\textbar{} ≤ 0.44), their sign differed between models, and
neither localization nor steering explained them fully. We can only
speculate about the reason. Although judging whether a review is genuine
resembles judging whether a text was written by an AI on the surface,
the two differ in what the models bring to them: The cues that separate
truthful from deceptive reviews are subtle and specific to this corpus
(human readers, too, judge these reviews near chance (Ott et al.,
\protect\hyperlink{ref-ott2011finding}{2011})), the models' answers were
near chance before and after post-training, and we found no examples of
this judgment in the fine-tuning data. It is conceivable that
post-training is largely irrelevant to a judgment for which neither the
base model nor the training data provide a usable mapping from the
represented information to the answer; what it changed here was a
general heuristic (review polarity) and the response bias, not the use
of the valid cues.

\hypertarget{d-steering-at-the-answer-position}{%
\subsection{D Steering at the answer
position}\label{d-steering-at-the-answer-position}}

The probes and the answers could dissociate either because the answers
ignore the authorship information or because they use it with the wrong
sign. As a check on the second possibility, we shifted a model's
internal state at the position where the answer is produced along the
direction that separates AI from human texts (estimated on held-out
groups; Section 2.7), and measured how the answer changed. The primary
test used shifts of ±2 within-class standard deviations, which kept
48--77\% of the shifted texts within the range of the observed texts (±1
SD: 60--94\%). We compared the effect with 100 control directions per
cell, each drawn once and applied to all texts: 50 random directions
within the high-variance subspace of the activations and 50 differences
between random halves of the texts within each class. The controls took
either steps of the same length as the authorship direction
(norm-matched) or steps of two standard deviations along their own
direction (scale-matched), which kept them within the range of the texts
as often as the authorship direction (49--69\%). As a positive control,
we shifted along the direction that the model's own answers follow (the
decision direction). In 21 of 25 cells (model × corpus × layer,
including the truth judgments of Section 3.6 and the review judgments of
Appendix C), the decision direction moved the answer more than the
norm-matched controls (rank-based p ≤ .05), so that the test can detect
a direction that the answer reads; against the scale-matched controls,
this was the case in 16 of 25 cells. Where the positive control failed,
a null result for the authorship direction is not informative.

In six of the 17 authorship cells, the authorship direction moved the
answer more than the controls (rank-based p ≤ .06 against both control
families; with 100 controls, the smallest possible p is .01, so that no
cell would survive a correction for 17 tests), and in all six it moved
it the wrong way: Shifting a text toward ``AI'' made the model answer
``human'' more often. This was the case for the short chats in Gemma 3
12B (layer 31: −7.38 nats; rank-based p = .01 against both control
families), Gemma 3 4B (layer 17: −1.74; p = .04 norm-matched, .06
scale-matched), and Qwen3 (layers 24 and 30: −1.18 and −2.06; p ≤ .03),
and for the newspaper segments in Qwen3 (−0.43 and −0.85; p ≤ .04). In
these cells, the authorship direction pointed against the decision
direction (cosine −0.59 to −0.92), and the answers were clearly below
chance (Table 2). Note that these are the cells with the most negative
d′, and that this is expected: The authorship direction is the
difference between the class means at the position from which the answer
is read, so that, if the answer depends approximately linearly on this
state, shifting along it must reproduce the observed difference between
the classes. Across the 25 cells, the effect of a shift by one class
distance correlated with the observed difference (r = .91). The test
therefore does not provide evidence that is independent of the answers'
d′; it shows that the inverse ordering is carried by the class-mean
difference at the answer position, more than by typical directions. In
the remaining cells, the authorship direction did not move the answer
more than arbitrary directions: Gemma 3 4B on the newspaper segments
(−1.04 and −2.00 at layers 22 and 28; p = .26--.42), Apertus and OLMo 3
on the textbook summaries before and after SFT, and Gemma 3 12B on the
newspaper segments, on which its answers were at chance (+0.03). In some
of these cells, the positive control did not exceed the scale-matched
controls either (Gemma 3 12B on the newspaper segments, Gemma 3 4B at
layer 28, Apertus at layers 21 and 27 and OLMo 3 after SFT; Table D1),
so that their null results are weaker evidence. In the strongest cell
(Gemma 3 12B on the chats), the decision direction itself did not
clearly exceed the norm-matched controls (p = .069; scale-matched: p =
.01), and only 48\% of the shifted texts stayed within the observed
range. The norm-matched controls in the subspace of leading components
moved the state far outside the observed range in the Gemma 3 and Qwen3
cells, so that the scale-matched controls are the fairer comparison. In
summary, where the answers ordered the texts inversely, the class-mean
difference at the answer position carried this inverse ordering;
steering at earlier layers or at the text positions, which would be an
independent test, was not carried out.

A complementary, descriptive comparison concerns the size of the
effects. Shifting a text by one class-mean distance along the authorship
direction, i.e., from the average human text to the average AI text,
changed the answer in the same direction as the observed difference
between AI and human texts in all 16 cells in which the bootstrap
interval of this difference excluded zero, and in 23 of the 25 cells
overall (e.g., Qwen3 on the newspaper segments: −0.50 and −0.66 nats
vs.~an observed difference of −0.59). As stated above, this agreement is
expected if the answer depends approximately linearly on the internal
state; it is consistent with the patching results (Section 3.4).

Steering did not, however, reveal what post-training changes for
authorship. In the Apertus base model, shifting toward ``AI'' moved the
answer in the correct direction, increasingly so with depth (+0.08,
+0.21, and +0.35 nats at layers 16, 21, and 27), but not more than the
control directions (p ≥ .15), and after SFT the effect was close to zero
(−0.04 to +0.04). In OLMo 3 Think, neither the base nor the SFT model
used the direction (−0.01 and +0.09).\footnote{An earlier comparison, in
  which the SFT model received its chat template and the base model the
  plain prompt, had suggested a reversed use after SFT (−0.27); with the
  same plain prompt for both checkpoints, this was not the case.} The
change of the answers during SFT is therefore localized by the stitching
and patching analyses (Section 3.4), not by steering.

\textbf{Table D1.} Steering at the last prompt position (L: layer).
Dist.: distance between the class means along the class direction, in
within-class SD. In range: share of steered texts (±2 SD) within the
1st--99th percentile of the observed projections. Effect: directional
effect on dv (nats) of ±2 SD along the class direction (authorship:
toward ``AI''; truth and reviews: toward ``true''). p: rank among 100
fixed control directions (norm-matched and scale-matched; minimum .01).
Decision: effect of the decision direction (positive control). Per
dist.: effect of one class distance, extrapolated from ±1 SD. Observed:
difference of the unsteered dv between the classes (AI − human, true −
false). fin.: final post-trained model.

\begin{longtable}[]{@{}llrrrrrrrrr@{}}
\toprule
\begin{minipage}[b]{0.12\columnwidth}\raggedright
Judgment\strut
\end{minipage} & \begin{minipage}[b]{0.23\columnwidth}\raggedright
Model\strut
\end{minipage} & \begin{minipage}[b]{0.04\columnwidth}\raggedleft
L\strut
\end{minipage} & \begin{minipage}[b]{0.04\columnwidth}\raggedleft
Dist.\strut
\end{minipage} & \begin{minipage}[b]{0.04\columnwidth}\raggedleft
In range\strut
\end{minipage} & \begin{minipage}[b]{0.04\columnwidth}\raggedleft
Effect\strut
\end{minipage} & \begin{minipage}[b]{0.04\columnwidth}\raggedleft
p (norm)\strut
\end{minipage} & \begin{minipage}[b]{0.04\columnwidth}\raggedleft
p (scale)\strut
\end{minipage} & \begin{minipage}[b]{0.04\columnwidth}\raggedleft
Decision\strut
\end{minipage} & \begin{minipage}[b]{0.04\columnwidth}\raggedleft
Per dist.\strut
\end{minipage} & \begin{minipage}[b]{0.04\columnwidth}\raggedleft
Observed\strut
\end{minipage}\tabularnewline
\midrule
\endhead
\begin{minipage}[t]{0.12\columnwidth}\raggedright
Chats\strut
\end{minipage} & \begin{minipage}[t]{0.23\columnwidth}\raggedright
Gemma 3 12B (fin.)\strut
\end{minipage} & \begin{minipage}[t]{0.04\columnwidth}\raggedleft
31\strut
\end{minipage} & \begin{minipage}[t]{0.04\columnwidth}\raggedleft
0.9\strut
\end{minipage} & \begin{minipage}[t]{0.04\columnwidth}\raggedleft
0.48\strut
\end{minipage} & \begin{minipage}[t]{0.04\columnwidth}\raggedleft
−7.38\strut
\end{minipage} & \begin{minipage}[t]{0.04\columnwidth}\raggedleft
.01\strut
\end{minipage} & \begin{minipage}[t]{0.04\columnwidth}\raggedleft
.01\strut
\end{minipage} & \begin{minipage}[t]{0.04\columnwidth}\raggedleft
+5.77\strut
\end{minipage} & \begin{minipage}[t]{0.04\columnwidth}\raggedleft
−2.96\strut
\end{minipage} & \begin{minipage}[t]{0.04\columnwidth}\raggedleft
−4.70\strut
\end{minipage}\tabularnewline
\begin{minipage}[t]{0.12\columnwidth}\raggedright
Chats\strut
\end{minipage} & \begin{minipage}[t]{0.23\columnwidth}\raggedright
Gemma 3 4B (fin.)\strut
\end{minipage} & \begin{minipage}[t]{0.04\columnwidth}\raggedleft
17\strut
\end{minipage} & \begin{minipage}[t]{0.04\columnwidth}\raggedleft
0.8\strut
\end{minipage} & \begin{minipage}[t]{0.04\columnwidth}\raggedleft
0.62\strut
\end{minipage} & \begin{minipage}[t]{0.04\columnwidth}\raggedleft
−1.74\strut
\end{minipage} & \begin{minipage}[t]{0.04\columnwidth}\raggedleft
.04\strut
\end{minipage} & \begin{minipage}[t]{0.04\columnwidth}\raggedleft
.06\strut
\end{minipage} & \begin{minipage}[t]{0.04\columnwidth}\raggedleft
+3.08\strut
\end{minipage} & \begin{minipage}[t]{0.04\columnwidth}\raggedleft
−0.65\strut
\end{minipage} & \begin{minipage}[t]{0.04\columnwidth}\raggedleft
−3.33\strut
\end{minipage}\tabularnewline
\begin{minipage}[t]{0.12\columnwidth}\raggedright
Chats\strut
\end{minipage} & \begin{minipage}[t]{0.23\columnwidth}\raggedright
Qwen3 8B (fin.)\strut
\end{minipage} & \begin{minipage}[t]{0.04\columnwidth}\raggedleft
24\strut
\end{minipage} & \begin{minipage}[t]{0.04\columnwidth}\raggedleft
1.2\strut
\end{minipage} & \begin{minipage}[t]{0.04\columnwidth}\raggedleft
0.63\strut
\end{minipage} & \begin{minipage}[t]{0.04\columnwidth}\raggedleft
−1.18\strut
\end{minipage} & \begin{minipage}[t]{0.04\columnwidth}\raggedleft
.02\strut
\end{minipage} & \begin{minipage}[t]{0.04\columnwidth}\raggedleft
.01\strut
\end{minipage} & \begin{minipage}[t]{0.04\columnwidth}\raggedleft
+1.52\strut
\end{minipage} & \begin{minipage}[t]{0.04\columnwidth}\raggedleft
−0.72\strut
\end{minipage} & \begin{minipage}[t]{0.04\columnwidth}\raggedleft
−1.23\strut
\end{minipage}\tabularnewline
\begin{minipage}[t]{0.12\columnwidth}\raggedright
Chats\strut
\end{minipage} & \begin{minipage}[t]{0.23\columnwidth}\raggedright
Qwen3 8B (fin.)\strut
\end{minipage} & \begin{minipage}[t]{0.04\columnwidth}\raggedleft
30\strut
\end{minipage} & \begin{minipage}[t]{0.04\columnwidth}\raggedleft
0.9\strut
\end{minipage} & \begin{minipage}[t]{0.04\columnwidth}\raggedleft
0.60\strut
\end{minipage} & \begin{minipage}[t]{0.04\columnwidth}\raggedleft
−2.06\strut
\end{minipage} & \begin{minipage}[t]{0.04\columnwidth}\raggedleft
.03\strut
\end{minipage} & \begin{minipage}[t]{0.04\columnwidth}\raggedleft
.02\strut
\end{minipage} & \begin{minipage}[t]{0.04\columnwidth}\raggedleft
+2.30\strut
\end{minipage} & \begin{minipage}[t]{0.04\columnwidth}\raggedleft
−0.96\strut
\end{minipage} & \begin{minipage}[t]{0.04\columnwidth}\raggedleft
−1.23\strut
\end{minipage}\tabularnewline
\begin{minipage}[t]{0.12\columnwidth}\raggedright
News\strut
\end{minipage} & \begin{minipage}[t]{0.23\columnwidth}\raggedright
Gemma 3 12B (fin.)\strut
\end{minipage} & \begin{minipage}[t]{0.04\columnwidth}\raggedleft
31\strut
\end{minipage} & \begin{minipage}[t]{0.04\columnwidth}\raggedleft
0.7\strut
\end{minipage} & \begin{minipage}[t]{0.04\columnwidth}\raggedleft
0.54\strut
\end{minipage} & \begin{minipage}[t]{0.04\columnwidth}\raggedleft
+0.03\strut
\end{minipage} & \begin{minipage}[t]{0.04\columnwidth}\raggedleft
.97\strut
\end{minipage} & \begin{minipage}[t]{0.04\columnwidth}\raggedleft
.94\strut
\end{minipage} & \begin{minipage}[t]{0.04\columnwidth}\raggedleft
+1.36\strut
\end{minipage} & \begin{minipage}[t]{0.04\columnwidth}\raggedleft
+0.01\strut
\end{minipage} & \begin{minipage}[t]{0.04\columnwidth}\raggedleft
+0.14\strut
\end{minipage}\tabularnewline
\begin{minipage}[t]{0.12\columnwidth}\raggedright
News\strut
\end{minipage} & \begin{minipage}[t]{0.23\columnwidth}\raggedright
Gemma 3 4B (fin.)\strut
\end{minipage} & \begin{minipage}[t]{0.04\columnwidth}\raggedleft
22\strut
\end{minipage} & \begin{minipage}[t]{0.04\columnwidth}\raggedleft
0.8\strut
\end{minipage} & \begin{minipage}[t]{0.04\columnwidth}\raggedleft
0.60\strut
\end{minipage} & \begin{minipage}[t]{0.04\columnwidth}\raggedleft
−1.04\strut
\end{minipage} & \begin{minipage}[t]{0.04\columnwidth}\raggedleft
.42\strut
\end{minipage} & \begin{minipage}[t]{0.04\columnwidth}\raggedleft
.26\strut
\end{minipage} & \begin{minipage}[t]{0.04\columnwidth}\raggedleft
+3.83\strut
\end{minipage} & \begin{minipage}[t]{0.04\columnwidth}\raggedleft
−0.40\strut
\end{minipage} & \begin{minipage}[t]{0.04\columnwidth}\raggedleft
−0.40\strut
\end{minipage}\tabularnewline
\begin{minipage}[t]{0.12\columnwidth}\raggedright
News\strut
\end{minipage} & \begin{minipage}[t]{0.23\columnwidth}\raggedright
Gemma 3 4B (fin.)\strut
\end{minipage} & \begin{minipage}[t]{0.04\columnwidth}\raggedleft
28\strut
\end{minipage} & \begin{minipage}[t]{0.04\columnwidth}\raggedleft
0.5\strut
\end{minipage} & \begin{minipage}[t]{0.04\columnwidth}\raggedleft
0.51\strut
\end{minipage} & \begin{minipage}[t]{0.04\columnwidth}\raggedleft
−2.00\strut
\end{minipage} & \begin{minipage}[t]{0.04\columnwidth}\raggedleft
.37\strut
\end{minipage} & \begin{minipage}[t]{0.04\columnwidth}\raggedleft
.26\strut
\end{minipage} & \begin{minipage}[t]{0.04\columnwidth}\raggedleft
+3.60\strut
\end{minipage} & \begin{minipage}[t]{0.04\columnwidth}\raggedleft
−0.49\strut
\end{minipage} & \begin{minipage}[t]{0.04\columnwidth}\raggedleft
−0.40\strut
\end{minipage}\tabularnewline
\begin{minipage}[t]{0.12\columnwidth}\raggedright
News\strut
\end{minipage} & \begin{minipage}[t]{0.23\columnwidth}\raggedright
Qwen3 8B (fin.)\strut
\end{minipage} & \begin{minipage}[t]{0.04\columnwidth}\raggedleft
24\strut
\end{minipage} & \begin{minipage}[t]{0.04\columnwidth}\raggedleft
2.3\strut
\end{minipage} & \begin{minipage}[t]{0.04\columnwidth}\raggedleft
0.75\strut
\end{minipage} & \begin{minipage}[t]{0.04\columnwidth}\raggedleft
−0.43\strut
\end{minipage} & \begin{minipage}[t]{0.04\columnwidth}\raggedleft
.04\strut
\end{minipage} & \begin{minipage}[t]{0.04\columnwidth}\raggedleft
.03\strut
\end{minipage} & \begin{minipage}[t]{0.04\columnwidth}\raggedleft
+0.67\strut
\end{minipage} & \begin{minipage}[t]{0.04\columnwidth}\raggedleft
−0.50\strut
\end{minipage} & \begin{minipage}[t]{0.04\columnwidth}\raggedleft
−0.59\strut
\end{minipage}\tabularnewline
\begin{minipage}[t]{0.12\columnwidth}\raggedright
News\strut
\end{minipage} & \begin{minipage}[t]{0.23\columnwidth}\raggedright
Qwen3 8B (fin.)\strut
\end{minipage} & \begin{minipage}[t]{0.04\columnwidth}\raggedleft
30\strut
\end{minipage} & \begin{minipage}[t]{0.04\columnwidth}\raggedleft
1.6\strut
\end{minipage} & \begin{minipage}[t]{0.04\columnwidth}\raggedleft
0.65\strut
\end{minipage} & \begin{minipage}[t]{0.04\columnwidth}\raggedleft
−0.85\strut
\end{minipage} & \begin{minipage}[t]{0.04\columnwidth}\raggedleft
.03\strut
\end{minipage} & \begin{minipage}[t]{0.04\columnwidth}\raggedleft
.03\strut
\end{minipage} & \begin{minipage}[t]{0.04\columnwidth}\raggedleft
+1.05\strut
\end{minipage} & \begin{minipage}[t]{0.04\columnwidth}\raggedleft
−0.66\strut
\end{minipage} & \begin{minipage}[t]{0.04\columnwidth}\raggedleft
−0.59\strut
\end{minipage}\tabularnewline
\begin{minipage}[t]{0.12\columnwidth}\raggedright
Textbook\strut
\end{minipage} & \begin{minipage}[t]{0.23\columnwidth}\raggedright
Apertus 1.5 8B (base)\strut
\end{minipage} & \begin{minipage}[t]{0.04\columnwidth}\raggedleft
16\strut
\end{minipage} & \begin{minipage}[t]{0.04\columnwidth}\raggedleft
3.2\strut
\end{minipage} & \begin{minipage}[t]{0.04\columnwidth}\raggedleft
0.70\strut
\end{minipage} & \begin{minipage}[t]{0.04\columnwidth}\raggedleft
+0.08\strut
\end{minipage} & \begin{minipage}[t]{0.04\columnwidth}\raggedleft
.48\strut
\end{minipage} & \begin{minipage}[t]{0.04\columnwidth}\raggedleft
.45\strut
\end{minipage} & \begin{minipage}[t]{0.04\columnwidth}\raggedleft
+0.37\strut
\end{minipage} & \begin{minipage}[t]{0.04\columnwidth}\raggedleft
+0.11\strut
\end{minipage} & \begin{minipage}[t]{0.04\columnwidth}\raggedleft
+0.15\strut
\end{minipage}\tabularnewline
\begin{minipage}[t]{0.12\columnwidth}\raggedright
Textbook\strut
\end{minipage} & \begin{minipage}[t]{0.23\columnwidth}\raggedright
Apertus 1.5 8B (base)\strut
\end{minipage} & \begin{minipage}[t]{0.04\columnwidth}\raggedleft
21\strut
\end{minipage} & \begin{minipage}[t]{0.04\columnwidth}\raggedleft
2.2\strut
\end{minipage} & \begin{minipage}[t]{0.04\columnwidth}\raggedleft
0.67\strut
\end{minipage} & \begin{minipage}[t]{0.04\columnwidth}\raggedleft
+0.21\strut
\end{minipage} & \begin{minipage}[t]{0.04\columnwidth}\raggedleft
.37\strut
\end{minipage} & \begin{minipage}[t]{0.04\columnwidth}\raggedleft
.24\strut
\end{minipage} & \begin{minipage}[t]{0.04\columnwidth}\raggedleft
+0.57\strut
\end{minipage} & \begin{minipage}[t]{0.04\columnwidth}\raggedleft
+0.23\strut
\end{minipage} & \begin{minipage}[t]{0.04\columnwidth}\raggedleft
+0.15\strut
\end{minipage}\tabularnewline
\begin{minipage}[t]{0.12\columnwidth}\raggedright
Textbook\strut
\end{minipage} & \begin{minipage}[t]{0.23\columnwidth}\raggedright
Apertus 1.5 8B (base)\strut
\end{minipage} & \begin{minipage}[t]{0.04\columnwidth}\raggedleft
27\strut
\end{minipage} & \begin{minipage}[t]{0.04\columnwidth}\raggedleft
1.6\strut
\end{minipage} & \begin{minipage}[t]{0.04\columnwidth}\raggedleft
0.62\strut
\end{minipage} & \begin{minipage}[t]{0.04\columnwidth}\raggedleft
+0.35\strut
\end{minipage} & \begin{minipage}[t]{0.04\columnwidth}\raggedleft
.22\strut
\end{minipage} & \begin{minipage}[t]{0.04\columnwidth}\raggedleft
.15\strut
\end{minipage} & \begin{minipage}[t]{0.04\columnwidth}\raggedleft
+0.68\strut
\end{minipage} & \begin{minipage}[t]{0.04\columnwidth}\raggedleft
+0.28\strut
\end{minipage} & \begin{minipage}[t]{0.04\columnwidth}\raggedleft
+0.15\strut
\end{minipage}\tabularnewline
\begin{minipage}[t]{0.12\columnwidth}\raggedright
Textbook\strut
\end{minipage} & \begin{minipage}[t]{0.23\columnwidth}\raggedright
Apertus 1.5 8B (SFT)\strut
\end{minipage} & \begin{minipage}[t]{0.04\columnwidth}\raggedleft
16\strut
\end{minipage} & \begin{minipage}[t]{0.04\columnwidth}\raggedleft
3.0\strut
\end{minipage} & \begin{minipage}[t]{0.04\columnwidth}\raggedleft
0.74\strut
\end{minipage} & \begin{minipage}[t]{0.04\columnwidth}\raggedleft
+0.04\strut
\end{minipage} & \begin{minipage}[t]{0.04\columnwidth}\raggedleft
.67\strut
\end{minipage} & \begin{minipage}[t]{0.04\columnwidth}\raggedleft
.73\strut
\end{minipage} & \begin{minipage}[t]{0.04\columnwidth}\raggedleft
+0.32\strut
\end{minipage} & \begin{minipage}[t]{0.04\columnwidth}\raggedleft
+0.05\strut
\end{minipage} & \begin{minipage}[t]{0.04\columnwidth}\raggedleft
−0.00\strut
\end{minipage}\tabularnewline
\begin{minipage}[t]{0.12\columnwidth}\raggedright
Textbook\strut
\end{minipage} & \begin{minipage}[t]{0.23\columnwidth}\raggedright
Apertus 1.5 8B (SFT)\strut
\end{minipage} & \begin{minipage}[t]{0.04\columnwidth}\raggedleft
21\strut
\end{minipage} & \begin{minipage}[t]{0.04\columnwidth}\raggedleft
2.8\strut
\end{minipage} & \begin{minipage}[t]{0.04\columnwidth}\raggedleft
0.72\strut
\end{minipage} & \begin{minipage}[t]{0.04\columnwidth}\raggedleft
−0.03\strut
\end{minipage} & \begin{minipage}[t]{0.04\columnwidth}\raggedleft
.83\strut
\end{minipage} & \begin{minipage}[t]{0.04\columnwidth}\raggedleft
.87\strut
\end{minipage} & \begin{minipage}[t]{0.04\columnwidth}\raggedleft
+0.37\strut
\end{minipage} & \begin{minipage}[t]{0.04\columnwidth}\raggedleft
−0.04\strut
\end{minipage} & \begin{minipage}[t]{0.04\columnwidth}\raggedleft
−0.00\strut
\end{minipage}\tabularnewline
\begin{minipage}[t]{0.12\columnwidth}\raggedright
Textbook\strut
\end{minipage} & \begin{minipage}[t]{0.23\columnwidth}\raggedright
Apertus 1.5 8B (SFT)\strut
\end{minipage} & \begin{minipage}[t]{0.04\columnwidth}\raggedleft
27\strut
\end{minipage} & \begin{minipage}[t]{0.04\columnwidth}\raggedleft
2.5\strut
\end{minipage} & \begin{minipage}[t]{0.04\columnwidth}\raggedleft
0.77\strut
\end{minipage} & \begin{minipage}[t]{0.04\columnwidth}\raggedleft
−0.04\strut
\end{minipage} & \begin{minipage}[t]{0.04\columnwidth}\raggedleft
.74\strut
\end{minipage} & \begin{minipage}[t]{0.04\columnwidth}\raggedleft
.80\strut
\end{minipage} & \begin{minipage}[t]{0.04\columnwidth}\raggedleft
+0.36\strut
\end{minipage} & \begin{minipage}[t]{0.04\columnwidth}\raggedleft
−0.05\strut
\end{minipage} & \begin{minipage}[t]{0.04\columnwidth}\raggedleft
−0.00\strut
\end{minipage}\tabularnewline
\begin{minipage}[t]{0.12\columnwidth}\raggedright
Textbook\strut
\end{minipage} & \begin{minipage}[t]{0.23\columnwidth}\raggedright
OLMo 3 7B Think (base)\strut
\end{minipage} & \begin{minipage}[t]{0.04\columnwidth}\raggedleft
21\strut
\end{minipage} & \begin{minipage}[t]{0.04\columnwidth}\raggedleft
1.6\strut
\end{minipage} & \begin{minipage}[t]{0.04\columnwidth}\raggedleft
0.71\strut
\end{minipage} & \begin{minipage}[t]{0.04\columnwidth}\raggedleft
−0.01\strut
\end{minipage} & \begin{minipage}[t]{0.04\columnwidth}\raggedleft
.95\strut
\end{minipage} & \begin{minipage}[t]{0.04\columnwidth}\raggedleft
.93\strut
\end{minipage} & \begin{minipage}[t]{0.04\columnwidth}\raggedleft
+0.42\strut
\end{minipage} & \begin{minipage}[t]{0.04\columnwidth}\raggedleft
−0.01\strut
\end{minipage} & \begin{minipage}[t]{0.04\columnwidth}\raggedleft
−0.02\strut
\end{minipage}\tabularnewline
\begin{minipage}[t]{0.12\columnwidth}\raggedright
Textbook\strut
\end{minipage} & \begin{minipage}[t]{0.23\columnwidth}\raggedright
OLMo 3 7B Think (SFT)\strut
\end{minipage} & \begin{minipage}[t]{0.04\columnwidth}\raggedleft
21\strut
\end{minipage} & \begin{minipage}[t]{0.04\columnwidth}\raggedleft
2.4\strut
\end{minipage} & \begin{minipage}[t]{0.04\columnwidth}\raggedleft
0.75\strut
\end{minipage} & \begin{minipage}[t]{0.04\columnwidth}\raggedleft
+0.09\strut
\end{minipage} & \begin{minipage}[t]{0.04\columnwidth}\raggedleft
.28\strut
\end{minipage} & \begin{minipage}[t]{0.04\columnwidth}\raggedleft
.58\strut
\end{minipage} & \begin{minipage}[t]{0.04\columnwidth}\raggedleft
+0.15\strut
\end{minipage} & \begin{minipage}[t]{0.04\columnwidth}\raggedleft
+0.12\strut
\end{minipage} & \begin{minipage}[t]{0.04\columnwidth}\raggedleft
+0.14\strut
\end{minipage}\tabularnewline
\begin{minipage}[t]{0.12\columnwidth}\raggedright
Negated statements\strut
\end{minipage} & \begin{minipage}[t]{0.23\columnwidth}\raggedright
Apertus 1.5 8B (base)\strut
\end{minipage} & \begin{minipage}[t]{0.04\columnwidth}\raggedleft
21\strut
\end{minipage} & \begin{minipage}[t]{0.04\columnwidth}\raggedleft
7.9\strut
\end{minipage} & \begin{minipage}[t]{0.04\columnwidth}\raggedleft
0.64\strut
\end{minipage} & \begin{minipage}[t]{0.04\columnwidth}\raggedleft
−0.14\strut
\end{minipage} & \begin{minipage}[t]{0.04\columnwidth}\raggedleft
.13\strut
\end{minipage} & \begin{minipage}[t]{0.04\columnwidth}\raggedleft
.33\strut
\end{minipage} & \begin{minipage}[t]{0.04\columnwidth}\raggedleft
+0.27\strut
\end{minipage} & \begin{minipage}[t]{0.04\columnwidth}\raggedleft
−0.54\strut
\end{minipage} & \begin{minipage}[t]{0.04\columnwidth}\raggedleft
−0.31\strut
\end{minipage}\tabularnewline
\begin{minipage}[t]{0.12\columnwidth}\raggedright
Negated statements\strut
\end{minipage} & \begin{minipage}[t]{0.23\columnwidth}\raggedright
Apertus 1.5 8B (SFT)\strut
\end{minipage} & \begin{minipage}[t]{0.04\columnwidth}\raggedleft
21\strut
\end{minipage} & \begin{minipage}[t]{0.04\columnwidth}\raggedleft
14.2\strut
\end{minipage} & \begin{minipage}[t]{0.04\columnwidth}\raggedleft
0.58\strut
\end{minipage} & \begin{minipage}[t]{0.04\columnwidth}\raggedleft
+0.36\strut
\end{minipage} & \begin{minipage}[t]{0.04\columnwidth}\raggedleft
.02\strut
\end{minipage} & \begin{minipage}[t]{0.04\columnwidth}\raggedleft
.02\strut
\end{minipage} & \begin{minipage}[t]{0.04\columnwidth}\raggedleft
+0.36\strut
\end{minipage} & \begin{minipage}[t]{0.04\columnwidth}\raggedleft
+2.51\strut
\end{minipage} & \begin{minipage}[t]{0.04\columnwidth}\raggedleft
+3.08\strut
\end{minipage}\tabularnewline
\begin{minipage}[t]{0.12\columnwidth}\raggedright
Negated statements\strut
\end{minipage} & \begin{minipage}[t]{0.23\columnwidth}\raggedright
OLMo 2 7B (base)\strut
\end{minipage} & \begin{minipage}[t]{0.04\columnwidth}\raggedleft
21\strut
\end{minipage} & \begin{minipage}[t]{0.04\columnwidth}\raggedleft
9.5\strut
\end{minipage} & \begin{minipage}[t]{0.04\columnwidth}\raggedleft
0.57\strut
\end{minipage} & \begin{minipage}[t]{0.04\columnwidth}\raggedleft
+0.09\strut
\end{minipage} & \begin{minipage}[t]{0.04\columnwidth}\raggedleft
.58\strut
\end{minipage} & \begin{minipage}[t]{0.04\columnwidth}\raggedleft
.54\strut
\end{minipage} & \begin{minipage}[t]{0.04\columnwidth}\raggedleft
+0.42\strut
\end{minipage} & \begin{minipage}[t]{0.04\columnwidth}\raggedleft
+0.43\strut
\end{minipage} & \begin{minipage}[t]{0.04\columnwidth}\raggedleft
+0.22\strut
\end{minipage}\tabularnewline
\begin{minipage}[t]{0.12\columnwidth}\raggedright
Negated statements\strut
\end{minipage} & \begin{minipage}[t]{0.23\columnwidth}\raggedright
OLMo 2 7B (SFT)\strut
\end{minipage} & \begin{minipage}[t]{0.04\columnwidth}\raggedleft
21\strut
\end{minipage} & \begin{minipage}[t]{0.04\columnwidth}\raggedleft
7.7\strut
\end{minipage} & \begin{minipage}[t]{0.04\columnwidth}\raggedleft
0.59\strut
\end{minipage} & \begin{minipage}[t]{0.04\columnwidth}\raggedleft
+1.07\strut
\end{minipage} & \begin{minipage}[t]{0.04\columnwidth}\raggedleft
.01\strut
\end{minipage} & \begin{minipage}[t]{0.04\columnwidth}\raggedleft
.04\strut
\end{minipage} & \begin{minipage}[t]{0.04\columnwidth}\raggedleft
+1.14\strut
\end{minipage} & \begin{minipage}[t]{0.04\columnwidth}\raggedleft
+4.13\strut
\end{minipage} & \begin{minipage}[t]{0.04\columnwidth}\raggedleft
+3.64\strut
\end{minipage}\tabularnewline
\begin{minipage}[t]{0.12\columnwidth}\raggedright
Reviews\strut
\end{minipage} & \begin{minipage}[t]{0.23\columnwidth}\raggedright
Gemma 3 4B (base)\strut
\end{minipage} & \begin{minipage}[t]{0.04\columnwidth}\raggedleft
23\strut
\end{minipage} & \begin{minipage}[t]{0.04\columnwidth}\raggedleft
0.8\strut
\end{minipage} & \begin{minipage}[t]{0.04\columnwidth}\raggedleft
0.66\strut
\end{minipage} & \begin{minipage}[t]{0.04\columnwidth}\raggedleft
+0.21\strut
\end{minipage} & \begin{minipage}[t]{0.04\columnwidth}\raggedleft
.68\strut
\end{minipage} & \begin{minipage}[t]{0.04\columnwidth}\raggedleft
.35\strut
\end{minipage} & \begin{minipage}[t]{0.04\columnwidth}\raggedleft
+1.03\strut
\end{minipage} & \begin{minipage}[t]{0.04\columnwidth}\raggedleft
+0.09\strut
\end{minipage} & \begin{minipage}[t]{0.04\columnwidth}\raggedleft
+0.20\strut
\end{minipage}\tabularnewline
\begin{minipage}[t]{0.12\columnwidth}\raggedright
Reviews\strut
\end{minipage} & \begin{minipage}[t]{0.23\columnwidth}\raggedright
Gemma 3 4B (fin.)\strut
\end{minipage} & \begin{minipage}[t]{0.04\columnwidth}\raggedleft
23\strut
\end{minipage} & \begin{minipage}[t]{0.04\columnwidth}\raggedleft
0.7\strut
\end{minipage} & \begin{minipage}[t]{0.04\columnwidth}\raggedleft
0.60\strut
\end{minipage} & \begin{minipage}[t]{0.04\columnwidth}\raggedleft
+0.85\strut
\end{minipage} & \begin{minipage}[t]{0.04\columnwidth}\raggedleft
.66\strut
\end{minipage} & \begin{minipage}[t]{0.04\columnwidth}\raggedleft
.53\strut
\end{minipage} & \begin{minipage}[t]{0.04\columnwidth}\raggedleft
+4.45\strut
\end{minipage} & \begin{minipage}[t]{0.04\columnwidth}\raggedleft
+0.31\strut
\end{minipage} & \begin{minipage}[t]{0.04\columnwidth}\raggedleft
+0.55\strut
\end{minipage}\tabularnewline
\begin{minipage}[t]{0.12\columnwidth}\raggedright
Reviews\strut
\end{minipage} & \begin{minipage}[t]{0.23\columnwidth}\raggedright
Qwen3 8B (base)\strut
\end{minipage} & \begin{minipage}[t]{0.04\columnwidth}\raggedleft
24\strut
\end{minipage} & \begin{minipage}[t]{0.04\columnwidth}\raggedleft
1.3\strut
\end{minipage} & \begin{minipage}[t]{0.04\columnwidth}\raggedleft
0.70\strut
\end{minipage} & \begin{minipage}[t]{0.04\columnwidth}\raggedleft
−0.01\strut
\end{minipage} & \begin{minipage}[t]{0.04\columnwidth}\raggedleft
.98\strut
\end{minipage} & \begin{minipage}[t]{0.04\columnwidth}\raggedleft
.96\strut
\end{minipage} & \begin{minipage}[t]{0.04\columnwidth}\raggedleft
+0.53\strut
\end{minipage} & \begin{minipage}[t]{0.04\columnwidth}\raggedleft
−0.01\strut
\end{minipage} & \begin{minipage}[t]{0.04\columnwidth}\raggedleft
+0.06\strut
\end{minipage}\tabularnewline
\begin{minipage}[t]{0.12\columnwidth}\raggedright
Reviews\strut
\end{minipage} & \begin{minipage}[t]{0.23\columnwidth}\raggedright
Qwen3 8B (fin.)\strut
\end{minipage} & \begin{minipage}[t]{0.04\columnwidth}\raggedleft
24\strut
\end{minipage} & \begin{minipage}[t]{0.04\columnwidth}\raggedleft
1.2\strut
\end{minipage} & \begin{minipage}[t]{0.04\columnwidth}\raggedleft
0.68\strut
\end{minipage} & \begin{minipage}[t]{0.04\columnwidth}\raggedleft
+0.50\strut
\end{minipage} & \begin{minipage}[t]{0.04\columnwidth}\raggedleft
.16\strut
\end{minipage} & \begin{minipage}[t]{0.04\columnwidth}\raggedleft
.20\strut
\end{minipage} & \begin{minipage}[t]{0.04\columnwidth}\raggedleft
+1.08\strut
\end{minipage} & \begin{minipage}[t]{0.04\columnwidth}\raggedleft
+0.30\strut
\end{minipage} & \begin{minipage}[t]{0.04\columnwidth}\raggedleft
+1.08\strut
\end{minipage}\tabularnewline
\bottomrule
\end{longtable}

\hypertarget{e-few-shot-gains}{%
\subsection{E Few-shot gains}\label{e-few-shot-gains}}

\textbf{Table E1.} Gain in d′ from 0 to 16 labeled examples (mean of the
per-set d′ minus zero-shot d′), for base and final post-trained models
on the same texts, and the difference of the gains with a
grouped-bootstrap CI.

\begin{longtable}[]{@{}llrrr@{}}
\toprule
Corpus & Model & Gain base & Gain post-trained & Difference {[}95\%
CI{]}\tabularnewline
\midrule
\endhead
Chats & Gemma 3 4B & +0.69 & +0.45 & +0.24 {[}−0.09,
+0.60{]}\tabularnewline
Chats & OLMo 2 7B & +0.55 & +0.01 & +0.54 {[}+0.27,
+0.83{]}\tabularnewline
Chats & Qwen3 8B & +0.83 & +0.63 & +0.19 {[}−0.06,
+0.45{]}\tabularnewline
Textbook & Gemma 3 4B & +0.50 & +0.33 & +0.18 {[}−0.06,
+0.41{]}\tabularnewline
Textbook & Qwen3 8B & −0.13 & +0.18 & −0.31 {[}−0.55,
−0.09{]}\tabularnewline
\bottomrule
\end{longtable}

\hypertarget{f-signal-detection-invariance}{%
\subsection{F Signal detection
invariance}\label{f-signal-detection-invariance}}

\textbf{Table F1.} Mean absolute change of the criterion c and of the
sensitivity d′ under three manipulations of the plain prompt, averaged
over corpora and answer orders: stated AI share 20 → 80\% (from the
slope over 20, 50, and 80\%), answer order (``KI oder Mensch''
vs.~``Mensch oder KI'', without base-rate sentence), and presence of a
base-rate sentence (50\% vs.~none). The changes of d′ had no consistent
direction (mean signed change ≤ 0.03). In 56\% of the conditions, the
model gave the same answer to at least 98\% of the texts, so that c
could hardly move.

\begin{longtable}[]{@{}llrrrrrr@{}}
\toprule
\begin{minipage}[b]{0.10\columnwidth}\raggedright
Model\strut
\end{minipage} & \begin{minipage}[b]{0.10\columnwidth}\raggedright
Checkpoint\strut
\end{minipage} & \begin{minipage}[b]{0.10\columnwidth}\raggedleft
Base rate 20→80\%: c\strut
\end{minipage} & \begin{minipage}[b]{0.10\columnwidth}\raggedleft
Base rate 20→80\%: d′\strut
\end{minipage} & \begin{minipage}[b]{0.10\columnwidth}\raggedleft
Answer order: c\strut
\end{minipage} & \begin{minipage}[b]{0.10\columnwidth}\raggedleft
Answer order: d′\strut
\end{minipage} & \begin{minipage}[b]{0.10\columnwidth}\raggedleft
Base-rate sentence: c\strut
\end{minipage} & \begin{minipage}[b]{0.10\columnwidth}\raggedleft
Base-rate sentence: d′\strut
\end{minipage}\tabularnewline
\midrule
\endhead
\begin{minipage}[t]{0.10\columnwidth}\raggedright
Apertus 1.5 8B\strut
\end{minipage} & \begin{minipage}[t]{0.10\columnwidth}\raggedright
base\strut
\end{minipage} & \begin{minipage}[t]{0.10\columnwidth}\raggedleft
0.63\strut
\end{minipage} & \begin{minipage}[t]{0.10\columnwidth}\raggedleft
0.24\strut
\end{minipage} & \begin{minipage}[t]{0.10\columnwidth}\raggedleft
1.09\strut
\end{minipage} & \begin{minipage}[t]{0.10\columnwidth}\raggedleft
0.30\strut
\end{minipage} & \begin{minipage}[t]{0.10\columnwidth}\raggedleft
0.37\strut
\end{minipage} & \begin{minipage}[t]{0.10\columnwidth}\raggedleft
0.17\strut
\end{minipage}\tabularnewline
\begin{minipage}[t]{0.10\columnwidth}\raggedright
Apertus 1.5 8B\strut
\end{minipage} & \begin{minipage}[t]{0.10\columnwidth}\raggedright
final\strut
\end{minipage} & \begin{minipage}[t]{0.10\columnwidth}\raggedleft
0.97\strut
\end{minipage} & \begin{minipage}[t]{0.10\columnwidth}\raggedleft
0.18\strut
\end{minipage} & \begin{minipage}[t]{0.10\columnwidth}\raggedleft
0.08\strut
\end{minipage} & \begin{minipage}[t]{0.10\columnwidth}\raggedleft
0.16\strut
\end{minipage} & \begin{minipage}[t]{0.10\columnwidth}\raggedleft
0.14\strut
\end{minipage} & \begin{minipage}[t]{0.10\columnwidth}\raggedleft
0.05\strut
\end{minipage}\tabularnewline
\begin{minipage}[t]{0.10\columnwidth}\raggedright
Gemma 3 12B\strut
\end{minipage} & \begin{minipage}[t]{0.10\columnwidth}\raggedright
base\strut
\end{minipage} & \begin{minipage}[t]{0.10\columnwidth}\raggedleft
1.67\strut
\end{minipage} & \begin{minipage}[t]{0.10\columnwidth}\raggedleft
0.09\strut
\end{minipage} & \begin{minipage}[t]{0.10\columnwidth}\raggedleft
1.27\strut
\end{minipage} & \begin{minipage}[t]{0.10\columnwidth}\raggedleft
0.22\strut
\end{minipage} & \begin{minipage}[t]{0.10\columnwidth}\raggedleft
0.40\strut
\end{minipage} & \begin{minipage}[t]{0.10\columnwidth}\raggedleft
0.18\strut
\end{minipage}\tabularnewline
\begin{minipage}[t]{0.10\columnwidth}\raggedright
Gemma 3 12B\strut
\end{minipage} & \begin{minipage}[t]{0.10\columnwidth}\raggedright
final\strut
\end{minipage} & \begin{minipage}[t]{0.10\columnwidth}\raggedleft
2.93\strut
\end{minipage} & \begin{minipage}[t]{0.10\columnwidth}\raggedleft
0.10\strut
\end{minipage} & \begin{minipage}[t]{0.10\columnwidth}\raggedleft
0.26\strut
\end{minipage} & \begin{minipage}[t]{0.10\columnwidth}\raggedleft
0.08\strut
\end{minipage} & \begin{minipage}[t]{0.10\columnwidth}\raggedleft
0.61\strut
\end{minipage} & \begin{minipage}[t]{0.10\columnwidth}\raggedleft
0.19\strut
\end{minipage}\tabularnewline
\begin{minipage}[t]{0.10\columnwidth}\raggedright
Gemma 3 4B\strut
\end{minipage} & \begin{minipage}[t]{0.10\columnwidth}\raggedright
base\strut
\end{minipage} & \begin{minipage}[t]{0.10\columnwidth}\raggedleft
0.42\strut
\end{minipage} & \begin{minipage}[t]{0.10\columnwidth}\raggedleft
0.07\strut
\end{minipage} & \begin{minipage}[t]{0.10\columnwidth}\raggedleft
0.65\strut
\end{minipage} & \begin{minipage}[t]{0.10\columnwidth}\raggedleft
0.10\strut
\end{minipage} & \begin{minipage}[t]{0.10\columnwidth}\raggedleft
0.29\strut
\end{minipage} & \begin{minipage}[t]{0.10\columnwidth}\raggedleft
0.05\strut
\end{minipage}\tabularnewline
\begin{minipage}[t]{0.10\columnwidth}\raggedright
Gemma 3 4B\strut
\end{minipage} & \begin{minipage}[t]{0.10\columnwidth}\raggedright
final\strut
\end{minipage} & \begin{minipage}[t]{0.10\columnwidth}\raggedleft
3.28\strut
\end{minipage} & \begin{minipage}[t]{0.10\columnwidth}\raggedleft
0.15\strut
\end{minipage} & \begin{minipage}[t]{0.10\columnwidth}\raggedleft
0.53\strut
\end{minipage} & \begin{minipage}[t]{0.10\columnwidth}\raggedleft
0.10\strut
\end{minipage} & \begin{minipage}[t]{0.10\columnwidth}\raggedleft
0.65\strut
\end{minipage} & \begin{minipage}[t]{0.10\columnwidth}\raggedleft
0.18\strut
\end{minipage}\tabularnewline
\begin{minipage}[t]{0.10\columnwidth}\raggedright
OLMo 2 7B\strut
\end{minipage} & \begin{minipage}[t]{0.10\columnwidth}\raggedright
base\strut
\end{minipage} & \begin{minipage}[t]{0.10\columnwidth}\raggedleft
1.09\strut
\end{minipage} & \begin{minipage}[t]{0.10\columnwidth}\raggedleft
0.05\strut
\end{minipage} & \begin{minipage}[t]{0.10\columnwidth}\raggedleft
3.11\strut
\end{minipage} & \begin{minipage}[t]{0.10\columnwidth}\raggedleft
0.29\strut
\end{minipage} & \begin{minipage}[t]{0.10\columnwidth}\raggedleft
0.40\strut
\end{minipage} & \begin{minipage}[t]{0.10\columnwidth}\raggedleft
0.12\strut
\end{minipage}\tabularnewline
\begin{minipage}[t]{0.10\columnwidth}\raggedright
OLMo 2 7B\strut
\end{minipage} & \begin{minipage}[t]{0.10\columnwidth}\raggedright
final\strut
\end{minipage} & \begin{minipage}[t]{0.10\columnwidth}\raggedleft
0.47\strut
\end{minipage} & \begin{minipage}[t]{0.10\columnwidth}\raggedleft
0.13\strut
\end{minipage} & \begin{minipage}[t]{0.10\columnwidth}\raggedleft
0.16\strut
\end{minipage} & \begin{minipage}[t]{0.10\columnwidth}\raggedleft
0.12\strut
\end{minipage} & \begin{minipage}[t]{0.10\columnwidth}\raggedleft
0.28\strut
\end{minipage} & \begin{minipage}[t]{0.10\columnwidth}\raggedleft
0.09\strut
\end{minipage}\tabularnewline
\begin{minipage}[t]{0.10\columnwidth}\raggedright
OLMo 3 7B Think\strut
\end{minipage} & \begin{minipage}[t]{0.10\columnwidth}\raggedright
base\strut
\end{minipage} & \begin{minipage}[t]{0.10\columnwidth}\raggedleft
0.93\strut
\end{minipage} & \begin{minipage}[t]{0.10\columnwidth}\raggedleft
0.30\strut
\end{minipage} & \begin{minipage}[t]{0.10\columnwidth}\raggedleft
0.90\strut
\end{minipage} & \begin{minipage}[t]{0.10\columnwidth}\raggedleft
0.21\strut
\end{minipage} & \begin{minipage}[t]{0.10\columnwidth}\raggedleft
0.32\strut
\end{minipage} & \begin{minipage}[t]{0.10\columnwidth}\raggedleft
0.13\strut
\end{minipage}\tabularnewline
\begin{minipage}[t]{0.10\columnwidth}\raggedright
OLMo 3 7B Think\strut
\end{minipage} & \begin{minipage}[t]{0.10\columnwidth}\raggedright
final\strut
\end{minipage} & \begin{minipage}[t]{0.10\columnwidth}\raggedleft
0.10\strut
\end{minipage} & \begin{minipage}[t]{0.10\columnwidth}\raggedleft
0.09\strut
\end{minipage} & \begin{minipage}[t]{0.10\columnwidth}\raggedleft
0.68\strut
\end{minipage} & \begin{minipage}[t]{0.10\columnwidth}\raggedleft
0.13\strut
\end{minipage} & \begin{minipage}[t]{0.10\columnwidth}\raggedleft
0.34\strut
\end{minipage} & \begin{minipage}[t]{0.10\columnwidth}\raggedleft
0.07\strut
\end{minipage}\tabularnewline
\begin{minipage}[t]{0.10\columnwidth}\raggedright
Qwen3 8B\strut
\end{minipage} & \begin{minipage}[t]{0.10\columnwidth}\raggedright
base\strut
\end{minipage} & \begin{minipage}[t]{0.10\columnwidth}\raggedleft
1.73\strut
\end{minipage} & \begin{minipage}[t]{0.10\columnwidth}\raggedleft
0.04\strut
\end{minipage} & \begin{minipage}[t]{0.10\columnwidth}\raggedleft
0.54\strut
\end{minipage} & \begin{minipage}[t]{0.10\columnwidth}\raggedleft
0.12\strut
\end{minipage} & \begin{minipage}[t]{0.10\columnwidth}\raggedleft
1.19\strut
\end{minipage} & \begin{minipage}[t]{0.10\columnwidth}\raggedleft
0.23\strut
\end{minipage}\tabularnewline
\begin{minipage}[t]{0.10\columnwidth}\raggedright
Qwen3 8B\strut
\end{minipage} & \begin{minipage}[t]{0.10\columnwidth}\raggedright
final\strut
\end{minipage} & \begin{minipage}[t]{0.10\columnwidth}\raggedleft
0.07\strut
\end{minipage} & \begin{minipage}[t]{0.10\columnwidth}\raggedleft
0.24\strut
\end{minipage} & \begin{minipage}[t]{0.10\columnwidth}\raggedleft
0.06\strut
\end{minipage} & \begin{minipage}[t]{0.10\columnwidth}\raggedleft
0.12\strut
\end{minipage} & \begin{minipage}[t]{0.10\columnwidth}\raggedleft
0.03\strut
\end{minipage} & \begin{minipage}[t]{0.10\columnwidth}\raggedleft
0.10\strut
\end{minipage}\tabularnewline
\begin{minipage}[t]{0.10\columnwidth}\raggedright
All\strut
\end{minipage} & \begin{minipage}[t]{0.10\columnwidth}\raggedright
\strut
\end{minipage} & \begin{minipage}[t]{0.10\columnwidth}\raggedleft
1.19\strut
\end{minipage} & \begin{minipage}[t]{0.10\columnwidth}\raggedleft
0.14\strut
\end{minipage} & \begin{minipage}[t]{0.10\columnwidth}\raggedleft
0.78\strut
\end{minipage} & \begin{minipage}[t]{0.10\columnwidth}\raggedleft
0.16\strut
\end{minipage} & \begin{minipage}[t]{0.10\columnwidth}\raggedleft
0.42\strut
\end{minipage} & \begin{minipage}[t]{0.10\columnwidth}\raggedleft
0.13\strut
\end{minipage}\tabularnewline
\bottomrule
\end{longtable}

\textbf{Table F2.} Use of the stated base rate: mean shift of dv (nats)
between a stated share of 20\% and 80\% AI texts, averaged over three
corpora and both answer orders (six cells; range over cells in
brackets), its ratio to the shift of the prior log-odds (ln 16 = 2.77),
and the mean between-text SD of the shift within a cell. Wording: share
named first in the base-rate sentence.

\begin{longtable}[]{@{}lllrrr@{}}
\toprule
Model & Checkpoint & Wording & Shift {[}range{]} & Ratio to ln 16 &
Between-text SD\tabularnewline
\midrule
\endhead
Apertus 1.5 8B & base & AI first & 0.41 {[}0.33, 0.47{]} & 0.15 &
0.14\tabularnewline
Apertus 1.5 8B & final & AI first & 6.25 {[}3.06, 8.85{]} & 2.26 &
2.01\tabularnewline
Gemma 3 12B & base & AI first & 0.50 {[}0.42, 0.55{]} & 0.18 &
0.11\tabularnewline
Gemma 3 12B & final & AI first & 10.93 {[}6.48, 14.83{]} & 3.94 &
2.34\tabularnewline
Gemma 3 4B & base & AI first & 0.18 {[}0.14, 0.21{]} & 0.06 &
0.08\tabularnewline
Gemma 3 4B & base & human first & 0.16 {[}0.12, 0.20{]} & 0.06 &
0.08\tabularnewline
Gemma 3 4B & final & AI first & 10.65 {[}7.84, 12.78{]} & 3.84 &
1.54\tabularnewline
Gemma 3 4B & final & human first & 5.61 {[}4.08, 8.53{]} & 2.03 &
1.80\tabularnewline
OLMo 2 7B & base & AI first & 1.00 {[}0.54, 1.49{]} & 0.36 &
0.18\tabularnewline
OLMo 2 7B & base & human first & 1.35 {[}0.70, 2.13{]} & 0.49 &
0.23\tabularnewline
OLMo 2 7B & final & AI first & 5.39 {[}4.25, 6.54{]} & 1.94 &
0.89\tabularnewline
OLMo 2 7B & final & human first & 4.37 {[}3.75, 4.88{]} & 1.58 &
0.88\tabularnewline
OLMo 3 7B Think & base & AI first & 1.17 {[}0.56, 1.85{]} & 0.42 &
0.24\tabularnewline
OLMo 3 7B Think & base & human first & 1.07 {[}0.41, 1.89{]} & 0.38 &
0.21\tabularnewline
OLMo 3 7B Think & final & AI first & 1.21 {[}0.81, 1.44{]} & 0.44 &
0.19\tabularnewline
OLMo 3 7B Think & final & human first & 1.48 {[}1.00, 2.07{]} & 0.53 &
0.22\tabularnewline
Qwen3 8B & base & AI first & 1.40 {[}1.06, 1.68{]} & 0.51 &
0.27\tabularnewline
Qwen3 8B & final & AI first & 3.67 {[}3.28, 4.43{]} & 1.32 &
0.46\tabularnewline
\bottomrule
\end{longtable}

\hypertarget{g-individual-post-training-contrasts}{%
\subsection{G Individual post-training
contrasts}\label{g-individual-post-training-contrasts}}

Among the individual comparisons, six decreases survived Holm correction
within this family of 18 tests (all p\_Holm ≤ .013): Gemma 3 12B (Δd′ =
−0.95 {[}−1.28, −0.66{]}), Qwen3 (−0.57 {[}−0.79, −0.36{]}), Gemma 3 4B
(−0.50 {[}−0.72, −0.29{]}), Apertus (−0.47 {[}−0.71, −0.25{]}) and OLMo
2 (−0.32 {[}−0.52, −0.12{]}) on the textbook summaries, and Qwen3 on the
newspaper segments (−0.65 {[}−0.93, −0.42{]}). On the short chats, three
decreases of similar size (Gemma 3 4B: −0.51; Apertus: −0.47; Gemma 3
12B: −0.45; uncorrected p ≤ .008) did not survive Holm correction
(p\_Holm = .059--.074), because the chat estimates were less precise (61
independent originals compared with 119 pairs for the textbook
summaries); they remained significant under false-discovery-rate
control. Two increases were significant only without correction or under
false-discovery-rate control: OLMo 2 on the short chats (+0.35 {[}0.06,
0.63{]}) and OLMo 3 Think on the textbook summaries, the latter in a
condition with low answer probability. Note that we defined the family
of tests post hoc as the comparisons that address our hypothesis; Holm
correction across all 75 contrasts that we tested (of 129 computed),
including comparisons between adjacent post-training stages and across
prompt formats, left five of the six decreases significant. Note also
that three of the six decreases did not hold with another wording of the
prompt: For OLMo 2 on the textbook summaries, the change was +0.18 with
the paraphrased prompt, and for Qwen3 it was +0.10 on the textbook
summaries and +0.04 on the newspaper segments with the question placed
before the text (Section 3.2). Hence, post-training lowered the
sensitivity of the answers on average, but neither uniformly nor in a
single direction across models and text types.

\hypertarget{h-localization-further-analyses}{%
\subsection{H Localization: further
analyses}\label{h-localization-further-analyses}}

\emph{Sublayer types.} Both sublayer types contributed. Exchanging only
the attention blocks or only the feed-forward (MLP) blocks of the first
8 Apertus layers had partial effects in both directions (e.g., restoring
the base model's blocks under SFT: attention +0.17 {[}−0.06, 0.40{]},
MLP +0.18 {[}0.03, 0.37{]}, both together +0.59). In OLMo 2, the base
model's MLP blocks of the first 16 layers restored discrimination almost
completely (−0.09 {[}−0.29, 0.11{]}), whereas its attention blocks did
so only partly (−0.35 {[}−0.55, −0.19{]}). Within the limits just
described, the change thus appears to be distributed over the early
layers and to involve both attention and feed-forward blocks.

\emph{Question placed before the text.} See Section 3.4 for the
interpretation.

\textbf{Table H1.} Activation patching at layer 16 with the question
placed before the text (textbook summaries): d′ of the recipient after
receiving the donor's states at one set of positions; change relative to
the unpatched recipient with 95\% CI (grouped bootstrap) for the three
sets of interest.

\begin{longtable}[]{@{}lrrrrrr@{}}
\toprule
\begin{minipage}[b]{0.12\columnwidth}\raggedright
Recipient ← donor\strut
\end{minipage} & \begin{minipage}[b]{0.12\columnwidth}\raggedleft
Unpatched\strut
\end{minipage} & \begin{minipage}[b]{0.12\columnwidth}\raggedleft
Donor\strut
\end{minipage} & \begin{minipage}[b]{0.12\columnwidth}\raggedleft
Text\strut
\end{minipage} & \begin{minipage}[b]{0.12\columnwidth}\raggedleft
Question (before text)\strut
\end{minipage} & \begin{minipage}[b]{0.12\columnwidth}\raggedleft
After text\strut
\end{minipage} & \begin{minipage}[b]{0.12\columnwidth}\raggedleft
All positions\strut
\end{minipage}\tabularnewline
\midrule
\endhead
\begin{minipage}[t]{0.12\columnwidth}\raggedright
Apertus base ← SFT\strut
\end{minipage} & \begin{minipage}[t]{0.12\columnwidth}\raggedleft
+0.36\strut
\end{minipage} & \begin{minipage}[t]{0.12\columnwidth}\raggedleft
−0.19\strut
\end{minipage} & \begin{minipage}[t]{0.12\columnwidth}\raggedleft
+0.31 (−0.05 {[}−0.15, 0.05{]})\strut
\end{minipage} & \begin{minipage}[t]{0.12\columnwidth}\raggedleft
+0.55 (+0.19 {[}0.06, 0.32{]})\strut
\end{minipage} & \begin{minipage}[t]{0.12\columnwidth}\raggedleft
+0.13 (−0.23 {[}−0.46, −0.02{]})\strut
\end{minipage} & \begin{minipage}[t]{0.12\columnwidth}\raggedleft
−0.17\strut
\end{minipage}\tabularnewline
\begin{minipage}[t]{0.12\columnwidth}\raggedright
Apertus SFT ← base\strut
\end{minipage} & \begin{minipage}[t]{0.12\columnwidth}\raggedleft
−0.19\strut
\end{minipage} & \begin{minipage}[t]{0.12\columnwidth}\raggedleft
+0.36\strut
\end{minipage} & \begin{minipage}[t]{0.12\columnwidth}\raggedleft
−0.07 (+0.12 {[}0.04, 0.21{]})\strut
\end{minipage} & \begin{minipage}[t]{0.12\columnwidth}\raggedleft
+0.10 (+0.29 {[}0.21, 0.38{]})\strut
\end{minipage} & \begin{minipage}[t]{0.12\columnwidth}\raggedleft
+0.44 (+0.63 {[}0.43, 0.86{]})\strut
\end{minipage} & \begin{minipage}[t]{0.12\columnwidth}\raggedleft
+0.50\strut
\end{minipage}\tabularnewline
\begin{minipage}[t]{0.12\columnwidth}\raggedright
OLMo 2 base ← SFT\strut
\end{minipage} & \begin{minipage}[t]{0.12\columnwidth}\raggedleft
+0.09\strut
\end{minipage} & \begin{minipage}[t]{0.12\columnwidth}\raggedleft
−0.42\strut
\end{minipage} & \begin{minipage}[t]{0.12\columnwidth}\raggedleft
+0.06 (−0.04 {[}−0.13, 0.07{]})\strut
\end{minipage} & \begin{minipage}[t]{0.12\columnwidth}\raggedleft
−0.08 (−0.18 {[}−0.30, −0.04{]})\strut
\end{minipage} & \begin{minipage}[t]{0.12\columnwidth}\raggedleft
−0.16 (−0.25 {[}−0.39, −0.10{]})\strut
\end{minipage} & \begin{minipage}[t]{0.12\columnwidth}\raggedleft
−0.43\strut
\end{minipage}\tabularnewline
\begin{minipage}[t]{0.12\columnwidth}\raggedright
OLMo 2 SFT ← base\strut
\end{minipage} & \begin{minipage}[t]{0.12\columnwidth}\raggedleft
−0.42\strut
\end{minipage} & \begin{minipage}[t]{0.12\columnwidth}\raggedleft
+0.09\strut
\end{minipage} & \begin{minipage}[t]{0.12\columnwidth}\raggedleft
−0.40 (+0.01 {[}−0.06, 0.08{]})\strut
\end{minipage} & \begin{minipage}[t]{0.12\columnwidth}\raggedleft
−0.33 (+0.09 {[}−0.03, 0.21{]})\strut
\end{minipage} & \begin{minipage}[t]{0.12\columnwidth}\raggedleft
−0.17 (+0.24 {[}0.09, 0.40{]})\strut
\end{minipage} & \begin{minipage}[t]{0.12\columnwidth}\raggedleft
+0.08\strut
\end{minipage}\tabularnewline
\bottomrule
\end{longtable}

\hypertarget{supplementary-figures}{%
\subsection{Supplementary figures}\label{supplementary-figures}}

\setcounter{figure}{0}\renewcommand{\thefigure}{S\arabic{figure}}

\begin{figure}
\centering
\includegraphics[width=1\textwidth,height=0.68\textheight]{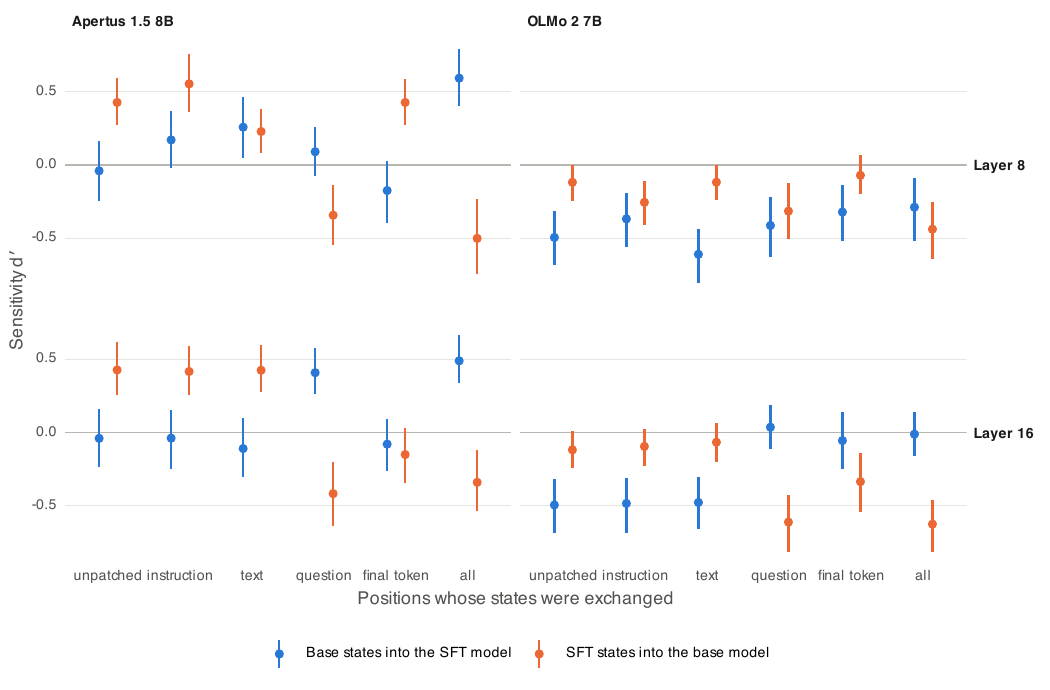}
\caption{Activation patching at layers 8 and 16 (textbook summaries). As
Figure 5B, with layer 8 added. At layer 8 the picture is mixed: in
Apertus, exchanging the states of the text positions still has a partial
effect; in OLMo 2, even exchanging all positions transfers only part of
the difference (the change is not yet complete at layer 8; cf.~the
stitching curve in Figure 5A). At layer 16, the whole difference between
base and SFT model is carried by the question positions. Error bars are
95\% bootstrap confidence intervals.}
\end{figure}

\begin{figure}
\centering
\includegraphics[width=1\textwidth,height=0.68\textheight]{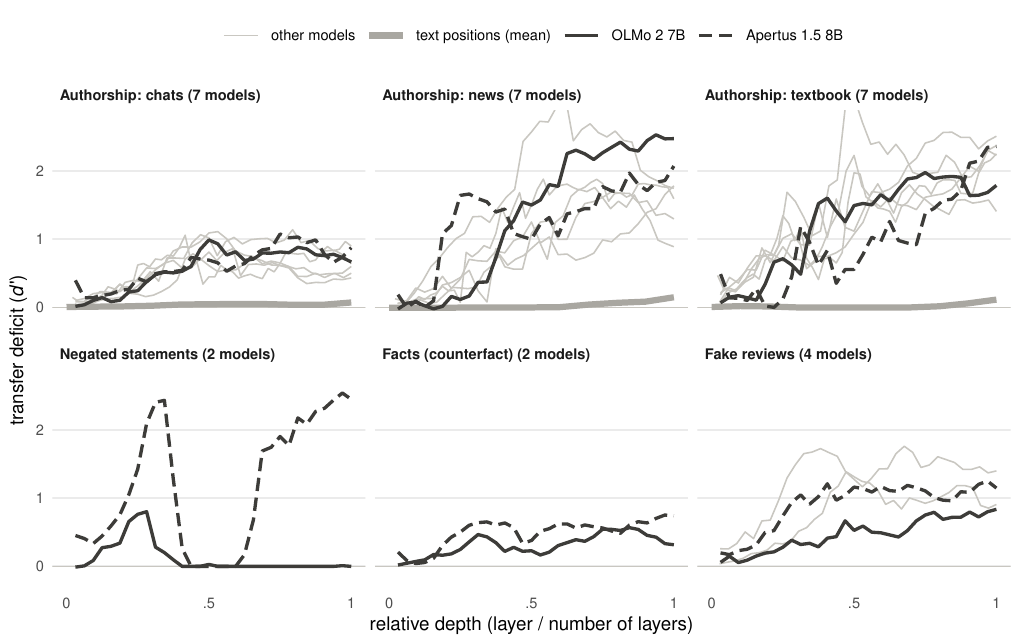}
\caption{Post-training changes the text positions little and reorganizes
the answer position. For each layer, a linear probe was trained on one
checkpoint and tested on the other (base vs.~final post-trained model,
both directions averaged); the transfer deficit is the loss in d′
relative to the probe trained on the checkpoint itself. A deficit of 0
means that the same direction separates the classes in both checkpoints.
Dark lines: OLMo 2 7B (solid) and Apertus 1.5 8B (dashed), the two
models included in all six sets; thin grey lines: the other models (OLMo
3 Think and Instruct, which share a base model, counted separately);
thick light line: mean over all models at the text positions (authorship
corpora only). At the text positions, probes transferred with little
loss at every depth (largest deficit 0.54; on the chats, where they were
well below ceiling, 0.16). At the last prompt position, at which the
answer is produced, the deficit grew with depth in all models and tasks,
although probes trained on each checkpoint itself still decoded the
information. For the negated statements, the deficit drops to zero at
mid-depth because both checkpoints are at ceiling there (AUC = 1); OLMo
2 shows no deficit in the upper layers for this set.}
\end{figure}

\end{document}